\documentclass{article}

\usepackage[final]{neurips_2022}

\usepackage[utf8]{inputenc} 
\usepackage[T1]{fontenc}    
\usepackage{hyperref}       
\usepackage{url}            
\usepackage{booktabs}       
\usepackage{amsfonts}       
\usepackage{nicefrac}       
\usepackage{microtype}      
\usepackage{xcolor}         
\usepackage{amsmath}         
\usepackage{graphics}   
\usepackage{graphicx} 
\usepackage{algorithm}  
\usepackage{algorithmic}  
\usepackage{color}
\usepackage{subfigure}
\title{CUP: Critic-Guided Policy Reuse}

\author{%
	Jin Zhang$^1$, Siyuan Li$^2$, {Chongjie Zhang}$^1$ \\
	$^1$Institute for Interdisciplinary Information Sciences, Tsinghua University, China \\
	$^2$School of Computer Science and Technology, Harbin Institute of Technology, China \\
	\texttt{jin-zhan20@mails.tsinghua.edu.cn}\\
	\texttt{lisiyuan199511@gmail.com}\\
	\texttt{chongjie@tsinghua.edu.cn} \\
}

\newtheorem{thm}{Theorem}
\begin{document}

\maketitle

\begin{abstract}
  The ability to reuse previous policies is an important aspect of human intelligence. To achieve efficient policy reuse, a Deep Reinforcement Learning (DRL) agent needs to decide when to reuse and which source policies to reuse. Previous methods solve this problem by introducing extra components to the underlying algorithm, such as hierarchical high-level policies over source policies, or estimations of source policies' value functions on the target task. However, training these components induces either optimization non-stationarity or heavy sampling cost, significantly impairing the effectiveness of transfer. To tackle this problem, we propose a novel policy reuse algorithm called Critic-gUided Policy reuse (CUP), which avoids training any extra components and efficiently reuses source policies. CUP utilizes the critic, a common component in actor-critic methods, to evaluate and choose source policies. At each state, CUP chooses the source policy that has the largest one-step improvement over the current target policy, and forms a guidance policy. The guidance policy is theoretically guaranteed to be a monotonic improvement over the current target policy. Then the target policy is regularized to imitate the guidance policy to perform efficient policy search. Empirical results demonstrate that CUP achieves efficient transfer and significantly outperforms baseline algorithms.
\end{abstract}

\section{Introduction}
Human intelligence can solve new tasks quickly by reusing previous policies \citep{guberman1991learning}. Despite remarkable success, current Deep Reinforcement Learning (DRL) agents lack this knowledge transfer ability \citep{silver2017mastering,vinyals2019grandmaster,ceron2021revisiting}, leading to enormous computation and sampling cost. As a consequence, a large number of works have been studying the problem of policy reuse in DRL, i.e.,  how to efficiently reuse source policies to speed up target policy learning \citep{fernandez2006probabilistic,barreto2018transfer,li2019hierarchical,yang2020efficient}.

A fundamental challenge towards policy reuse is: how does an agent with access to multiple source policies decide when and where to use them \citep{fernandez2006probabilistic,kurenkov2020ac,cheng2020policy}?
Previous methods solve this problem by introducing additional components to the underlying DRL algorithm, such as hierarchical high-level policies over source policies \citep{li2018context,li2019hierarchical,yang2020efficient}, or estimations of source policies' value functions on the target task \citep{barreto2017successor,barreto2018transfer,cheng2020policy}. However, training these components significantly impairs the effectiveness of transfer, as hierarchical structures induce optimization non-stationarity \citep{pateria2021hierarchical}, and estimating the value functions for every source policy is computationally expensive and with high sampling cost.
Thus, the objective of this study is to address the question:

\begin{quote}
    \emph{Can we achieve efficient transfer without training additional components?}
\end{quote}

Notice that actor-critic methods \citep{lillicrap2016continuous,fujimoto2018addressing,haarnoja2018soft2} learn a critic that approximates the actor's Q function and serves as a natural way to evaluate policies. Based on this observation, we propose a novel policy reuse algorithm that utilizes the critic to choose source policies. The proposed algorithm, called \emph{Critic-gUided Policy reuse (CUP)}, avoids training any additional components and achieves efficient transfer. At each state, CUP chooses the source policy that has the largest one-step improvement over the current target policy, thus forming a guidance policy. Then CUP guides learning by regularizing the target policy to imitate the guidance policy. This approach has the following advantages. First, the one-step improvement can be estimated simply by querying the critic, and no additional components are needed to be trained. Secondly, the guidance policy is theoretically guaranteed to be a monotonic improvement over the current target policy, which ensures that CUP can reuse the source policies to improve the current target policy. Finally, CUP is conceptually simple and easy to implement, introducing 
very few hyper-parameters to the underlying algorithm.

We evaluate CUP on Meta-World \citep{yu2020meta}, a popular reinforcement learning benchmark composed of multiple robot arm manipulation tasks. Empirical results demonstrate that CUP achieves efficient transfer and significantly outperforms baseline algorithms. 

\section{Preliminaries}
\label{pre}
Reinforcement learning (RL) deals with Markov Decision Processes (MDPs). A MDP can be modelled by a tuple $(\mathcal{S},\mathcal{A},r,p,\gamma)$, with state space $\mathcal{S}$, action space $\mathcal{A}$, reward function $r(s,a)$, transition function $p(s'|s,a)$, and discount factor $\gamma$ \citep{sutton2018reinforcement}. In this study, we focus on MDPs with continuous action spaces. RL's objective is to find a policy $\pi(a|s)$ that maximizes the cumulative discounted return $R(\pi)=\mathbb{E}_{\pi}\left[\sum_{t=0}^{\infty}\gamma^tr(s_t,a_t)\right]$. 

While CUP is generally applicable to a wide range of actor-critic algorithms, in this work we use SAC \citep{haarnoja2018soft2} as the underlying algorithm. The soft Q function and soft V function \citep{haarnoja2017reinforcement} of a policy $\pi$ are defined as:

\begin{equation}
    Q_{\pi}(s,a)=r(s,a)+\gamma\mathbb{E}_{s' \sim p(\cdot|s,a)}\left[V_{\pi}(s)\right]
\end{equation}

\begin{equation}
    V_{\pi}(s)=\mathbb{E}_{a \sim \pi(\cdot|s)}\left[Q_{\pi}(s,a)-\alpha\log\pi(a|s)\right],
\end{equation}
\ where $\alpha>0$ is the entropy weight. SAC's loss functions are defined as:
 \begin{equation}
\begin{split}
  &L_{critic}(Q_{\theta})=\mathbb{E}_{(s,a,r,s') \sim \mathcal{D}}\left[Q_{\theta}(s,a)-(r+\gamma{V_{\overline{\theta}}}(s'))\right]^2 \\
  &L_{actor}(\pi_{\phi})=\mathbb{E}_{s \sim \mathcal{D}}\left[\mathbb{E}_{a \sim \pi_{\phi}(\cdot|s)}\left[\alpha \log \pi_{\phi}(a|s)-Q_{\theta}(s,a)\right]\right]   \\
  &L_{entropy}(\alpha)=\mathbb{E}_{s \sim \mathcal{D}}\left[\mathbb{E}_{a \sim \pi_{\phi}(\cdot|s)}\left[-\alpha \log \pi_{\phi}(a|s)-\alpha \overline{\mathcal{H}}\right]\right],
\end{split}
\label{sacloss2}
\end{equation}{}
 \ where $\mathcal{D}$ is the replay buffer, $\overline{\mathcal{H}}$ is a hyper-parameter representing the target entropy, $\theta$ and $\phi$ are network parameters, $\overline{\theta}$ is target network's parameters, and ${V_{\overline{\theta}}}(s)=\mathbb{E}_{a \sim \pi(a|s)}[Q_{\overline{\theta}}(s,a)-\alpha \log \pi(a|s)]$ is the target soft value function.

We define the \textit{soft expected advantage} of action probability distribution $\pi_i(\cdot|s)$ over policy $\pi_j$ at state $s$ as:
 \begin{equation}
    EA_{\pi_j}(s,\pi_i)=\mathbb{E}_{a \sim \pi_i(\cdot|s)}\left[Q_{\pi_j}(s,a)-\alpha\log\pi_i(a|s)-V_{\pi_j}(s)\right].
    \label{ea}
\end{equation}
\ $EA_{\pi_j}(s,\pi_i)$ measures the one-step performance improvement brought by following $\pi_i$ instead of $\pi_j$ at state $s$, and following $\pi_j$ afterwards.

The field of policy reuse focuses on solving a target MDP $M$ efficiently by transferring knowledge from a set of source policies $\{\pi_1,\pi_2,...,\pi_n\}$. We denote the target policy learned on $M$ at iteration $t$ as $\pi_{tar}^{t}$, and its corresponding soft Q function as $Q_{\pi_{tar}^t}$. In this work, we assume that the source policies and the target policy share the same state and action spaces.



\section{Critic-Guided Policy Reuse}
This section presents CUP, an efficient policy reuse algorithm that does not require training any additional components. CUP is built upon actor-critic methods. In each iteration, CUP uses the critic to form a guidance policy from the source policies and the current target policy. Then CUP guides policy search by regularizing the target policy to imitate the guidance policy. Section \ref{3.1} presents how to form a guidance policy by aggregating source policies through the critic, and proves that the guidance policy is guaranteed to be a monotonic improvement over the current target policy. We also prove that the target policy is theoretically guaranteed to improve by imitating the guidance policy. Section \ref{3.2} presents the overall framework of CUP.

\subsection{Critic-Guided Source Policy Aggregation}
\label{3.1}
CUP utilizes action probabilities proposed by source policies to improve the current target policy, and forms a guidance policy. 
At iteration $t$ of target policy learning, for each state $s$, the agent has access to a set of candidate action probability distributions proposed by the $n$ source policies and the current target policy: $\Pi_{t}^{s}=\{\pi_1(\cdot |s),\pi_2(\cdot |s),...,\pi_n(\cdot |s),\pi_{tar}^{t}(\cdot |s)\}$. The guidance policy $\pi_g^t$ can be formed by combining the action probability distributions that have the largest soft expected advantage over $\pi_{tar}^{t}$ at each state $s$:
\begin{equation}
    \pi_g^t(\cdot |s)=\mathop{\arg\max}\limits_{\pi(\cdot |s) \in \Pi_{t}^s}EA_{\pi_{tar}^t}(s,\pi)=\mathop{\arg\max}\limits_{\pi(\cdot |s)\in \Pi_{t}^{s}}\mathbb{E}_{a \sim\pi(\cdot |s)}\left[{Q}_{\pi_{tar}^t}(s,a)-\alpha\log\pi(a|s)\right]\ for\ all \ s\in\mathcal{S}.
    \label{arrgrgate}
\end{equation}
\ The second equation holds as adding $V_{\pi_{tar}^{t}}(s)$ to all soft expected advantages does not affect the result of the ${\arg\max}$ operator. Eq. \ref{arrgrgate} implies that at each state, we can choose which source policy to follow simply by querying its expected soft Q value under $\pi_{tar}^{t}$. Noticing that with function approximation, the exact soft Q value cannot be acquired. The following theorem enables us to form the guidance policy with an approximated soft Q function, and guarantees that the guidance policy is a monotonic improvement over the current target policy.
\begin{thm}
Let $\widetilde{Q}_{\pi_{tar}^t}$ be an approximation of ${Q}_{\pi_{tar}^t}$ such that 
\begin{equation}
    |\widetilde{Q}_{\pi_{tar}^t}(s,a)-{Q}_{\pi_{tar}^t}(s,a)|\leq \epsilon \text{\ for\  all}\  s \in \mathcal{S}, a \in A.
\end{equation}
Define 
\begin{equation}
   \widetilde{\pi_g^t}(\cdot |s)=\mathop{\arg\max}\limits_{\pi(\cdot |s) \in \Pi_{t}^s}\mathbb{E}_{a \sim \pi(\cdot |s)}\left[\widetilde{Q}_{\pi_{tar}^t}(s,a)-\alpha\log\pi(a|s)\right] \text{\ for\  all\ } s \in \mathcal{S}.
   \label{approximatepi}
\end{equation}
Then, 
\begin{equation}
   V_{\widetilde{\pi_g^t}}(s)\geq V_{\pi_{tar}^t}(s) - \frac{2\epsilon}{1-\gamma} \text{\ for\  all\ } s \in \mathcal{S}.
\end{equation}
\label{thm2}
\end{thm}
\ Theorem \ref{thm2} provides a way to choose source policies using an approximation of the current target policy's soft Q value. As SAC learns such an approximation, the guidance policy can be formed without training any additional components.

The next question is, how to incorporate the guidance policy $\widetilde{\pi_g^t}$ into target policy learning? The following theorem demonstrates that policy improvement can be guaranteed if the target policy is optimized to stay close to the guidance policy.
\begin{thm}
If 
\begin{equation}
   D_{KL}\left(\pi_{tar}^{t+1}(\cdot|s)||\widetilde{\pi_g^t}(\cdot|s)\right)\leq \delta\  for\  all\  s \in \mathcal{S},
\end{equation}
\ then 
\begin{equation}
   V_{\pi_{tar}^{t+1}}(s) \geq V_{\pi_{tar}^t}(s)-\frac{\sqrt{2\ln{2}\delta}(\widetilde{R}_{max}+\alpha\mathcal{H}_{max}^{t+1})}{(1-\gamma)^2}- \frac{2\epsilon+\alpha\widetilde{\mathcal{H}}_{max}}{1-\gamma}\  for\  all\  s \in \mathcal{S},
\end{equation}
\ where $\widetilde{R}_{max}=\max\limits_{s,a}|r(s,a)|$ is the largest possible absolute value of the reward, $\mathcal{H}_{max}^{t+1}=\max\limits_{s}\mathcal{H}(\pi_{tar}^{t+1}(\cdot|s))$ is the largest entropy of $\pi_{tar}^{t+1}$, and $\widetilde{\mathcal{H}}_{max}=\max\limits_{s}\left|\mathcal{H}(\pi_{tar}^{t}(\cdot|s))-\mathcal{H}(\pi_{tar}^{t+1}(\cdot|s))\right|$ is the largest possible absolute difference of the policy entropy.
\label{thm3}
\end{thm}
\ According to Theorem \ref{thm3}, the target policy can be improved by minimizing the KL divergence between the target policy and the guidance policy. Thus we can use the KL divergence as an auxiliary loss to guide target policy learning. Proofs of this section are deferred to Appendix \ref{proof1} and Appendix \ref{proof2}. Theorem \ref{thm2} and Theorem \ref{thm3} can be extended to common ``hard'' value functions (deferred to Appendix \ref{hard}), so CUP is also applicable to actor-critic algorithms that uses ``hard'' Bellman updates, such as A3C \citep{mnih2016asynchronous}.
\subsection{CUP Framework}
\label{3.2}
\begin{figure}[ht]
\centering
        \includegraphics[width=\columnwidth]{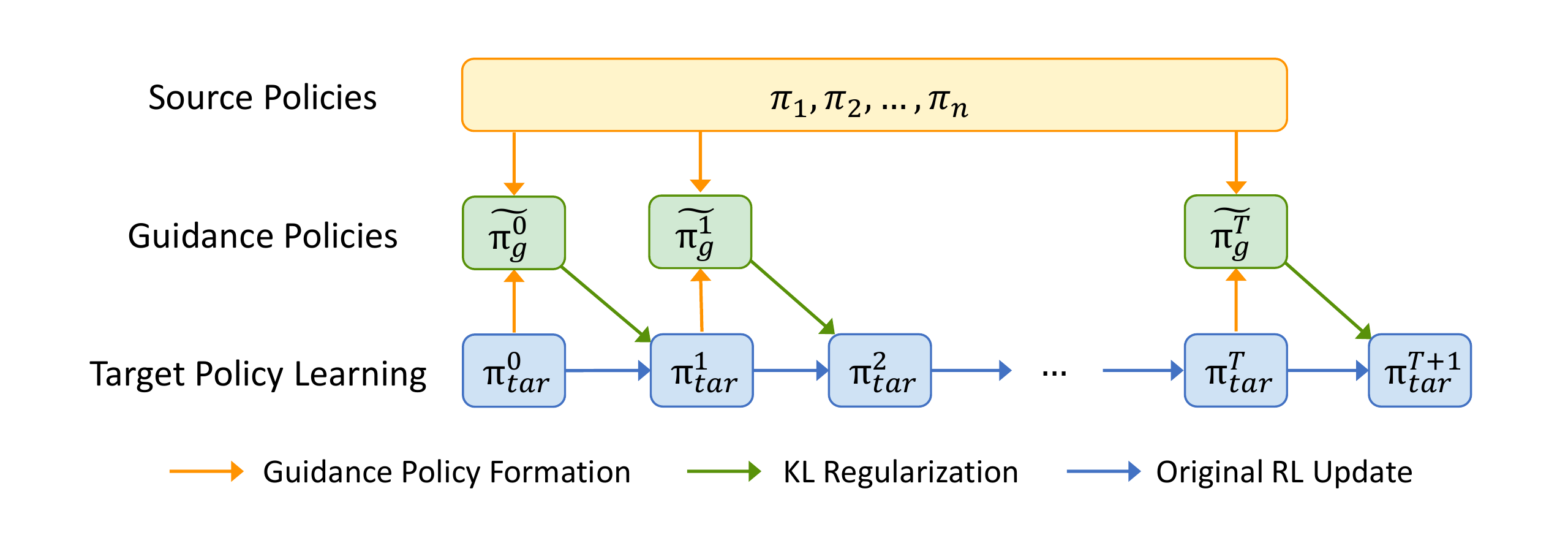}
\caption{CUP framework. In each iteration, CUP first forms a guidance policy by querying the critic, then guides policy learning by adding a KL regularization to policy search.}
\label{algorithm}
\end{figure}

In this subsection we propose the overall framework of CUP. As shown in Fig. \ref{algorithm}, at each iteration $t$, CUP first forms a guidance policy $\widetilde{\pi_g^t}$ according to Eq. \ref{approximatepi}, then provides additional guidance to policy search by regularizing the target policy $\pi_{tar}^{t+1}$ to imitate $\widetilde{\pi_g^t}$ \citep{wu2019behavior,fujimoto2021minimalist}. Specifically, CUP minimizes the following loss to optimize $\pi_{tar}^{t+1}$:
\begin{equation}
    L_{CUP}(\pi_{tar}^{t+1})=L_{actor}(\pi_{tar}^{t+1})+\mathbb{E}_{s \sim \mathcal{D}} \left[\beta_s D_{KL}\left(\pi_{tar}^{t+1}\left(\cdot|s\right)||\widetilde{\pi_g^t}\left(\cdot|s\right)\right)\right],
\end{equation}
\ where $L_{actor}$ is the original actor loss defined in Eq. \eqref{sacloss2}, and $\beta_s > 0$ is a hyper-parameter controlling the weight of regularization. In practice, we find that using a fixed weight for regularization has two problems. First, it is difficult to balance the scale between $L_{actor}$ and the regularization term, because $L_{actor}$ grows as the Q value gets larger. Secondly, a fixed weight cannot reflect the agent's confidence on $\widetilde{\pi_g^t}$. For example, when no source policies have positive soft expected advantages, $\widetilde{\pi_g^t}={\pi_{tar}^{t}}$. Then the agent should not imitate $\widetilde{\pi_g^t}$ anymore, as $\widetilde{\pi_g^t}$ cannot provide any guidance to further improve performance. Noticing that the soft expected advantage serves as a natural confidence measure, we weight the KL divergence with corresponding soft expected advantage at that state:
 \begin{equation}
    \beta_s=\beta_1 \min\left( \widetilde{EA}_{{\pi_{tar}^{t}}}(s,\widetilde{\pi_g^t}),\beta_2 |\widetilde{V}_{\pi_{tar}^{t}}\left(s)\right|\right),
    \label{12}
\end{equation}
 where  $\widetilde{EA}_{{\pi_{tar}^{t}}}(s,\widetilde{\pi_g^t})=\mathbb{E}_{a \sim \widetilde{\pi_g^t}(\cdot|s)}\left[\widetilde{Q}_{\pi_{tar}^{t}}(s,a)-\alpha\log\pi_g^t(a|s)-\widetilde{V}_{\pi_{tar}^{t}}(s)\right]$ is the approximated soft expected advantage, $\beta_1,\beta_2>0$ are two hyper-parameters, and $\widetilde{V}_{\pi_{tar}^{t}}(s)=\mathbb{E}_{a \sim \pi_{tar}^t(\cdot|s)}\left[\widetilde{Q}_{\pi_{tar}^{t}}\left(s,a\right)-\alpha\log\pi_{tar}^{t}(a|s)\right]$ is the approximated soft value function. This adaptive regularization weight automatically balances between the two losses, and ignores the regularization term at states where $\widetilde{\pi_g^t}$ cannot improve over $\pi_{tar}^t$ anymore. We further upper clip the expected advantage with the absolute value of $\beta_2 \widetilde{V}_{\pi_{tar}^{t}}$ to avoid the agent being overly confident about $\widetilde{\pi_g^t}$ due to function approximation error $\epsilon$.
 
CUP's pseudo-code is presented in Alg. \ref{CUP pseudo}. The modifications CUP made to SAC are marked in red. Additional implementation details are deferred to Appendix \ref{aid}.

 \begin{algorithm}[t]
   \caption{CUP}
   \label{alg:MetaCUP}
\begin{algorithmic}
   \STATE {\bfseries Require:} Source policies $\{\pi_1,\pi_2,...,\pi_n\}$, hyper-parameters $\lambda_{\theta_1}, \lambda_{\theta_2}, \lambda_{\pi}, \lambda_{\alpha}, \tau, {\overline{\mathcal{H}}}$, {\color{red}{$\beta_1,\beta_2$}}
   \STATE Initialize replay buffer $\mathcal{D}$
   \STATE Initialize actor $\pi_{\phi}$, entropy weight $\alpha$, critic $Q_{\theta_1}$,$Q_{\theta_2}$, target networks $Q_{{\overline{\theta}_1}} \leftarrow Q_{\theta_1},Q_{{\overline{\theta}_2}} \leftarrow Q_{\theta_2}$
   \WHILE{not done} 
   \FOR{each environment step}
   \STATE $a_t \sim \pi_{\theta}$
   \STATE $s_{t+1} \sim p(s_{t+1}|s_t,a_t)$
   \STATE $\mathcal{D} \leftarrow \mathcal{D} \cup \{s_t,a_t,r(s_t,a_t),s_{t+1}\}$
   \ENDFOR
   \FOR{each gradient step}
   \STATE Sample minibatch $b$ from $\mathcal{D}$
   \STATE {\color{red}Query source policies' action probabilities $\{\pi_1(\cdot|s),\pi_2(\cdot|s),...,\pi_n(\cdot|s)\}$ for states in $b$}
   \STATE {\color{red}Compute expected advantages according to Eq. \eqref{ea}, form $\widetilde{\pi_g^t}$ according to Eq. \eqref{approximatepi}}
   \STATE $\theta_i \leftarrow \theta_i -\lambda_Q \hat{\nabla}_{\theta_i}L_{critic}(Q_{\theta_i})$ for $i \in \{1,2\}$
   \STATE $\phi \leftarrow \phi - \lambda_{\pi} \hat{\nabla}_{\phi} {\color{red}L_{CUP}(\pi_\phi)}$
   \STATE $\alpha \leftarrow \alpha - \lambda_{\alpha}\hat{\nabla}_{\alpha}L_{entropy}(\alpha)$
   \STATE $\overline{\theta}_i \leftarrow \tau \theta_i + (1-\tau) \overline{\theta}_i$ for $i \in \{1,2\}$
   \ENDFOR
   \ENDWHILE
\end{algorithmic}
\label{CUP pseudo}
\end{algorithm}
\section{Experiments}
\label{exp}
We evaluate on Meta-World \citep{yu2020meta}, a popular reinforcement learning benchmark composed of multiple robot manipulation tasks. These tasks are both correlated (performed by the same Sawyer robot arm) and distinct (interacting with different objects and having different reward functions), and serve as a proper evaluation benchmark for policy reuse. The source policies are achieved by training on three representative tasks: Reach, Push, and Pick-Place. We choose several complex tasks as target tasks, including Hammer, Peg-Insert-Side, Push-Wall, Pick-Place-Wall, Push-Back, and Shelf-Place. Among these target tasks, Hammer and Peg-Insert-Side require interacting with objects unseen in the source tasks. In Push-Wall and Pick-Place-Wall, there is a wall between the object and the goal. In Push-Back, the goal distribution is different from Push. In Shelf-Place, the robot is required to put a block on a shelf, and the shelf is unseen in the source tasks. Video demonstrations of these tasks are available at \url{https://meta-world.github.io/}. Similar to the settings in \cite{yang2020multi}, in our experiments the goal position is randomly reset at the start of every episode. Codes are available at \url{https://github.com/NagisaZj/CUP}.


\subsection{Transfer Performance on Meta-World}
\label{tp}
\begin{figure}[t]
\centering
        \subfigure{\includegraphics[width=0.32\columnwidth]{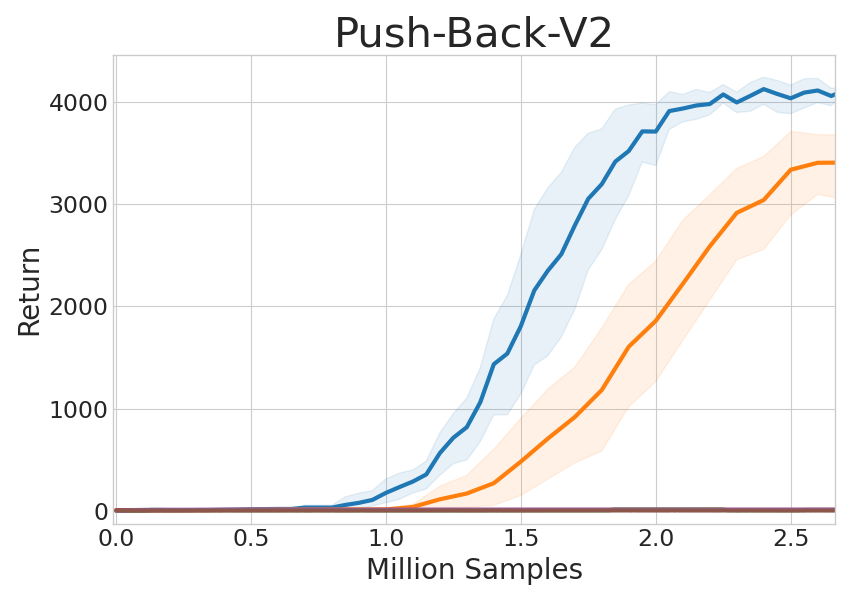}}
        \subfigure{\includegraphics[width=0.32\columnwidth]{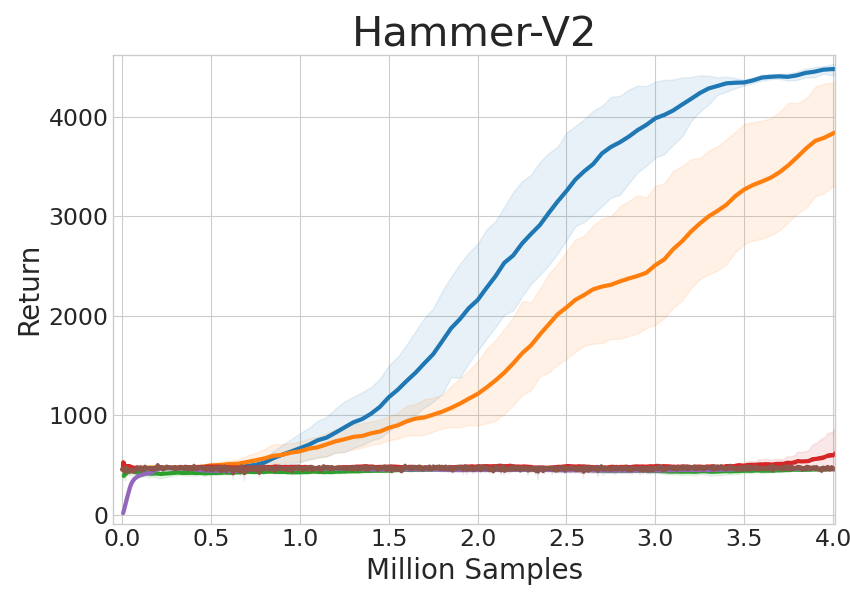}}
        \subfigure{\includegraphics[width=0.32\columnwidth]{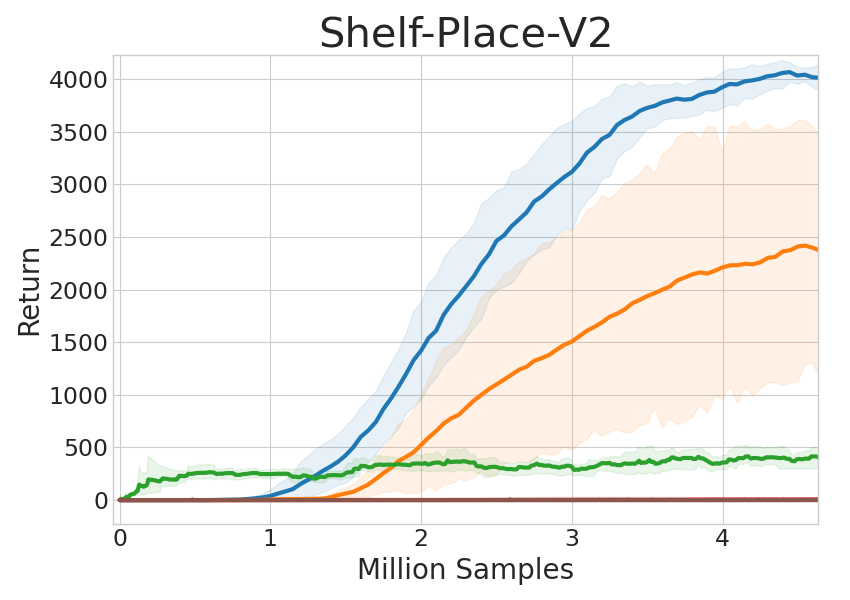}}
        \subfigure{\includegraphics[width=0.32\columnwidth]{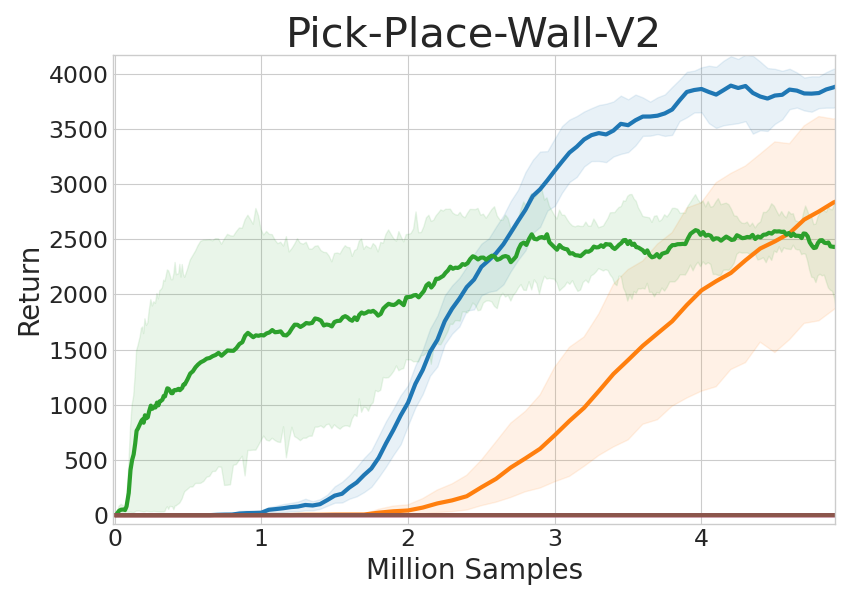}}
        \subfigure{\includegraphics[width=0.32\columnwidth]{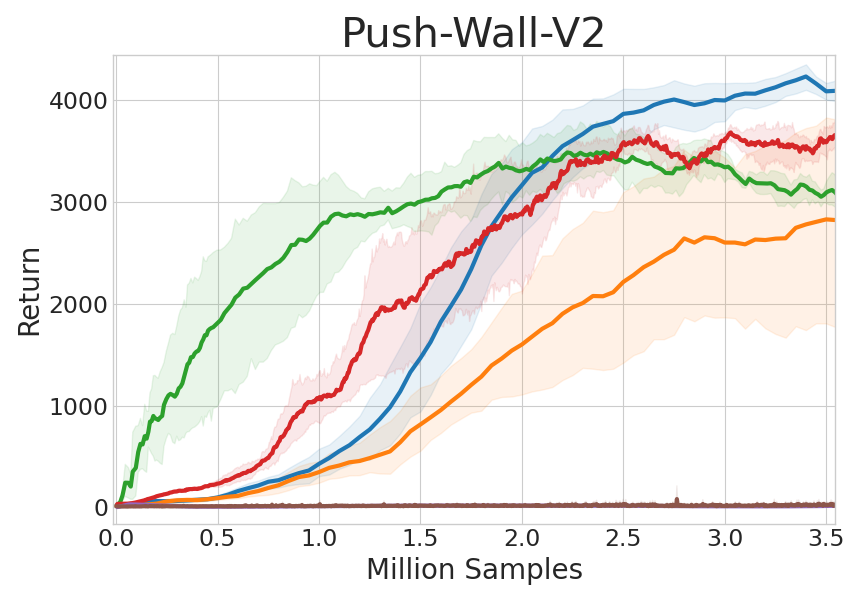}}
        \subfigure{\includegraphics[width=0.32\columnwidth]{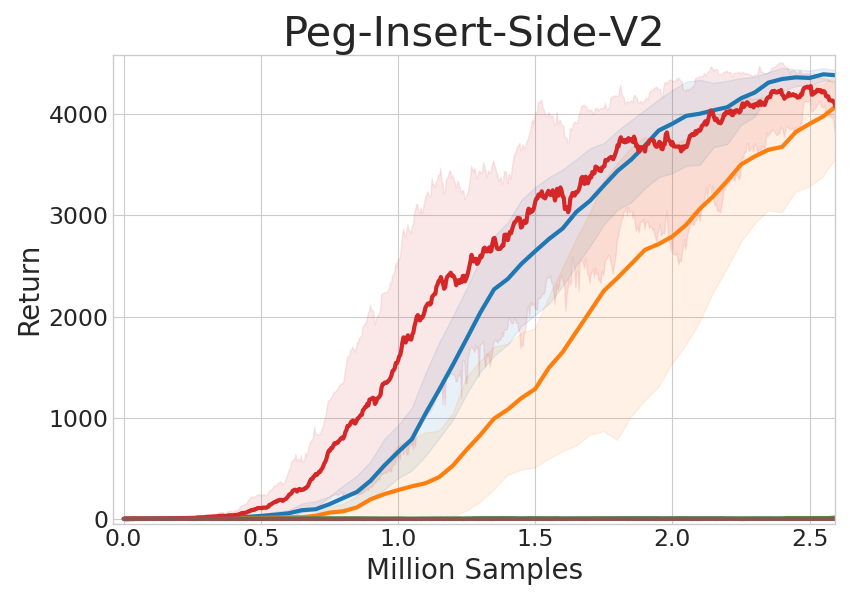}}
        \subfigure{\includegraphics[width=0.95\columnwidth]{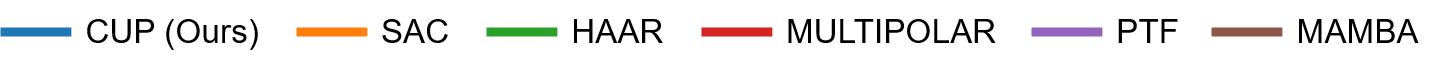}}
\caption{Evaluation of CUP and several baselines on various Meta-World tasks. Dashed areas represent 95\% bootstrapped confidence intervals. CUP achieves substantially better performance than baseline algorithms.}
\label{curves}
\end{figure}

We compare against several representative baseline algorithms, including HAAR \citep{li2019hierarchical}, PTF \citep{yang2020efficient}, MULTIPOLAR \citep{barekatain2021multipolar}, and MAMBA \citep{cheng2020policy}. Among these algorithms, HAAR and PTF learn hierarchical high-level policies over source policies. MAMBA aggregates source policies' V functions to form a baseline function, and performs policy improvement over the baseline function. MULTIPOLAR learns a weighted sum of source policies' action probabilities, and learns an additional network to predict residuals. We also compare against the original SAC algorithm. All the results are averaged over six random seeds. As shown in Figure \ref{curves}, CUP is the only algorithm that achieves efficient transfer on all six tasks, significantly outperforming the original SAC algorithm. HAAR has a jump-start performance on Push-Wall and Pick-Pick-Wall, but fails to further improve due to optimization non-stationarity induced by jointly training high-level and low-level policies. MULTIPOLAR achieves comparable performance on Push-Wall and Peg-Insert-Side, because the Push source policy is useful on Push-Wall (implied by HAAR's good jump-start performance), and learning residuals on Peg-Insert-Side is easier (implied by SAC's fast learning). In Pick-Place-Wall, the Pick-Place source policy is useful, but the residual is difficult to learn, so MULTIPOLAR does not work. For the remaining three tasks, the source policies are less useful, and MULTIPOLAR fails on these tasks.
PTF fails as its hierarchical policy only gets updated when the agent chooses similar actions to one of the source policies, which is quite rare when the source and target tasks are distinct. MAMBA fails as estimating all source policies' V functions accurately is sampling inefficient. Algorithm performance evaluated by success rate is deferred to Appendix \ref{sre}. 

\subsection{Analyzing the Guidance Policy}
\begin{figure}[tb]
\centering
        \includegraphics[width=0.8\columnwidth]{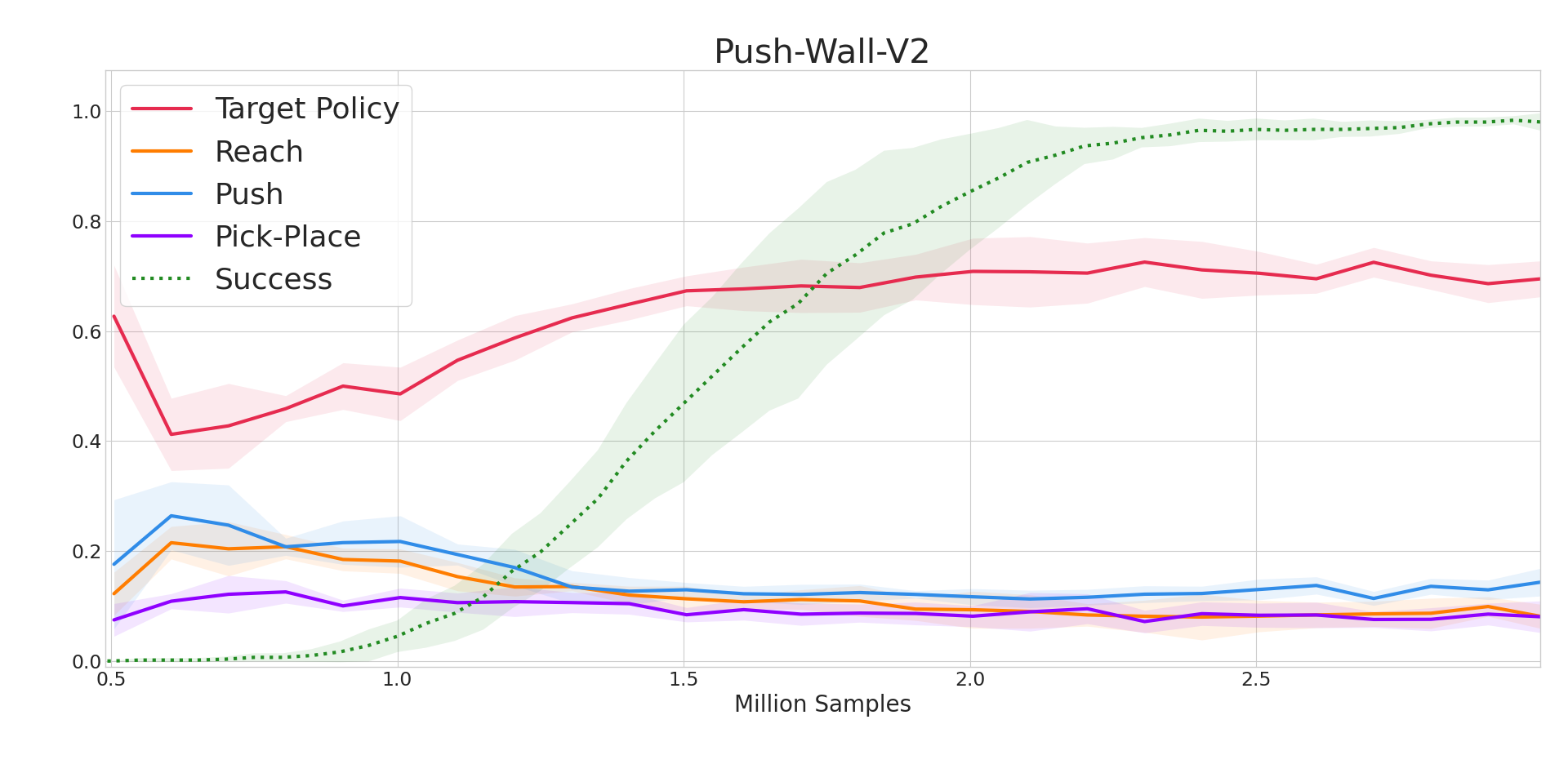}
\caption{Percentages of source policies being selected by CUP during training on Push-Wall. The green dashed line represents the target policy's success rate on the task.}  \label{pushwall1}
\end{figure}

\begin{figure}[tb]
\centering
        \includegraphics[width=0.8\columnwidth]{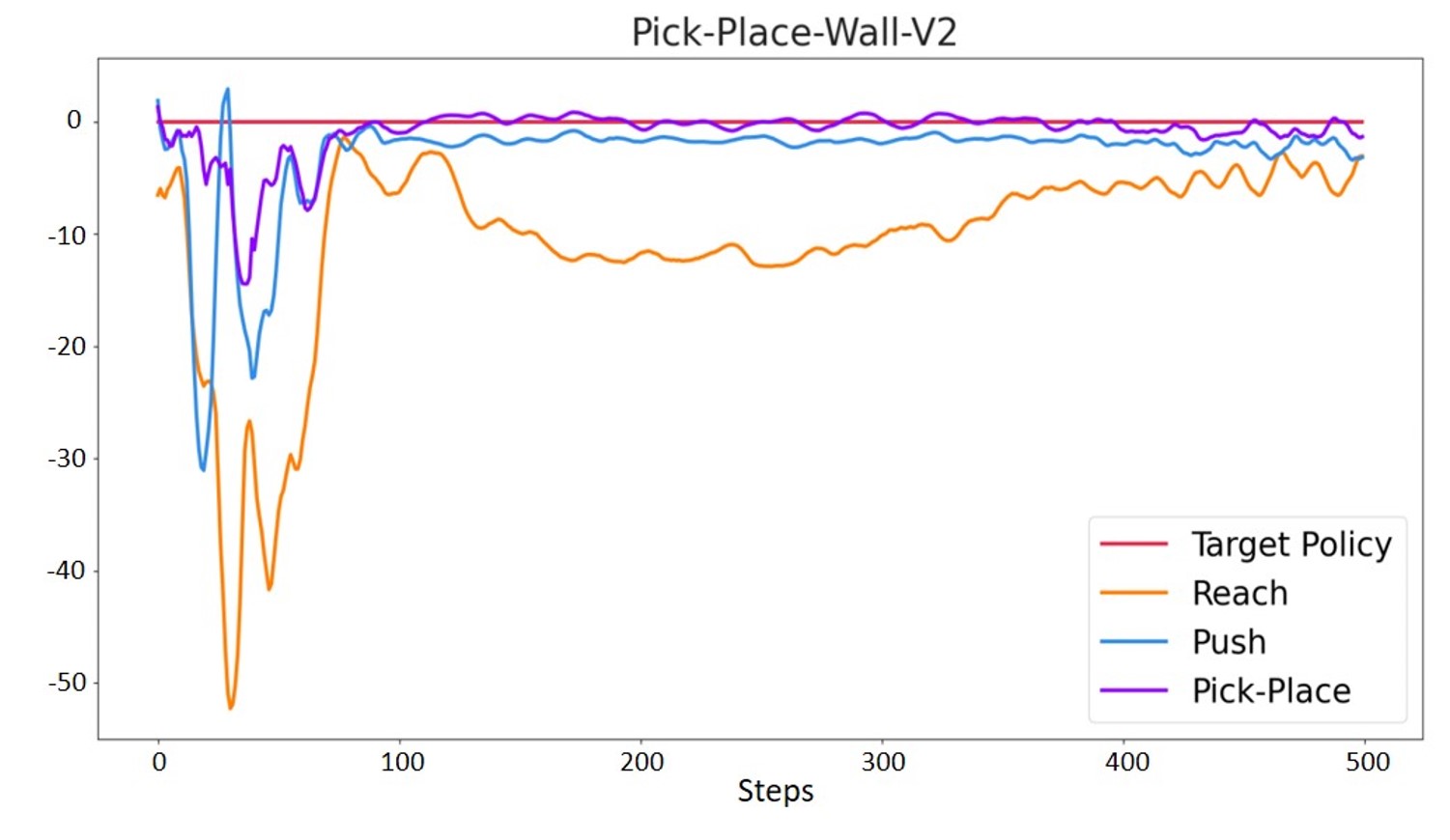}
\caption{Expected advantages of source policies at convergence on Pick-Place-Wall. The horizontal axis represents the environment steps of an episode.}
\label{eval}
\end{figure}

This subsection provides visualizations of CUP's source policy selection. Fig. \ref{pushwall1} shows the percentages of each source policy being selected throughout training on Push-Wall. At early stages of training, the source policies are selected more frequently as they have positive expected advantages, which means that they can be used to improve the current target policy. As training proceeds and the target policy becomes better, the source policies are selected less frequently. Among these three source policies, Push is chosen more frequently than the other two source policies, as it is more related to the target task. Figure \ref{eval} presents the source policies' expected advantages over an episode at convergence in Pick-Place-Wall. The Push source policy and Reach source policy almost always have negative expected advantages, which implies that these two source policies can hardly improve the current target policy anymore. Meanwhile, the Pick-Place source policy has expected advantages close to zero after 100 environment steps, which implies that the Pick-Place source policy is close to the target policy at these steps. Analyses on all six tasks as well as analyses on HAAR's source policy selection are deferred to Appendix \ref{agp} and Appendix \ref{anay}, respectively.

\subsection{Ablation Study}
\label{abl}
This subsection evaluates CUP's sensitivity to hyper-parameter settings and the number of source policies. We also evaluate CUP's robustness against random source policies, which do not provide meaningful candidate actions for solving target tasks.
\subsubsection{Hyper-Parameter Sensitivity}
\begin{figure}[h]
\centering
        \includegraphics[width=0.49\columnwidth]{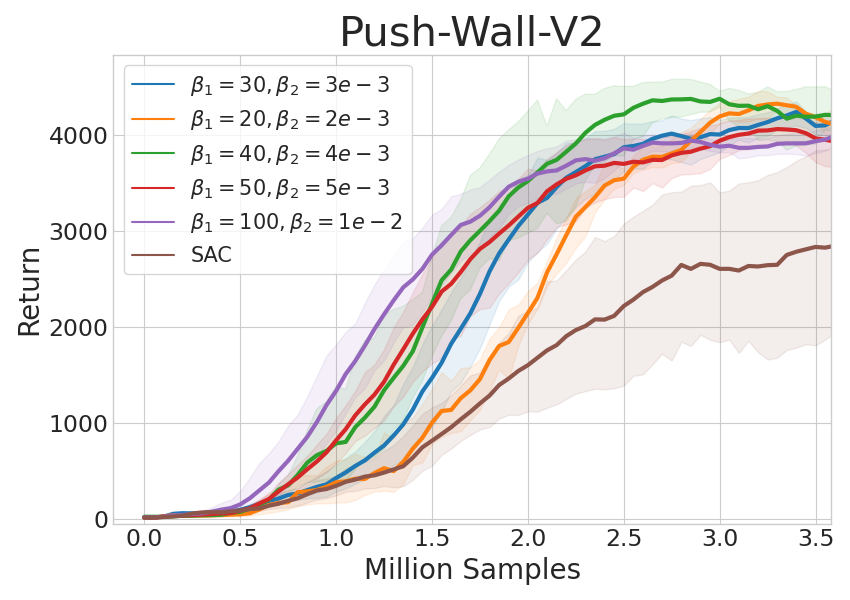}
        \includegraphics[width=0.49\columnwidth]{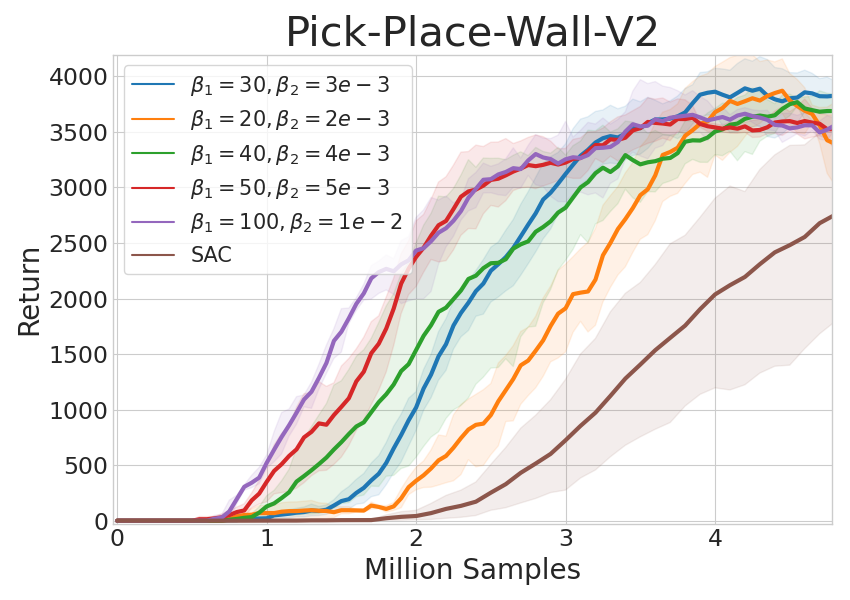}
\caption{Ablation studies on a wide range of hyper-parameters. CUP performs well on a wide range of hyper-parameters. }  \label{addtionalhyper}
\end{figure}

For all the experiments in Section \ref{tp}, we use the same set of hyper-parameters, which indicates that CUP is generally applicable to a wide range of tasks without particular fine-tuning. CUP introduces only two additional hyper-parameters to the underlying SAC algorithm, and we further test CUP's sensitivity to these additional hyper-parameters. As shown in Fig. \ref{addtionalhyper}, CUP is generally robust to the choice of hyper-parameters and achieves stable performance.

\subsubsection{Number of Source Policies}
\begin{figure}[t]
\centering
        \subfigure{\includegraphics[width=0.32\columnwidth]{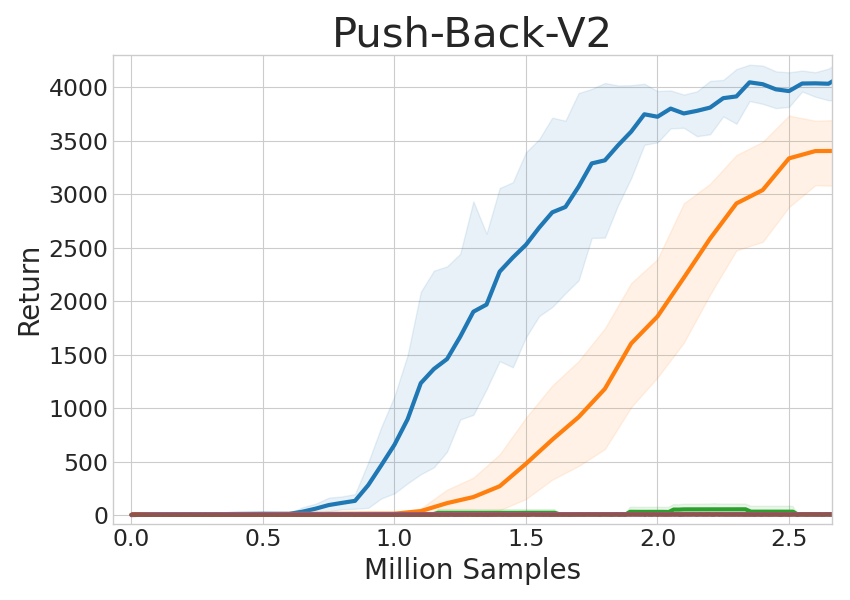}}
        \subfigure{\includegraphics[width=0.32\columnwidth]{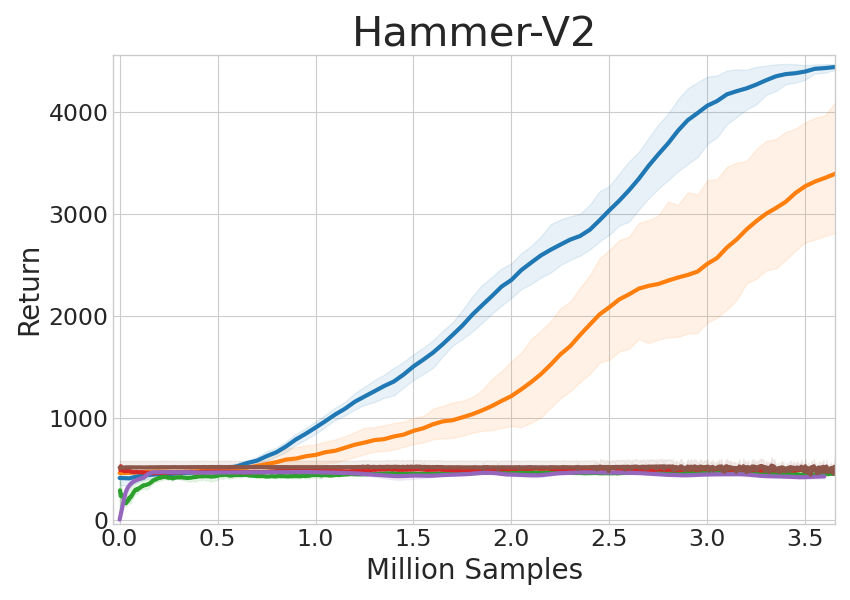}}
        \subfigure{\includegraphics[width=0.32\columnwidth]{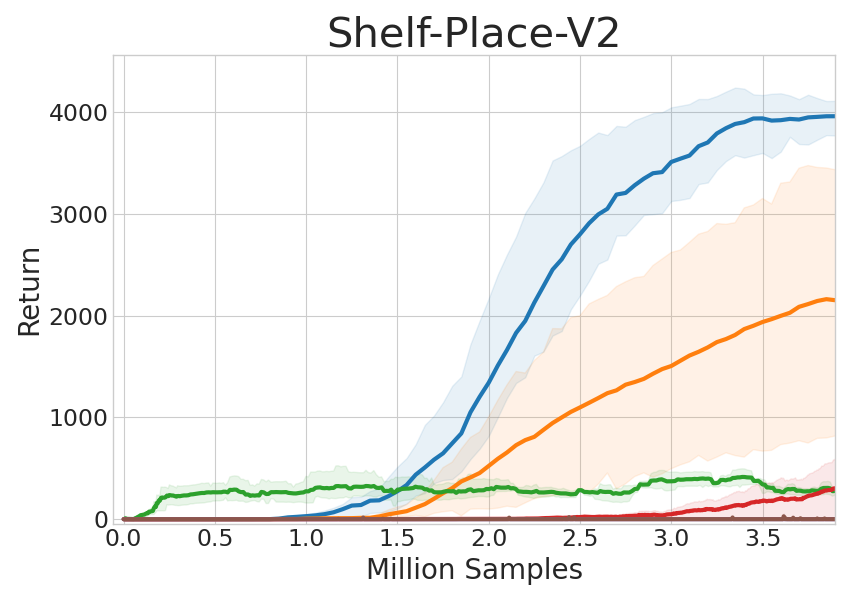}}
        \subfigure{\includegraphics[width=0.32\columnwidth]{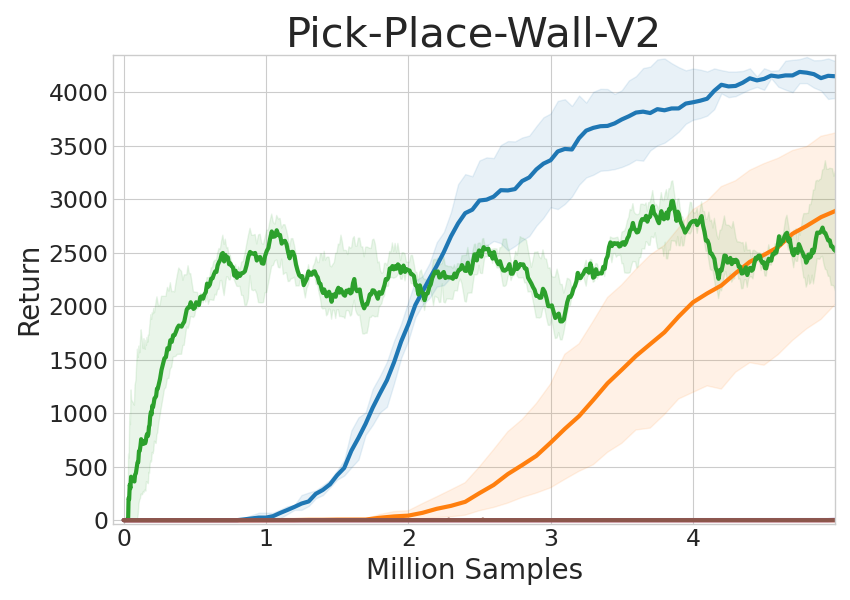}}
        \subfigure{\includegraphics[width=0.32\columnwidth]{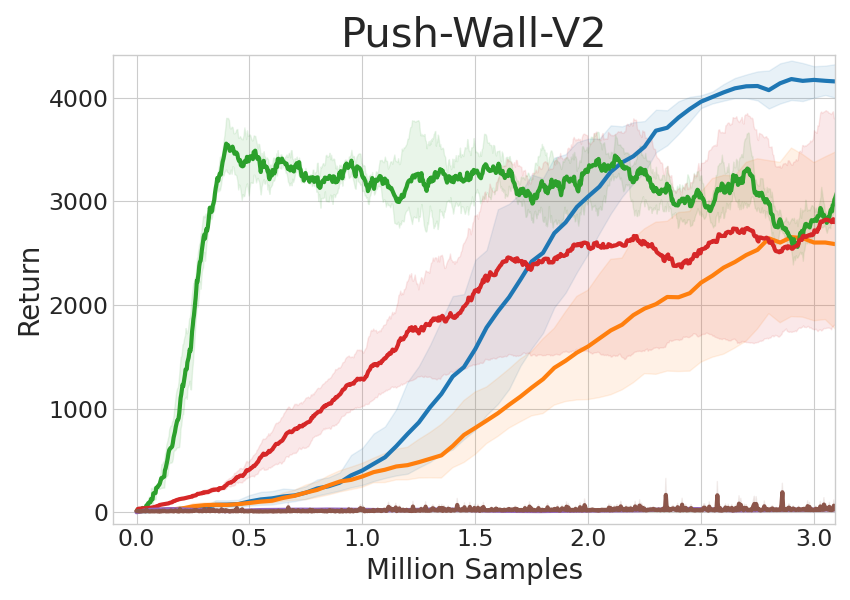}}
        \subfigure{\includegraphics[width=0.32\columnwidth]{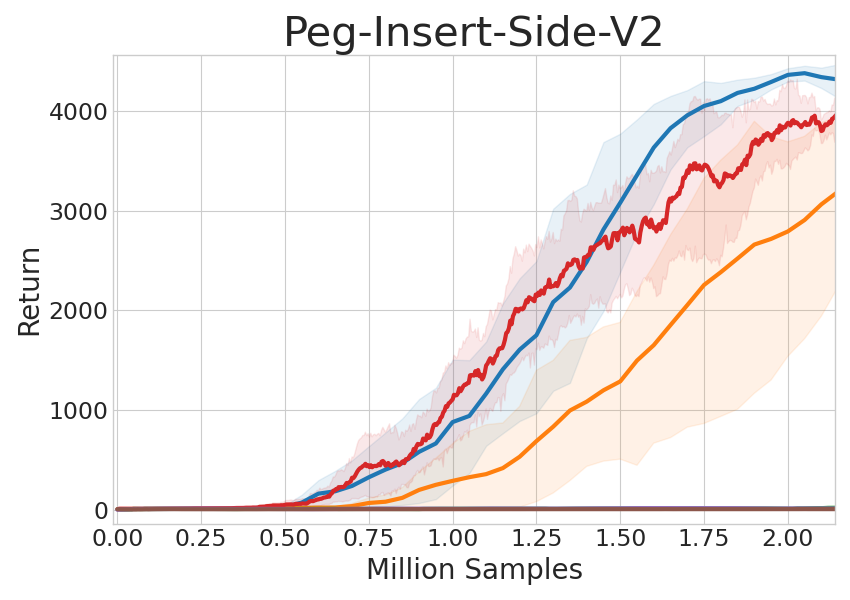}}
        \subfigure{\includegraphics[width=0.95\columnwidth]{figs/legends.png}}
\caption{Performance of CUP and baseline algorithms on various Meta-World tasks, with a set of six source policies.}
\label{curves2}
\end{figure}

\begin{figure}[t]
\centering
        \subfigure{\includegraphics[width=0.32\columnwidth]{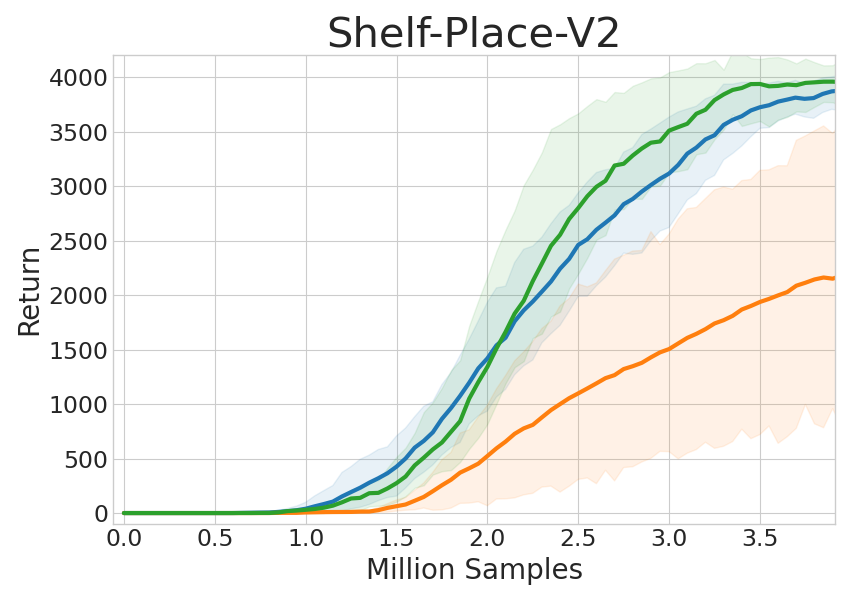}}
        \subfigure{\includegraphics[width=0.32\columnwidth]{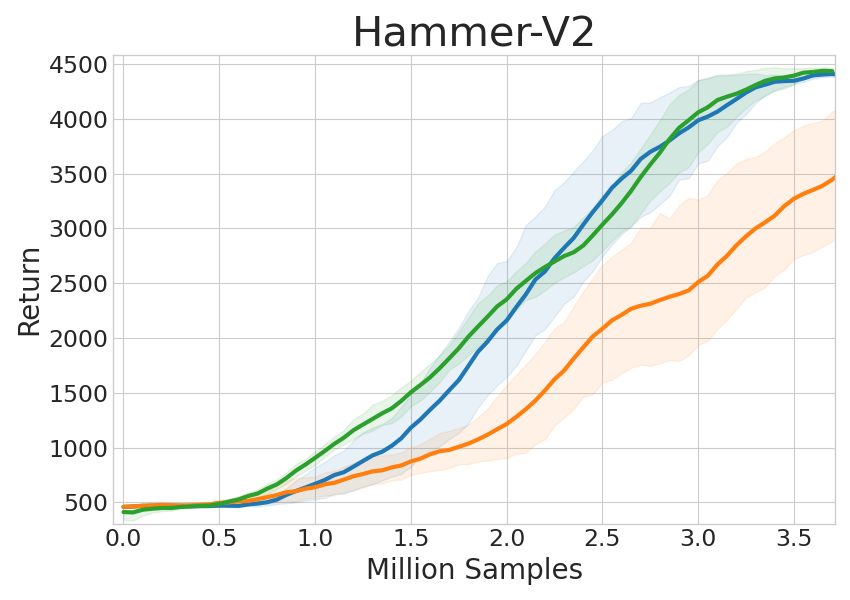}}
        \subfigure{\includegraphics[width=0.32\columnwidth]{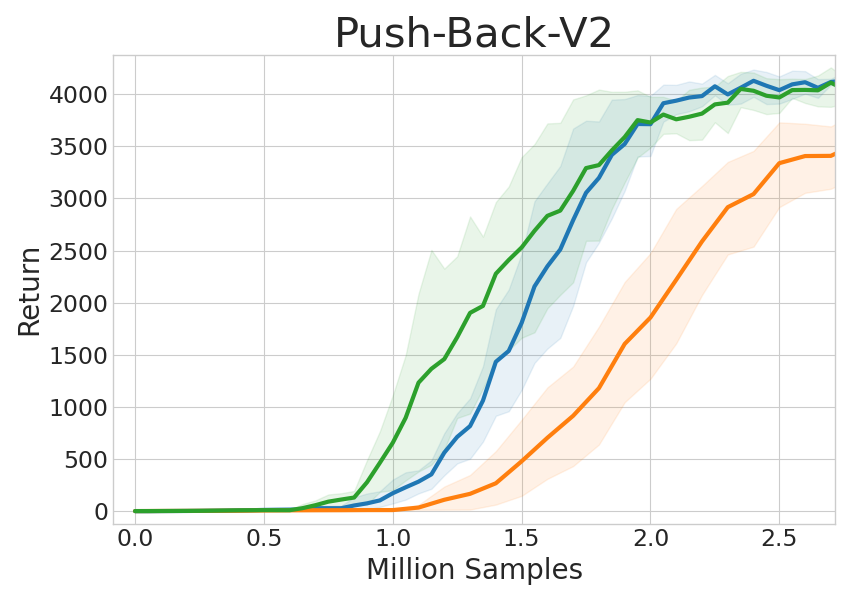}}
        \subfigure{\includegraphics[width=0.32\columnwidth]{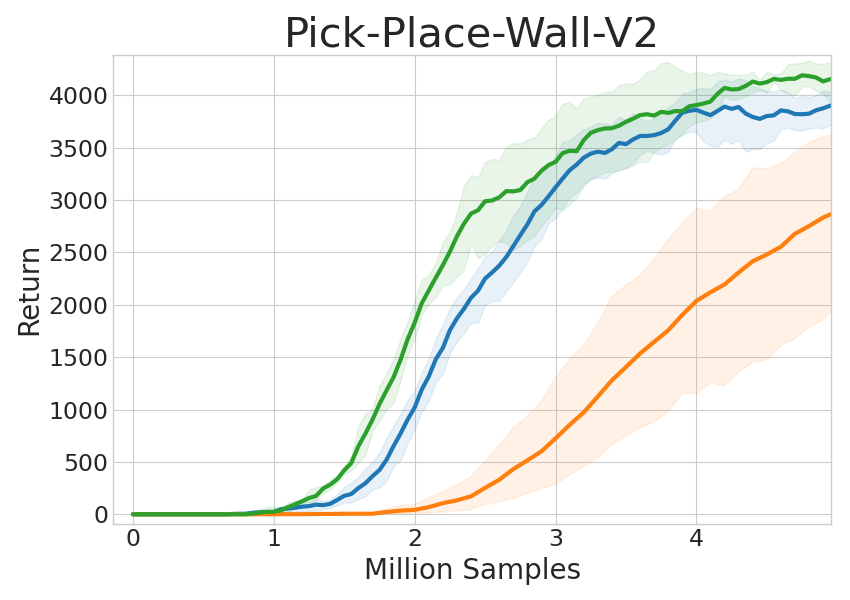}}
        \subfigure{\includegraphics[width=0.32\columnwidth]{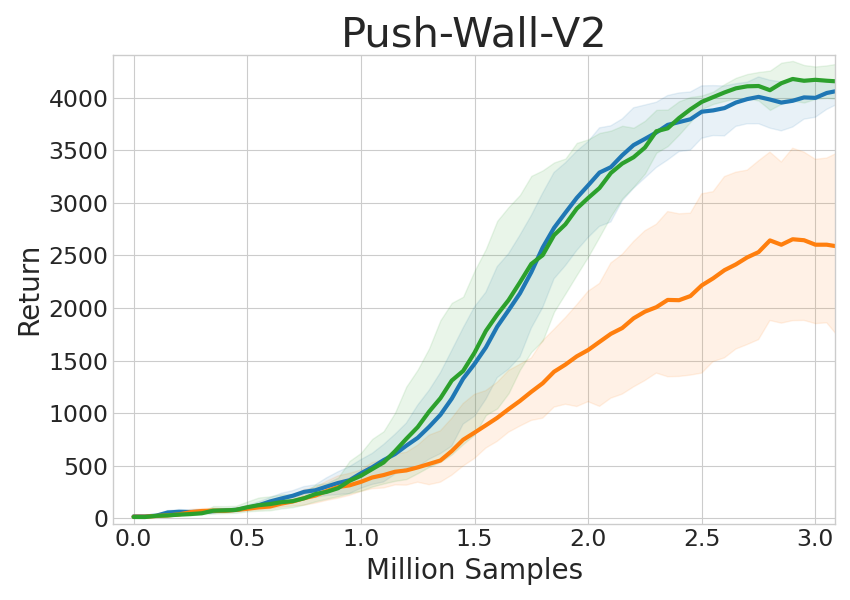}}
        \subfigure{\includegraphics[width=0.32\columnwidth]{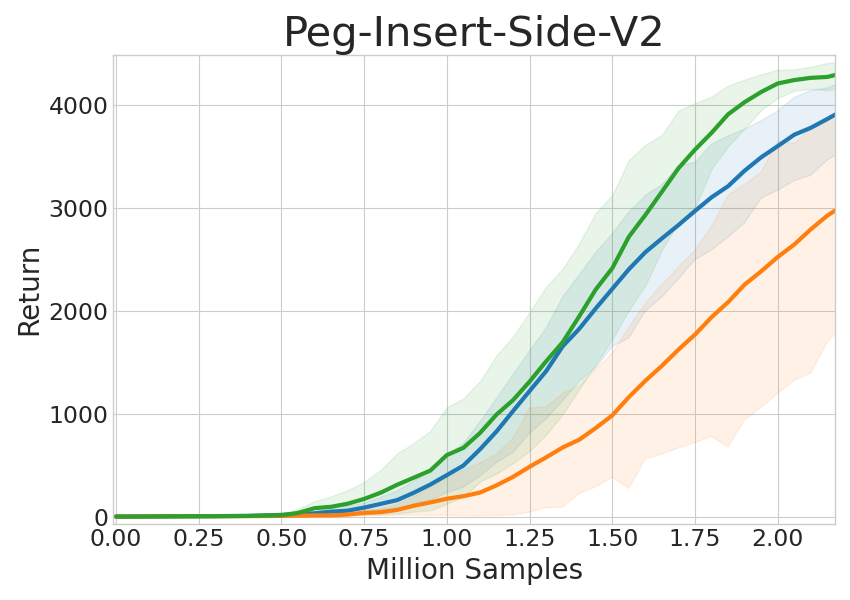}}
        \subfigure{\includegraphics[width=0.7\columnwidth]{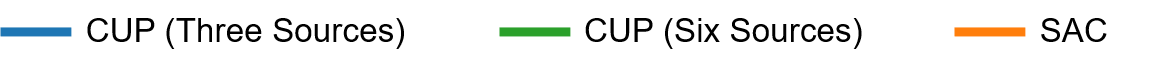}}
\caption{Comparison of CUP's performance with different number of source policies. }
\label{curves3}
\end{figure}

We evaluate CUP as well as baseline algorithms on a larger source policy set. We add three policies to the original source policy set, which solve three simple tasks including Drawer-Close, Push-Wall, and Coffee-Button. This forms a source policy set composed of six policies. As shown in Fig. \ref{curves2}, CUP is still the only algorithm that solves all the six target tasks efficiently. MULTIPOLAR suffers from a decrease in performance, which indicates that learning the weighted sum of source policies' actions becomes more difficult as the number of source policies grows. The rest of the baseline algorithms have similar performance to those using three source policies. Fig. \ref{curves3} provides a more direct comparison of CUP's performance with different number of source policies. CUP is able to utilize the additional source policies to further improve its performance, especially on Pick-Place-Wall and Peg-Insert-Side. Further detailed analysis is deferred to Appendix \ref{CAU}.
\subsubsection{Interference of Random Source Policies}
\begin{figure}[tb]
\centering
        \subfigure[]{\includegraphics[width=0.49\columnwidth]{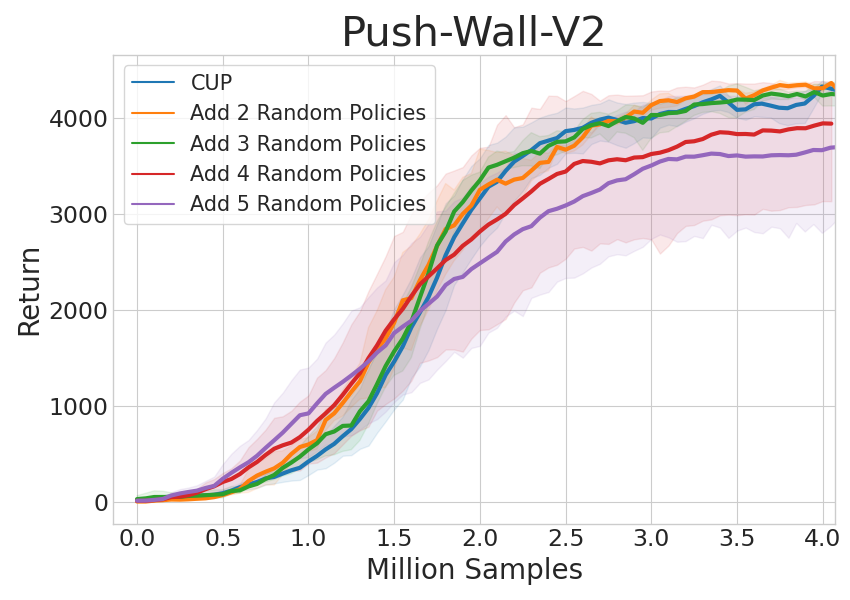}\label{useless}}
        \subfigure[]{\includegraphics[width=0.49\columnwidth]{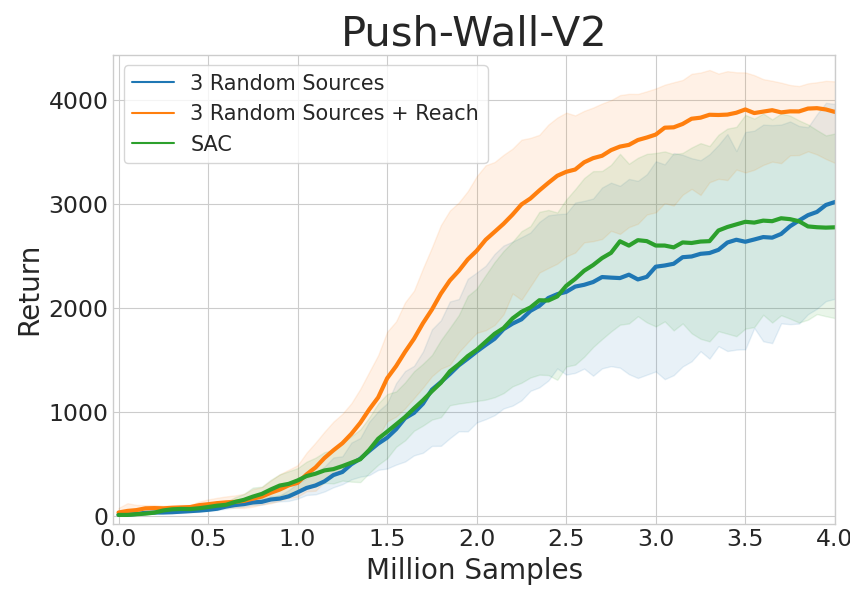}\label{useless2}}
\caption{Ablation studies on CUP's sensitivity to useless source policies. (a) Adding up to 3 random policies to the source policy set does not affect CUP's performance. (b) Ablation study in a setting where most source policies are useless. If none of the source policies are useful (3 Random Sources), CUP performs similarly to the original SAC. Even if only one of the four source policies is useful (3 Random Sources+Reach), CUP is still able to efficiently utilize the useful source policy to improve learning performance.}
\end{figure}

In order to evaluate the efficiency of CUP's critic-guided source policy aggregation, we add random policies to the set of source policies. As shown in Fig. \ref{useless}, adding up to 3 random source policies does not affect CUP's performance. This indicates that CUP can efficiently choose which source policy to follow even if there exist many source policies that are not meaningful. Adding 4 and 5 random source policies leads to a slight drop in performance. This drop is because that as the number of random policies grows, more random actions are sampled, and taking argmax over these actions' expected advantages is more likely to be affected by errors in value estimation.

To further investigate CUP's ability to ignore unsuitable source policies, we design another transfer setting that consists of another two source policy sets. The first set consists of three random policies that are useless for the target task, and the second set adds the Reach policy to the first set. As demonstrated in Fig. \ref{useless2}, when none of the source policies are useful, CUP performs similarly to the original SAC, and its sample efficiency is almost unaffected by the useless source policies. When there exists a useful source policy, CUP can efficiently utilize it to improve performance, even if there are many useless source policies.

\section{Related Work}
\paragraph{Policy reuse.}
A series of works on policy reuse utilize source policies for exploration in value-based algorithms \citep{fernandez2006probabilistic,li2018optimal,gimelfarb2021contextual}, but they are not applicable to policy gradient methods due to the off-policyness problem \citep{fujimoto2019off}. AC-Teach \citep{kurenkov2020ac} mitigates this problem by improving the actor over behavior policy's value estimation, but still fails in more complex tasks. One branch of methods train hierarchical high-level policies over source policies. CAPS \citep{li2018context} guarantees the optimality of the hierarchical policies by adding primitive skills to the low-level policy set, but is inapplicable to MDPs with continuous action spaces. HAAR \citep{li2019hierarchical} fine-tunes low-level policies to ensure optimality, but joint training of high-level and low-level policies induce optimization non-stationarity \citep{pateria2021hierarchical}. PTF \citep{yang2020efficient} trains a hierarchical policy,  which is imitated by the target policy. However, the hierarchical policy only gets updated when the target policy chooses similar actions to one of the source policies, so PTF fails in complex tasks with large action spaces. Another branch of works aggregate source policies via their Q functions or V functions on the target task. \cite{barreto2017successor} and \cite{barreto2018transfer} focus on the situation where source tasks and target tasks share the same dynamics, and aggregate source policies by choosing the policy that has the largest Q at each state. They use successor features to mitigate the heavy computation cost brought by estimating Q functions for all source policies. MAMBA \citep{cheng2020policy} forms a baseline function by aggregating source policies' V functions, and guides policy search by improving the policy over the baseline function. Finally, MULTIPOLAR \citep{barekatain2021multipolar} learns a weighted sum over source policies' actions, and learns an auxiliary network to predict residuals around the aggregated actions. MULTIPOLAR is computationally expensive, as it requires querying all the source policies at every sampling step. Our proposed method, CUP, focuses on the setting of learning continuous-action MDPs with actor-critic methods. CUP is both computationally and sampling efficient, as it does not require training any additional components.

\paragraph{Policy regularization.} Adding regularization to policy optimization is a common approach to induce prior knowledge into policy learning. Distral \citep{teh2017distral} achieves inter-task transfer by imitating an average policy distilled from policies of related tasks. In offline RL, policy regularization serves as a common technique to keep the policy close to the behavior policy used to collect the dataset \citep{wu2019behavior,nair2020awac,fujimoto2021minimalist}. CUP uses policy regularization as a means to provide additional guidance to policy search with the guidance policy.

\section{Conclusion}
In this study, we address the problem of reusing source policies without training any additional components. By utilizing the critic as a natural evaluation of source policies, we propose CUP, an efficient policy reuse algorithm without training any additional components. CUP is conceptually simple, easy to implement, and has theoretical guarantees. Empirical results demonstrate that CUP achieves efficient transfer on a wide range of tasks. As for future work, CUP assumes that all source policies and the target policy share the same state and action spaces, which limits CUP's application to more general scenarios. One possible future direction is to take inspiration from previous works that map the state and action spaces of an MDP to another MDP with similar high-level structure \citep{wan2020mutual,zhang2020learning,heng2022cross,van2020mdp,van2020plannable}. Another interesting direction is to incorporate CUP into the continual learning setting \citep{rolnick2019experience,khetarpal2020towards}, in which an agent gradually enriches its source policy set in an online manner.
\section*{Acknowledgements}
This work is supported in part by Science and Technology Innovation 2030 – “New Generation Artificial Intelligence” Major Project (No. 2018AAA0100904), National Natural Science Foundation of China (62176135), and China Academy of Launch Vehicle Technology (CALT2022-18).


\bibliography{example_paper}
\bibliographystyle{icml2021}

\section*{Checklist}


\begin{enumerate}

\item For all authors...
\begin{enumerate}
  \item Do the main claims made in the abstract and introduction accurately reflect the paper's contributions and scope?
    \answerYes{}
  \item Did you describe the limitations of your work?
    \answerYes{See Section \ref{pre}.}
  \item Did you discuss any potential negative societal impacts of your work?
    \answerYes{See Appendix \ref{psi}.}
  \item Have you read the ethics review guidelines and ensured that your paper conforms to them?
    \answerYes{}
\end{enumerate}

\item If you are including theoretical results...
\begin{enumerate}
  \item Did you state the full set of assumptions of all theoretical results?
    \answerYes{See Section \ref{3.1}.}
        \item Did you include complete proofs of all theoretical results?
    \answerYes{See Appendix \ref{proof1} and Appendix \ref{proof2}.}
\end{enumerate}

\item If you ran experiments...
\begin{enumerate}
  \item Did you include the code, data, and instructions needed to reproduce the main experimental results (either in the supplemental material or as a URL)?
    \answerYes{See the supplemental materials.}
  \item Did you specify all the training details (e.g., data splits, hyperparameters, how they were chosen)?
    \answerYes{See Appendix \ref{aid}.}
        \item Did you report error bars (e.g., with respect to the random seed after running experiments multiple times)?
    \answerYes{See Section \ref{exp}.}
        \item Did you include the total amount of compute and the type of resources used (e.g., type of GPUs, internal cluster, or cloud provider)?
    \answerYes{See Appendix \ref{aid}.}
\end{enumerate}

\item If you are using existing assets (e.g., code, data, models) or curating/releasing new assets...
\begin{enumerate}
  \item If your work uses existing assets, did you cite the creators?
    \answerYes{See Section \ref{exp}}
  \item Did you mention the license of the assets?
    \answerYes{See the supplemental materials.}
  \item Did you include any new assets either in the supplemental material or as a URL?
    \answerNA{}
  \item Did you discuss whether and how consent was obtained from people whose data you're using/curating?
    \answerNA{}
  \item Did you discuss whether the data you are using/curating contains personally identifiable information or offensive content?
    \answerNA{}
\end{enumerate}

\item If you used crowdsourcing or conducted research with human subjects...
\begin{enumerate}
  \item Did you include the full text of instructions given to participants and screenshots, if applicable?
    \answerNA{}
  \item Did you describe any potential participant risks, with links to Institutional Review Board (IRB) approvals, if applicable?
    \answerNA{}
  \item Did you include the estimated hourly wage paid to participants and the total amount spent on participant compensation?
    \answerNA{}
\end{enumerate}

\end{enumerate}

\appendix

\section{Broader Social Impact}
\label{psi}
We believe policy reuse serves as a promising way to transfer knowledge among AI agents. This ability will enable AI agents to master new skills efficiently. However, we are also aware of possible negative social impacts, such as plagiarizing other AI products by querying and reusing their policies.
\section{Proofs}
\subsection{Proof for Theorem \ref{thm2}}
\label{proof1}
\textit{Proof.} As $|\widetilde{Q}_{\pi_{tar}^t}(s,a)-{Q}_{\pi_{tar}^t}(s,a)|\leq \epsilon \text{\ for\  all}\  s \in \mathcal{S}, a \in A$, we have that for all $s_i \in \mathcal{S}$, the difference between the true value function $V_{\pi_{tar}^t}$ and the approximated value function $\widetilde{V}_{\pi_{tar}^t}$ is bounded:
\begin{align}
    & V_{\pi_{tar}^t}(s_i) \notag  \\
    &=\mathbb{E}_{a_i \sim \pi_{tar}^t(\cdot|s_i)}\left[Q_{\pi_{tar}^t}(s_{i},a_{i})-\alpha\log\pi_{tar}^{t}(a_i|s_i)\right] \notag \\
    & \leq \mathbb{E}_{a_i \sim \pi_{tar}^t(\cdot|s_i)}\left[\widetilde{Q}_{\pi_{tar}^t}(s_{i},a_{i})-\alpha\log\pi_{tar}^{t}(a_i|s_i)+\epsilon\right]  \notag \\
    & = \widetilde{V}_{\pi_{tar}^t}(s_i)+ \epsilon. \notag
\end{align}

\ As $\pi_{tar}^{t}(\cdot|s)$ is contained in $\Pi_t^s$, with $\widetilde{\pi_{g}^t}$ defined in Eq. \eqref{approximatepi}, it is obvious that for all $s \in \mathcal{S}$, $\mathbb{E}_{a \sim \widetilde{\pi_g^t} (\cdot|s)}\left[\widetilde{Q}_{\pi_{tar}^{t}}(s,a)-\alpha\log\widetilde{\pi_g^t}(a|s)\right] \geq \mathbb{E}_{a \sim \pi_{tar}^t (\cdot|s)}\left[\widetilde{Q}_{\pi_{tar}^{t}}(s,a)-\alpha\log\pi_{tar}^{t}(a|s)\right]=\widetilde{V}_{\pi_{tar}^{t}}(s)$. Then for all $s_i \in \mathcal{S}$,
\begin{align}
    & V_{\pi_{tar}^t}(s_i) \notag  \\
    & \leq \widetilde{V}_{\pi_{tar}^t}(s_i)+ \epsilon \notag \\
    & \leq \mathbb{E}_{a_i \sim \widetilde{\pi_{g}^t}(\cdot|s_i)}[\widetilde{Q}_{\pi_{tar}^{t}}(s_i,a_i)-\alpha\log\widetilde{\pi_g^t}(a_i|s_i)] + \epsilon  \notag \\
    & \leq \mathbb{E}_{a_i \sim \widetilde{\pi_{g}^t}(\cdot|s_i)}[{Q}_{\pi_{tar}^{t}}(s_i,a_i)-\alpha\log\widetilde{\pi_g^t}(a_i|s_i)]+ 2\epsilon  \notag \\
        &=\mathbb{E}_{a_i \sim \pi_{g}^t(a|s_i)}[r(s_i,a_i)-\alpha\log\widetilde{\pi_g^t}(a_i|s_i) +\gamma V_{\pi_{tar}^{t}}(s_{i+1})]+ 2\epsilon \notag \\
     & \vdots  \notag \\
     & \leq \mathbb{E}_{\widetilde{\pi_{g}^t}}[\sum_{\tau=0}^{\infty}\gamma^{\tau}(r(s_{i+\tau},a_{i+\tau})-\alpha\log\widetilde{\pi_g^t}(a_{i+\tau}|s_{i+\tau}))] + 2 \sum_{\tau=0}^{\infty}\gamma^{\tau}\epsilon \notag \\
     & =V_{\widetilde{\pi_g^t}} (s_i)+\frac{2\epsilon}{1-\gamma}. \notag
\end{align}
\subsection{Proof of Theorem \ref{thm3}}
\label{proof2}
\textit{Proof.} According to Pinsker’s inequality \citep{fedotov2003refinements}, $D_{KL}(\pi_{tar}^{t+1}(\cdot|s)||\widetilde{\pi_{g}^t}(\cdot|s))\geq \frac{1}{2\ln2}||\pi_{tar}^{t+1}(\cdot|s)-\widetilde{\pi_{g}^t}(\cdot|s)||_1^2$, where $||\cdot||_1$ is the L1 norm. So we have that for all $s \in \mathcal{S}$, $||\pi_{tar}^{t+1}(\cdot|s)-\widetilde{\pi_{g}^t}(\cdot|s)||_1 \leq \sqrt{2\ln2 \delta}$. According to the Performance Difference Lemma \citep{kakade2002approximately}, we have that for all $s \in \mathcal{S}$:
\begin{align}
    &V_{\widetilde{\pi_{g}^t}}(s)-V_{\pi_{tar}^{t+1}}(s) \notag  \\
    & = \frac{1}{1-\gamma}\mathbb{E}_{s' \sim \mu_{s}^{\widetilde{\pi_{g}^t}}(s')} \notag \\
    &\ \ \ \ \ \ \ \ \ \left[\mathbb{E}_{a \sim \widetilde{\pi_{g}^t}(\cdot|s')}[Q_{\pi_{tar}^{t+1}}(s',a)-\alpha\log\widetilde{\pi_{g}^t}(a|s)]-\mathbb{E}_{a \sim \widetilde{\pi_{tar}^{t+1}}(\cdot|s')}[Q_{\pi_{tar}^{t+1}}(s',a)-\alpha\log\widetilde{\pi_{tar}^{t+1}}(a|s)]\right] \notag \\
    & \leq \frac{1}{1-\gamma}\max\limits_{s' \in \mathcal{S}}\left[\mathbb{E}_{a \sim \widetilde{\pi_{g}^t}(\cdot|s')}[Q_{\pi_{tar}^{t+1}}(s',a)]-\mathbb{E}_{a \sim \pi_{tar}^{t+1}(\cdot|s')}[Q_{\pi_{tar}^{t+1}}(s',a)]\right] \notag \\
    & \ \ \ \ \ \ \ \ \ \ \ \ \ \ \ \ \ \ \ \ \ \ \ \ \ \ \ \ \ \ \ \ \ \ \ \ \ \ \ \ \ \ \ \ \ \ \ \ \ \ \ \ \ \ \ \ \ \ \ \ \ \ \ \ \ \ \ \ \ \ \ \ \ \ \ \ \ \ \ \ \ \ \ +\frac{\alpha}{1-\gamma}\max\limits_{s'' \in \mathcal{S}}\left|\mathcal{H}(\widetilde{\pi_{g}^t}(\cdot|s''))-\mathcal{H}(\pi_{tar}^t(\cdot|s''))\right| \notag \\
    & = \frac{1}{1-\gamma}\max\limits_{s' \in \mathcal{S}}\int\left(\widetilde{\pi_{g}^t}(\cdot|s)-\pi_{tar}^{t+1}(a|s)\right)Q_{\pi_{tar}^{t+1}}(s',a)da+\frac{\alpha}{1-\gamma}\widetilde{\mathcal{H}}_{max} \notag \\
    & \leq\frac{1}{1-\gamma} \max\limits_{s' \in \mathcal{S}}\int\left|\widetilde{\pi_{g}^t}(a|s)-\pi_{tar}^{t+1}(a|s)\right|\cdot\left|Q_{\pi_{tar}^{t+1}}(s',a)\right|da+\frac{\alpha}{1-\gamma}\widetilde{\mathcal{H}}_{max} \notag \\
    &\leq \frac{1}{1-\gamma} \max\limits_{s' \in \mathcal{S}}\int\left|\widetilde{\pi_{g}^t}(a|s)-\pi_{tar}^{t+1}(a|s)\right|\cdot\frac{\widetilde{R}_{max}+\alpha\mathcal{H}_{max}^{t+1}}{1-\gamma}da+\frac{\alpha}{1-\gamma}\widetilde{\mathcal{H}}_{max} \notag \\
    & = \frac{\widetilde{R}_{max}+\alpha\mathcal{H}_{max}^{t+1}}{(1-\gamma)^2}\max\limits_{s' \in \mathcal{S}}||\widetilde{\pi_{g}^t}(\cdot|s)-\pi_{tar}^{t+1}(\cdot|s)||_1+\frac{\alpha}{1-\gamma}\widetilde{\mathcal{H}}_{max} \notag \\
    & \leq \frac{\sqrt{2\ln2\delta}(\widetilde{R}_{max}+\alpha\mathcal{H}_{max}^{t+1})+\alpha(1-\gamma)\widetilde{\mathcal{H}}_{max}}{(1-\gamma)^2}, \notag \\
\end{align}
\ where $\mu_{s}^{\widetilde{\pi_{g}^t}}(s')=(1-\gamma)\sum_{t=0}^{\infty}\gamma^tp(s_t=s'|s_0=s,\widetilde{\pi_{g}^t})$\footnote{We slightly abuse the notation $s_0$ here to indicate that the agent start deterministically from state $s$.} is the normalized discounted state occupancy distribution. Note that
\begin{align}
&|Q_{\pi_{tar}^{t+1}}(s,a)|\notag \\
&=\left|\mathbb{E}_{\pi_{tar}^{t+1}}\left[\sum_{i=0}^{\infty}\gamma^i(r(s_{\tau+i},a_{\tau+i})-\alpha\log\pi_{tar}^{t+1}(\cdot|s))|s_\tau=s,a_\tau=a\right]\right| \notag \\
&\leq\mathbb{E}_{\pi}[\sum_{i=0}^{\infty}\gamma^i(\widetilde{R}_{max}+\gamma\mathcal{H}_{max}^{t+1})] \\
&=\frac{\widetilde{R}_{max}+\alpha\mathcal{H}_{max}^{t+1}}{1-\gamma}.
\end{align}
\ Eventually, we have $V_{\pi_{tar}^{t+1}}(s) \geq V_{\widetilde{\pi_{g}^t}}(s) - \frac{\sqrt{2\ln2\delta}(\widetilde{R}_{max}+\alpha\mathcal{H}_{max}^{t+1})+\alpha(1-\gamma)\widetilde{\mathcal{H}}_{max}}{(1-\gamma)^2} \geq V_{\pi_{tar}^t}(s)-\frac{\sqrt{2\ln2\delta}(\widetilde{R}_{max}+\alpha\mathcal{H}_{max}^{t+1})}{(1-\gamma)^2}- \frac{2\epsilon+\alpha\widetilde{\mathcal{H}}_{max}}{1-\gamma}$.
\subsection{Critic-Guided Source Policy Aggregation under ``Hard'' Value Functions}
\label{hard}
In this section we override the notation $Q$, $V$ to represent ``hard'' value functions, and override the notation $EA$ to represent the \textit{expected advantage}, which is defined as  $EA_{\pi_j}(s,\pi_i)=\mathbb{E}_{a \sim \pi_i(\cdot|s)}\left[Q_{\pi_j}(s,a)-V_{\pi_j}(s)\right]$. Then Theorem \ref{thm2} and Theorem \ref{thm3} can be extended as below.
\begin{thm}
Let $\widetilde{Q}_{\pi_{tar}^t}$ be an approximation of ${Q}_{\pi_{tar}^t}$ such that 
\begin{equation}
    |\widetilde{Q}_{\pi_{tar}^t}(s,a)-{Q}_{\pi_{tar}^t}(s,a)|\leq \epsilon \text{\ for\  all}\  s \in \mathcal{S}, a \in A.
\end{equation}
Define 
\begin{equation}
   \widetilde{\pi_g^t}(\cdot |s)=\mathop{\arg\max}\limits_{\pi(\cdot |s) \in \Pi_{t}^s}\mathbb{E}_{a \sim \pi(\cdot |s)}\left[\widetilde{Q}_{\pi_{tar}^t}(s,a)\right] \text{\ for\  each\ } s \in \mathcal{S}.
   \label{approximatepi2}
\end{equation}
Then, 
\begin{equation}
   V_{\widetilde{\pi_g^t}}(s)\geq V_{\pi_{tar}^t}(s) - \frac{2\epsilon}{1-\gamma} \text{\ for\  all\ } s \in \mathcal{S}.
\end{equation}
\label{thm22}
\end{thm}
\begin{thm}
If 
\begin{equation}
   D_{KL}\left(\pi_{tar}^{t+1}(\cdot|s)||\widetilde{\pi_g^t}(\cdot|s)\right)\leq \delta\  for\  all\  s \in \mathcal{S},
\end{equation}
\ then 
\begin{equation}
   V_{\pi_{tar}^{t+1}}(s) \geq V_{\pi_{tar}^t}(s)-\frac{\sqrt{2\ln{2}\delta}\widetilde{R}_{max}}{(1-\gamma)^2}- \frac{2\epsilon}{1-\gamma}\  for\  all\  s \in \mathcal{S},
\end{equation}
\ where $\widetilde{R}_{max}=\max\limits_{s,a}|r(s,a)|$ is the largest possible absolute value of the reward.
\label{thm33}
\end{thm}
Theorem \ref{thm22} and Theorem \ref{thm33} implies that CUP can still guarantee policy improvement under hard Bellman updates. Proofs are given below.

\paragraph{Proof for Theorem \ref{thm22}.} As $|\widetilde{Q}_{\pi_{tar}^t}(s,a)-{Q}_{\pi_{tar}^t}(s,a)|\leq \epsilon \text{\ for\  all}\  s \in \mathcal{S}, a \in A$, we have that for all $s_i \in \mathcal{S}$, the difference between the true value function $V_{\pi_{tar}^t}$ and the approximated value function $\widetilde{V}_{\pi_{tar}^t}$ is bounded:
\begin{align}
    & V_{\pi_{tar}^t}(s_i) \notag  \\
    &=\mathbb{E}_{a \sim \pi_{tar}^t(\cdot|s)}\left[Q_{\pi_{tar}^t}(s_{i},a_{i})\right] \notag \\
    & \leq \mathbb{E}_{a \sim \pi_{tar}^t(\cdot|s)}\left[\widetilde{Q}_{\pi_{tar}^t}(s_{i},a_{i})+\epsilon\right]  \notag \\
    & = \widetilde{V}_{\pi_{tar}^t}(s_i)+ \epsilon. \notag
\end{align}

As $\pi_{tar}^{t}(\cdot|s)$ is contained in $\Pi_t^s$, with $\pi_g^t$ defined in Eq. \eqref{approximatepi2}, it is obvious that for all $s \in \mathcal{S}$, $\mathbb{E}_{a \sim \widetilde{\pi_g^t} (\cdot|s)}\left[\widetilde{Q}_{\pi_{tar}^{t}}(s,a)\right] \geq \mathbb{E}_{a \sim \pi_{tar}^t (\cdot|s)}\left[\widetilde{Q}_{\pi_{tar}^{t}}(s,a)\right]=\widetilde{V}_{\pi_{tar}^{t}}(s)$. Then for all $s_i \in \mathcal{S}$,
\begin{align}
    & V_{\pi_{tar}^t}(s_i) \notag  \\
    & \leq \widetilde{V}_{\pi_{tar}^t}(s_i)+ \epsilon \notag \\
    & \leq \mathbb{E}_{a_i \sim \widetilde{\pi_{g}^t}(\cdot|s_i)}[\widetilde{Q}_{\pi_{tar}^{t}}(s_i,a_i)] + \epsilon  \notag \\
    & \leq \mathbb{E}_{a_i \sim \widetilde{\pi_{g}^t}(\cdot|s_i)}[{Q}_{\pi_{tar}^{t}}(s_i,a_i)]+ 2\epsilon  \notag \\
        &=\mathbb{E}_{a_i \sim \pi_{g}^t(a|s_i)}[r(s_i,a_i) +\gamma V_{\pi_{tar}^{t}}(s_{i+1})]+ 2\epsilon \notag \\
    & \leq  \mathbb{E}_{a_i \sim \widetilde{\pi_{g}^t}(\cdot|s_i)}[r(s_i,a_i)+\gamma( \widetilde{V}_{\pi_{tar}^{t}}(s_{i+1})+\epsilon)]+2\epsilon \notag \\
    &\leq \mathbb{E}_{a_i \sim \widetilde{\pi_{g}^t}(\cdot|s_i),a_{i+1} \sim  \widetilde{\pi_{g}^t}(\cdot|s_{i+1})}[r(s_i,a_i)+\gamma \widetilde{Q}_{\pi_{tar}^{t}}(s_{i+1},a_{i+1})]+(2+\gamma)\epsilon  \notag \\
    &\leq \mathbb{E}_{a_i \sim \widetilde{\pi_{g}^t}(\cdot|s_i),a_{i+1} \sim  \widetilde{\pi_{g}^t}(\cdot|s_{i+1})}[r(s_i,a_i)+\gamma( {Q}_{\pi_{tar}^{t}}(s_{i+1},a_{i+1})+\epsilon)]+(2+\gamma)\epsilon  \notag \\
    & = \mathbb{E}_{a_i \sim \widetilde{\pi_{g}^t}(\cdot|s_i),a_{i+1} \sim  \widetilde{\pi_{g}^t}(\cdot|s_{i+1})}[r(s_i,a_i)+\gamma r(s_{i+1},a_{i+1})+\gamma^2 V_{\pi_{tar}^{t}}(s_{i+2})]+(2+2\gamma)\epsilon \notag \\
     & \vdots  \notag \\
     & \leq \mathbb{E}_{\widetilde{\pi_{g}^t}}[\sum_{\tau=0}^{\infty}\gamma^{\tau}r(s_{i+\tau},a_{i+\tau})] + 2 \sum_{\tau=0}^{\infty}\gamma^{\tau}\epsilon \notag \\
     & =V_{\widetilde{\pi_g^t}} (s_i)+\frac{2\epsilon}{1-\gamma}. \notag
\end{align}

\paragraph{Proof for Theorem \ref{thm33}.} According to Pinsker’s inequality \citep{fedotov2003refinements}, $D_{KL}(\pi_{tar}^{t+1}(\cdot|s)||\widetilde{\pi_{g}^t}(\cdot|s))\geq \frac{1}{2\ln2}||\pi_{tar}^{t+1}(\cdot|s)-\widetilde{\pi_{g}^t}(\cdot|s)||_1^2$, where $||\cdot||_1$ is the L1 norm. So we have that for all $s \in \mathcal{S}$, $||\pi_{tar}^{t+1}(\cdot|s)-\widetilde{\pi_{g}^t}(\cdot|s)||_1 \leq \sqrt{2\ln2 \delta}$. According to the Performance Difference Lemma \citep{kakade2002approximately}, we have that for all $s \in \mathcal{S}$:
\begin{align}
    &V_{\widetilde{\pi_{g}^t}}(s)-V_{\pi_{tar}^{t+1}}(s) \notag  \\
    & = \frac{1}{1-\gamma}\mathbb{E}_{s' \sim \mu_{s}^{\widetilde{\pi_{g}^t}}(s')}\left[EA_{\pi_{tar}^{t+1}}(s,\widetilde{\pi_{g}^t})\right] \notag \\
    & \leq \frac{1}{1-\gamma}\max\limits_{s' \in \mathcal{S}}\left[\mathbb{E}_{a \sim \widetilde{\pi_{g}^t}(\cdot|s')}[Q_{\pi_{tar}^{t+1}}(s',a)]-\mathbb{E}_{a \sim \pi_{tar}^{t+1}(\cdot|s')}[Q_{\pi_{tar}^{t+1}}(s',a)]\right] \notag \\
    & = \frac{1}{1-\gamma}\max\limits_{s' \in \mathcal{S}}\int\left(\widetilde{\pi_{g}^t}(\cdot|s)-\pi_{tar}^{t+1}(a|s)\right)Q_{\pi_{tar}^{t+1}}(s',a)da \notag \\
    & \leq\frac{1}{1-\gamma} \max\limits_{s' \in \mathcal{S}}\int\left|\widetilde{\pi_{g}^t}(a|s)-\pi_{tar}^{t+1}(a|s)\right|\cdot\left|Q_{\pi_{tar}^{t+1}}(s',a)\right|da \notag \\
    &\leq \frac{1}{1-\gamma} \max\limits_{s' \in \mathcal{S}}\int\left|\widetilde{\pi_{g}^t}(a|s)-\pi_{tar}^{t+1}(a|s)\right|\cdot\frac{\widetilde{R}_{max}}{1-\gamma}da \notag \\
    & = \frac{\widetilde{R}_{max}}{(1-\gamma)^2}\max\limits_{s' \in \mathcal{S}}||\widetilde{\pi_{g}^t}(\cdot|s)-\pi_{tar}^{t+1}(\cdot|s)||_1 \notag \\
    & \leq \frac{\sqrt{2\ln2\delta}\widetilde{R}_{max}}{(1-\gamma)^2}. \notag \\
\end{align}
Note that $|Q_{\pi}(s,a)|=|\mathbb{E}_{\pi}[\sum_{i=0}^{\infty}\gamma^tr(s_{t+i},a_{t+i})|s_t=s,a_t=a]|\leq\mathbb{E}_{\pi}[\sum_{t=0}^{\infty}\gamma^t\widetilde{R}_{max}]=\frac{\widetilde{R}_{max}}{1-\gamma}$. Eventually, we have $V_{\pi_{tar}^{t+1}}(s) \geq V_{\widetilde{\pi_{g}^t}}(s) - \frac{\sqrt{2\ln2\delta}\widetilde{R}_{max}}{(1-\gamma)^2} \geq V_{\pi_{tar}^t}(s)-\frac{\sqrt{2\ln{2}\delta}\widetilde{R}_{max}}{(1-\gamma)^2}- \frac{2\epsilon}{1-\gamma}$.

\section{Discussion on the Influence of Over-Estimation}
As CUP takes an argmax over expected Q values, it may suffer from the value over-estimation issue in DRL \citep{ostrovski2021difficulty}. Although CUP may over-estimate values on rarely selected actions, this over-estimation serves as a kind of exploration mechanism, encouraging the agent to explore actions suggested by the source policies and potentially improving the learning target policy. If the source policies give unsuitable actions, then after exploration this over-estimation is resolved and these unsuitable actions will not be selected again. Results in Figure \ref{useless2} suggest that even if all source policies are random and do not give useful actions, CUP still performs similarly to the original SAC, and is almost unaffected by the over-estimation issue, as over-estimation is addressed after exploring these actions.

\section{Experimental Settings}
\subsection{Additional Implementation Details}
\label{aid}
To improve CUP's computation efficiency, we store the source polices' output $\{\pi_1(\cdot|s),\pi_2(\cdot|s),...,\pi_n(\cdot|s)\}$ in the replay buffer. As we can query source policies with batches of states, and each state in the buffer only need to be queried for once, CUP is computationally efficient. Empirically, CUP only takes about 30\% more wall-clock time than SAC to run the same number of environment steps. All experiments are run on GeForce GTX 2080 GPUs. The policy regularization is added after 0.5M environment environment steps to achieve more stable learning.

SAC utilizes two Q functions to mitigate the overestimation error. When CUP forms the guidance policy, we use the max value of the two target Q functions to estimate the expected advantage, which contributes to bolder exploration. Using target networks contributes to more stable training.

Equation \ref{arrgrgate} requires estimating expectations over Q values. In practice, to be efficient, we estimate the expectation by sampling a few actions (e.g., 3 actions) from each action probability distribution proposed by the source policies, and find it sufficient to achieve stable performance.

As for HAAR, we fix the source policies, and train a high-level policy as well as an additional low-level policy with HAAR's auxiliary rewards.
\subsection{Hyper-Parameter Details}
All hyper-parameters used in our experiments are listed in Table \ref{sample-table}. We use the same set of hyper-parameters for all six tasks. We also use the same set of hyper-parameters for both CUP and the SAC baseline. Most hyper-parameters are adopted from \cite{sodhani2021multi}.

\begin{table}[]
\caption{Detailed hyper-parameter settings for CUP.}
  \label{sample-table}
\begin{tabular}{c|c}
\hline
    \multicolumn{1}{l}{Hyper-Parameter}                                                                                          &\multicolumn{1}{l}{Hyper-Parameter Values}                                                                                                            \\ \hline
\multicolumn{1}{l}{batch size }    & \multicolumn{1}{l}{ 1280 }                                                                                                                                  \\
\multicolumn{1}{l}{non-linearity}                                                                                      & \multicolumn{1}{l}{ReLU}                                                                                                             \\
\multicolumn{1}{l}{actor/critic network structure}                                                                     & \multicolumn{1}{l}{\begin{tabular}[c]{@{}l@{}}fully connected networks, three \\ fully connected layers with 400 units\end{tabular}} \\
\multicolumn{1}{l}{policy initialization}                                                                              & \multicolumn{1}{l}{standard Gaussian}                                                                                                \\
\multicolumn{1}{l}{exploration parameters}                                                                             & \multicolumn{1}{l}{run a uniform exploration policy 50k steps}                                                                      \\
\multicolumn{1}{l}{learning rates for all networks}                                                                    & \multicolumn{1}{l}{3e-4}                                                                                                             \\
\multicolumn{1}{l}{\# of samples / \# of train steps per iteration}                                                    & \multicolumn{1}{l}{10 env steps / 1 training step}                                                                                   \\
\multicolumn{1}{l}{optimizer}                                                                                          & \multicolumn{1}{l}{adam}                                                                                                             \\
\multicolumn{1}{l}{Episode length (horizon)}                                                                           & \multicolumn{1}{l}{500}                                                                                                              \\
\multicolumn{1}{l}{beta for all optimizers}                                                                            & \multicolumn{1}{l}{(0.9, 0.999)}                                                                                                      \\
\multicolumn{1}{l}{discount}                                                                                           & \multicolumn{1}{l}{0.99}                                                                                                             \\
\multicolumn{1}{l}{reward scale}                                                                                       & \multicolumn{1}{l}{1.0}                                                                                                              \\
\multicolumn{1}{l}{temperature}                                                                                        & \multicolumn{1}{l}{learned}                                                                                                          \\
\multicolumn{1}{l}{\begin{tabular}[c]{@{}l@{}}\# of environment steps \\ before adding KL regularization\end{tabular}} & \multicolumn{1}{l}{500k}  \\
\multicolumn{1}{l}{$beta_1$}                                                                                        & \multicolumn{1}{l}{30} \\
\multicolumn{1}{l}{$\beta_2$}                                                                                        & \multicolumn{1}{l}{3e-3}                                                                                                            \\ \hline
\end{tabular}

\end{table}
\subsection{Discussions on Hyper-Parameter Design}
CUP has two additional hyper-parameters compared to SAC, $\beta_1$ and $\beta_2$. We provide some insight on choosing $\beta_1$ and $\beta_2$. Note that the maximum weight for the KL regularization is $\beta_1*\beta_2|\widetilde{V}_{\pi_{tar}^{t}}\left(s\right)|$, and the original actor loss $L_{actor}$ has roughly the same magnitude as $|\widetilde{V}_{\pi_{tar}^{t}}\left(s\right)|$. So $\beta_1*\beta_2$ roughly determines the maximum regularization weight. Following previous works on regularization [1,2], (0.1, 1) is a reasonable range for $\beta_1*\beta_2$. As a consequence, we choose (0.04, 1) as the range of $\beta_1*\beta_2$ for our hyper-parameter ablation studies. What's more, as $\beta_2$ upper bounds the maximum confidence on the expected advantage estimation (Section \ref{3.2}), $\beta_2$ should be decreased if a large variance in performance is observed. These two insights efficiently guide the design of $\beta_1$ and $\beta_2$. As shown in Section \ref{abl}, CUP achieves stable performance on a large range of $\beta_1$s and $\beta_2$s.

\section{Additional Experiment Results}
\subsection{Success Rate Evaluation}
\label{sre}
\begin{figure}[t]
\centering
        \subfigure{\includegraphics[width=0.32\columnwidth]{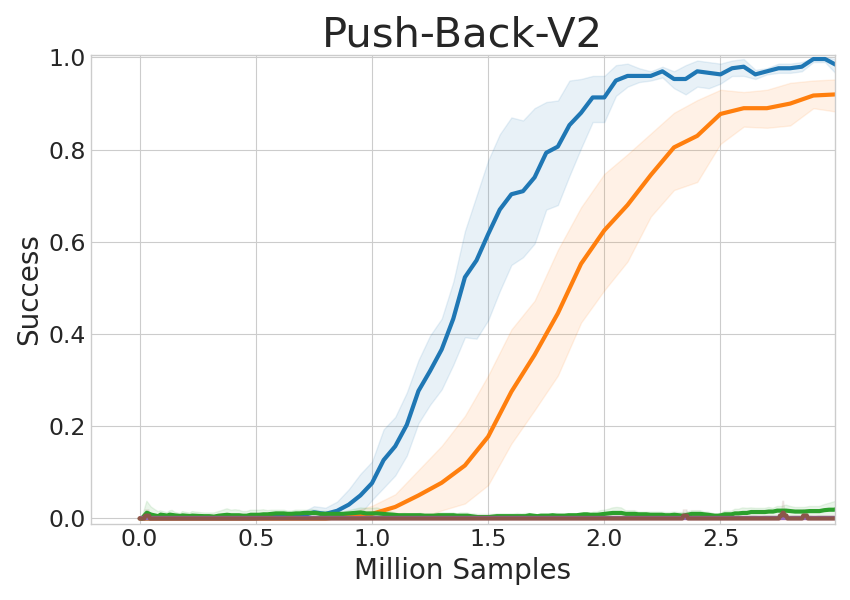}}
        \subfigure{\includegraphics[width=0.32\columnwidth]{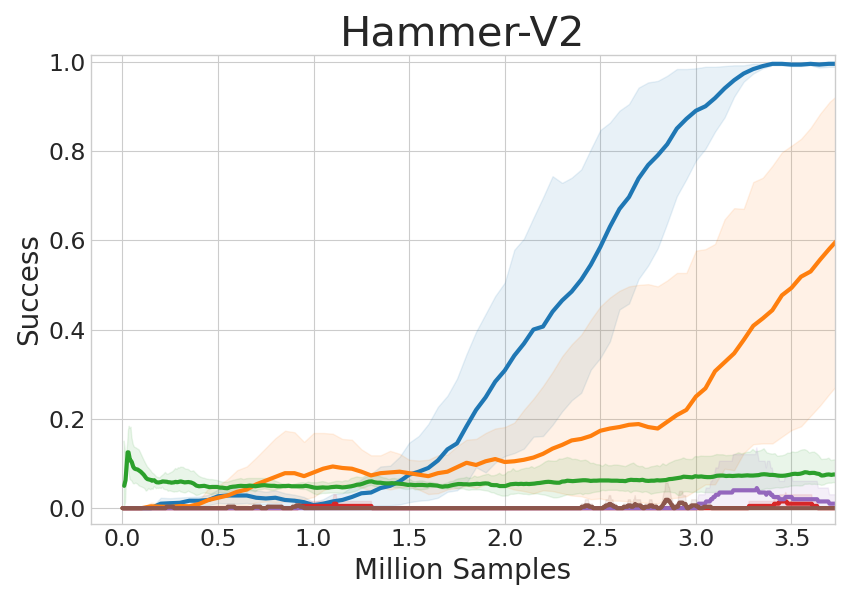}}
        \subfigure{\includegraphics[width=0.32\columnwidth]{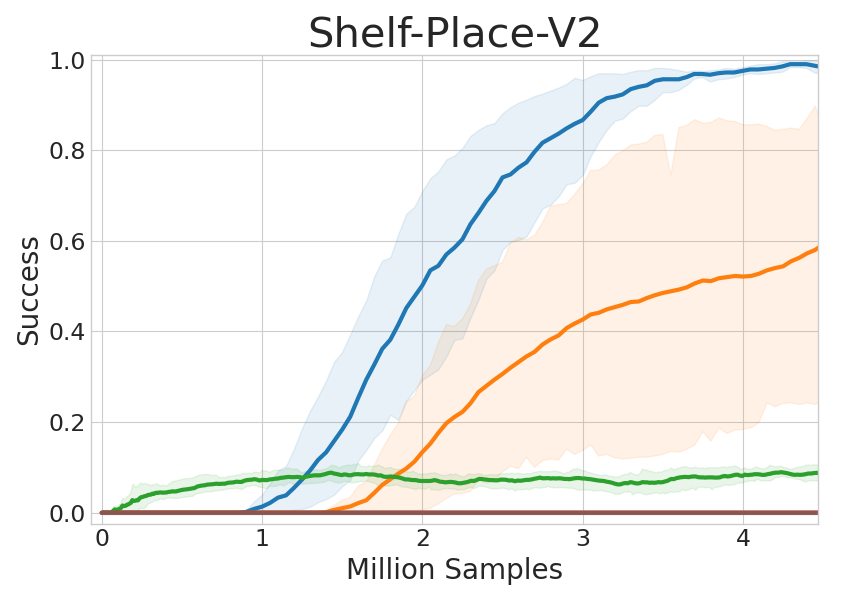}}
        \subfigure{\includegraphics[width=0.32\columnwidth]{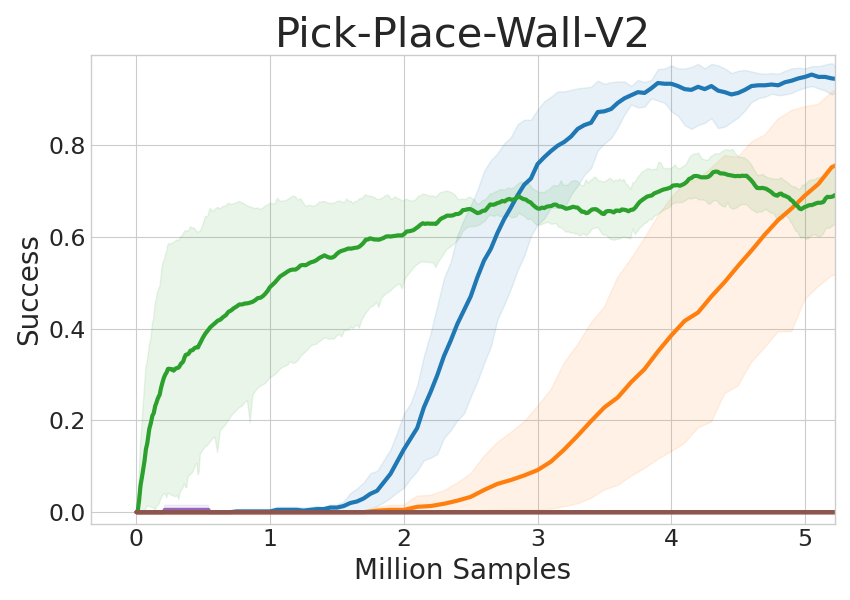}}
        \subfigure{\includegraphics[width=0.32\columnwidth]{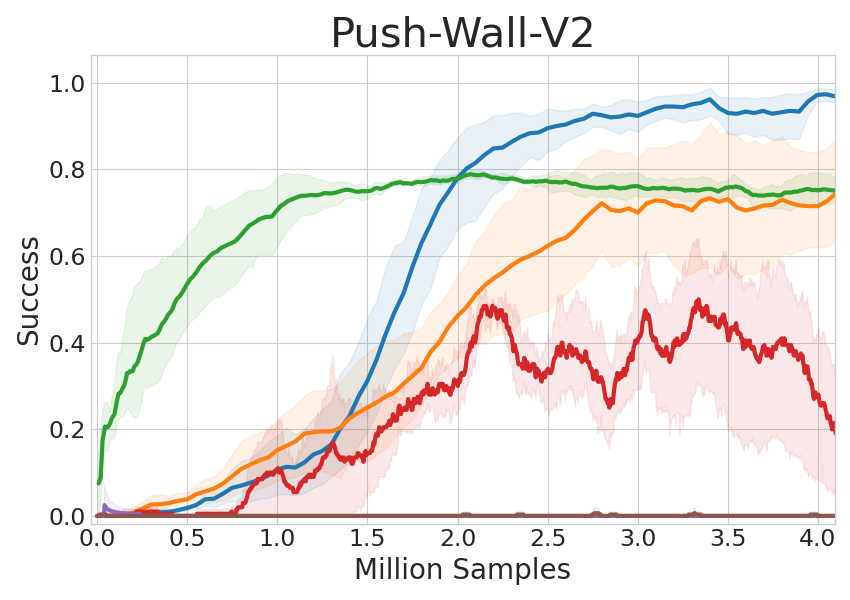}}
        \subfigure{\includegraphics[width=0.32\columnwidth]{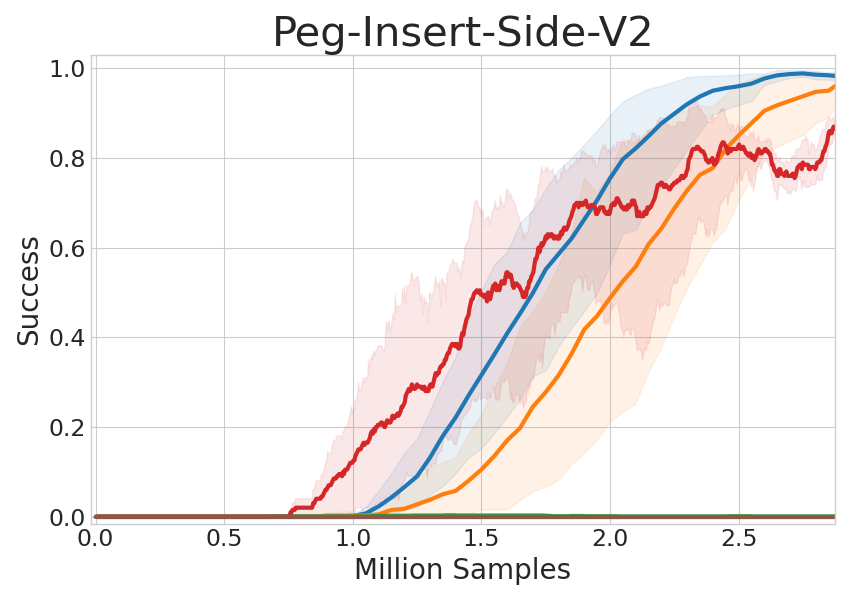}}
        \subfigure{\includegraphics[width=0.95\columnwidth]{figs/legends.png}}
\caption{Algorithm performance evaluated by success rate (three source policies). }
\label{curves4}
\end{figure}

\begin{figure}[t]
\centering
        \subfigure{\includegraphics[width=0.32\columnwidth]{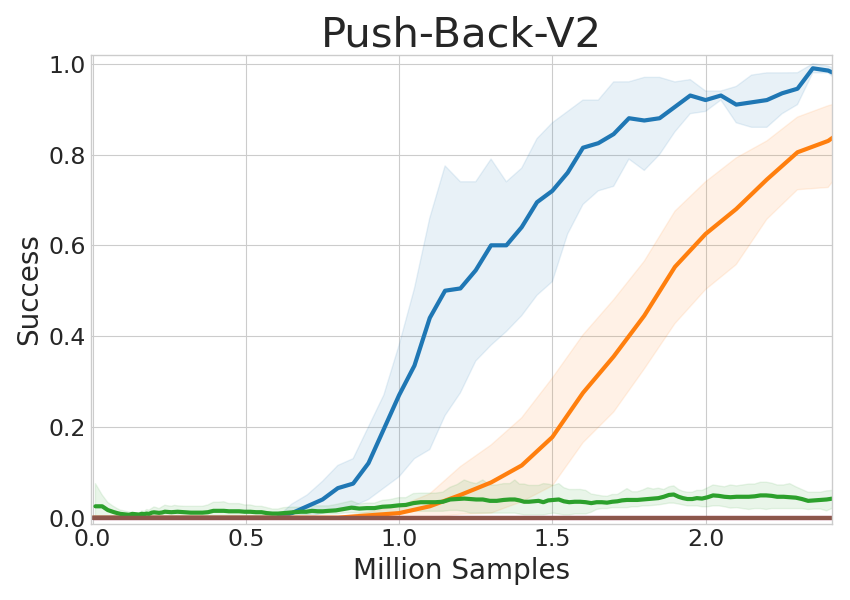}}
        \subfigure{\includegraphics[width=0.32\columnwidth]{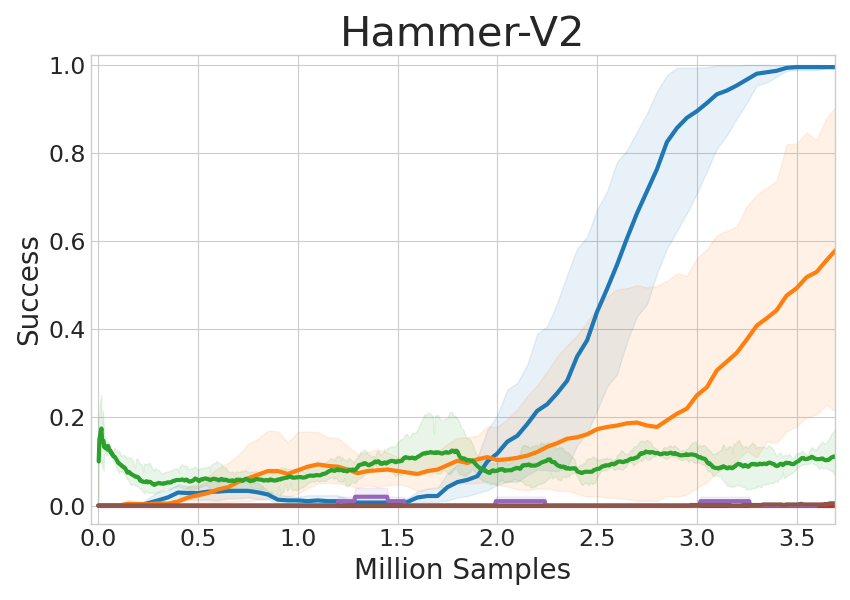}}
        \subfigure{\includegraphics[width=0.32\columnwidth]{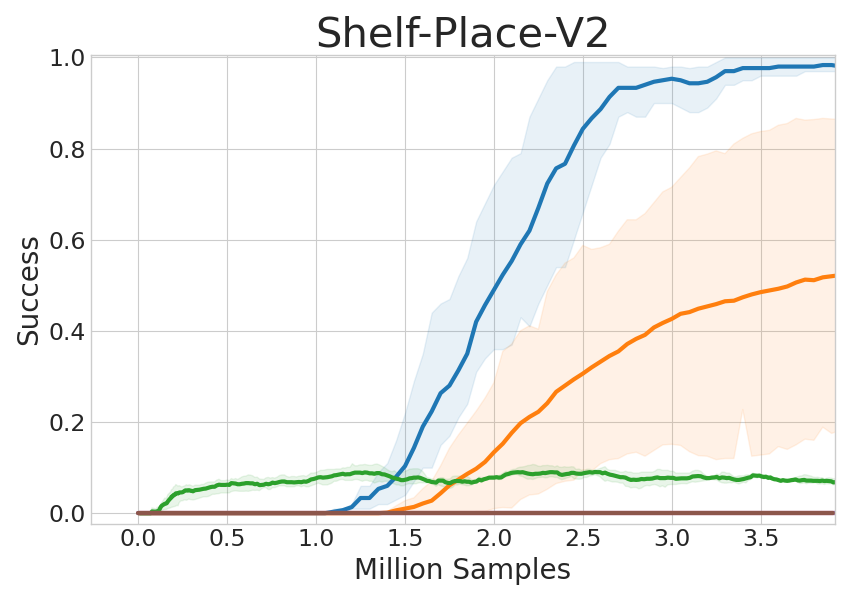}}
        \subfigure{\includegraphics[width=0.32\columnwidth]{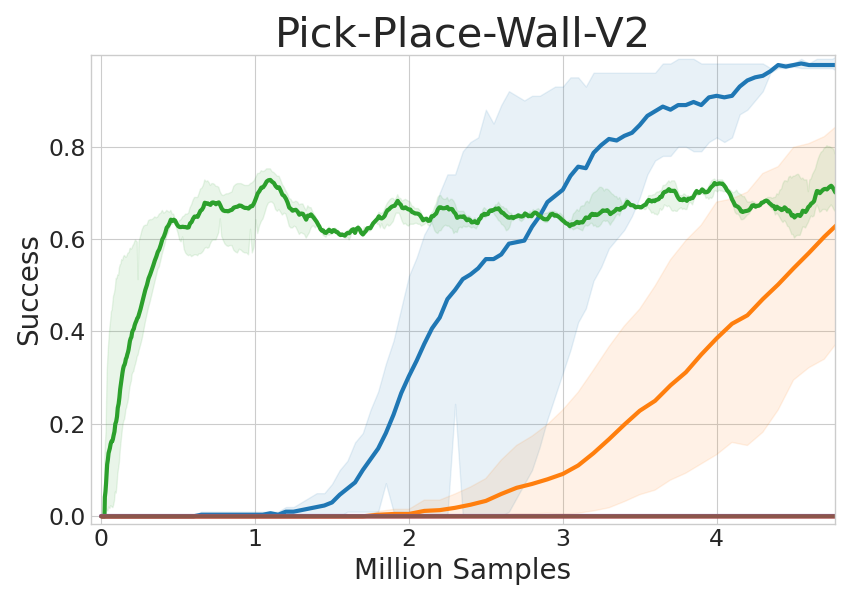}}
        \subfigure{\includegraphics[width=0.32\columnwidth]{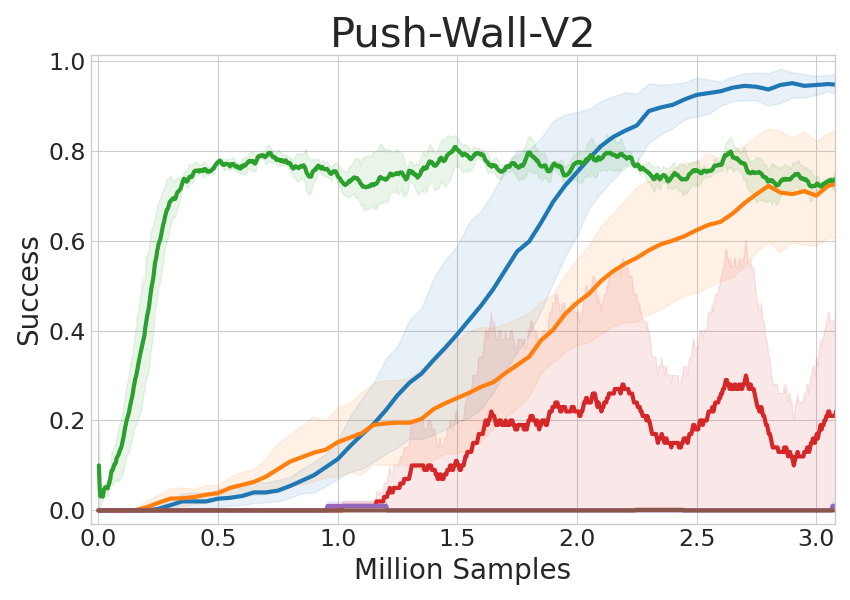}}
        \subfigure{\includegraphics[width=0.32\columnwidth]{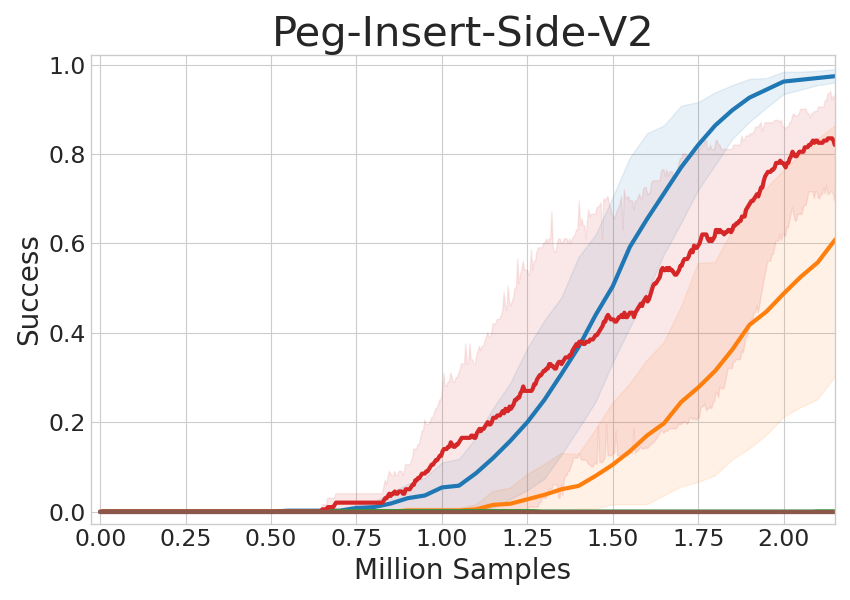}}
        \subfigure{\includegraphics[width=0.95\columnwidth]{figs/legends.png}}
\caption{Algorithm performance evaluated by success rate (six source policies).}
\label{curves22}
\end{figure}

Fig. \ref{curves4} and Fig.\ref{curves22} shows performance evaluated by success rates. The performance is consistent with performance evaluated by cumulative return.

\subsection{Guidance Policy Analysis for All Tasks}
\label{agp}
\begin{figure}[t]
\centering
        \subfigure{\includegraphics[width=0.49\columnwidth]{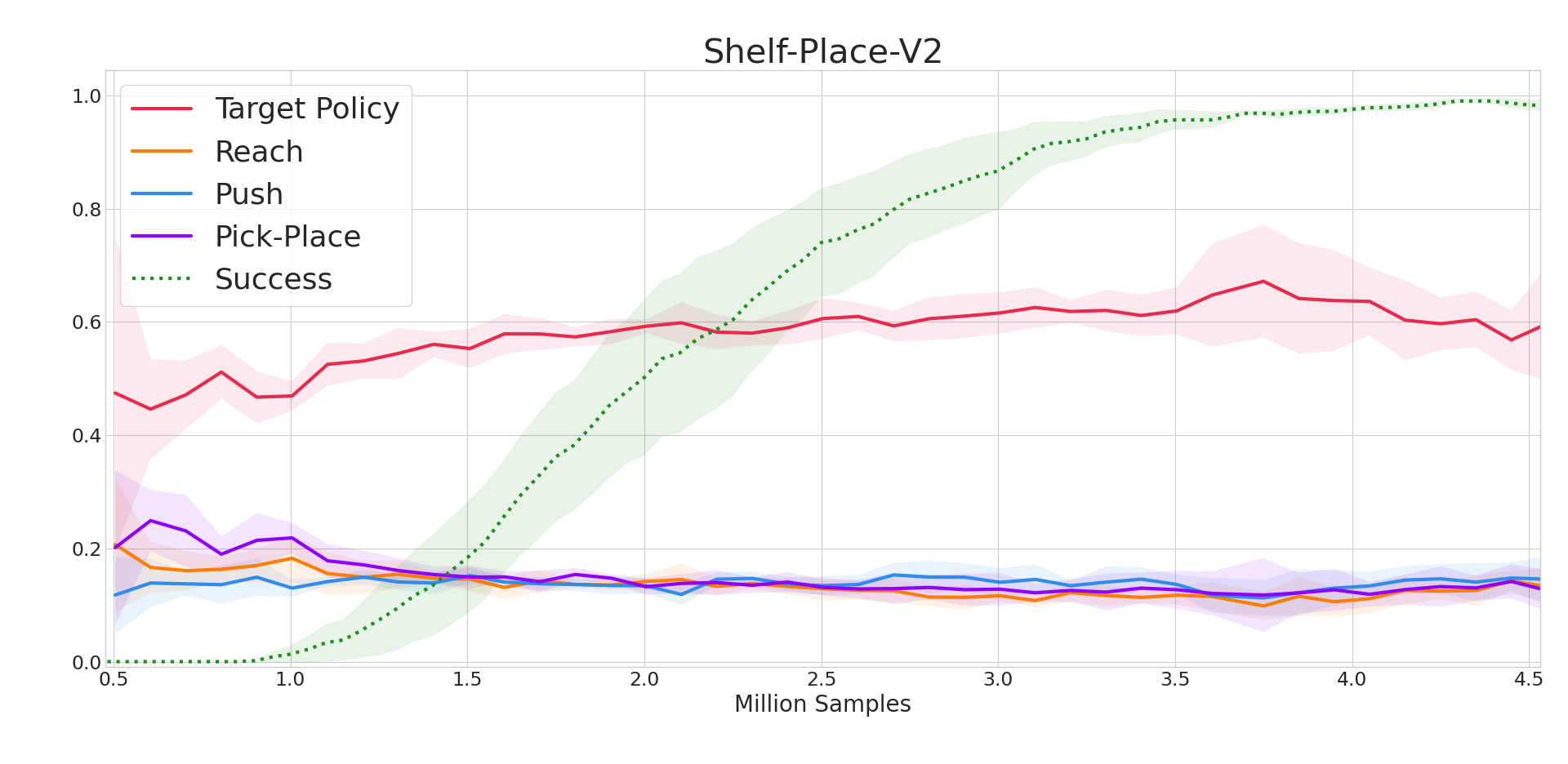}}
        \subfigure{\includegraphics[width=0.49\columnwidth]{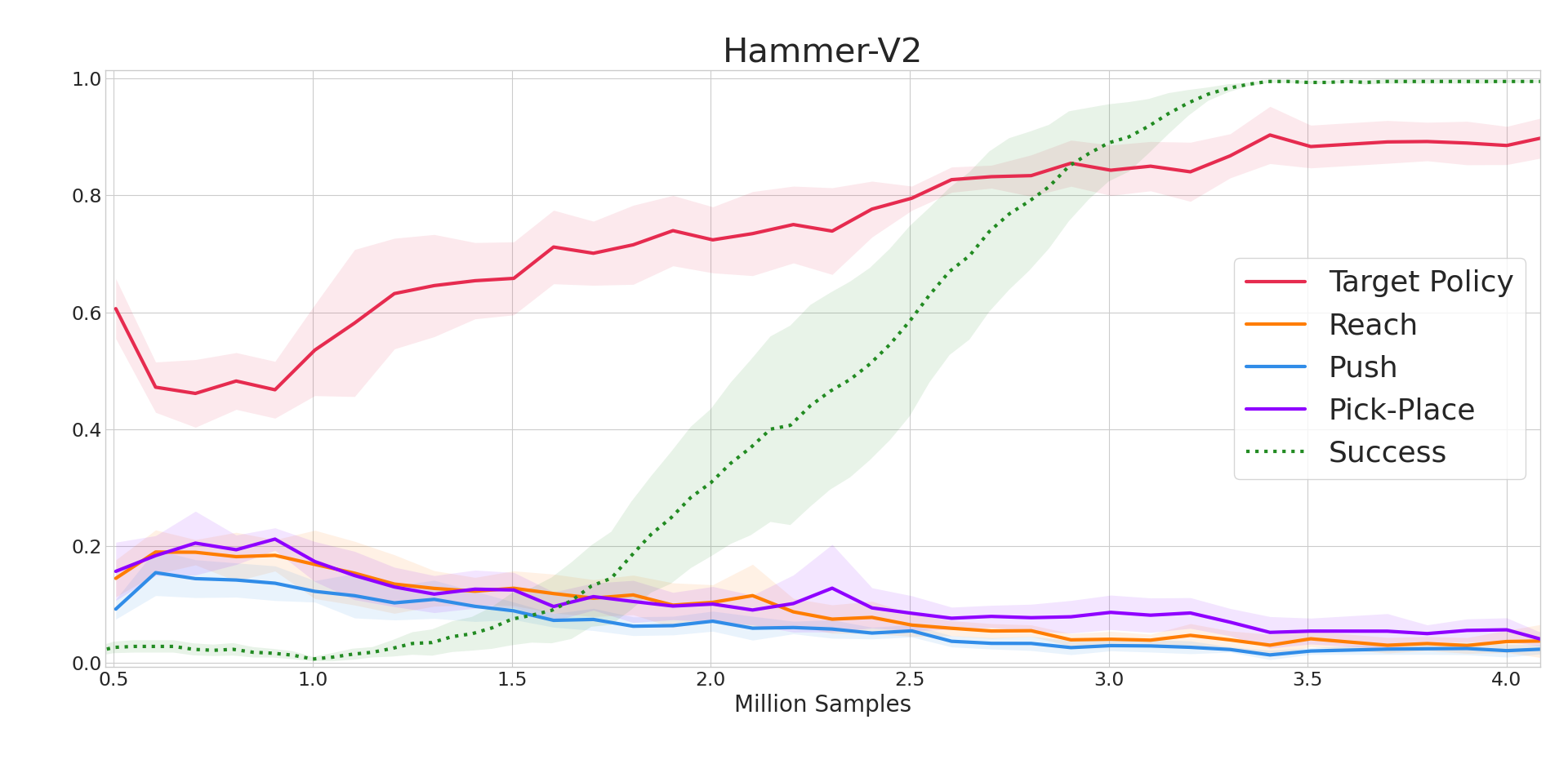}}
        \subfigure{\includegraphics[width=0.49\columnwidth]{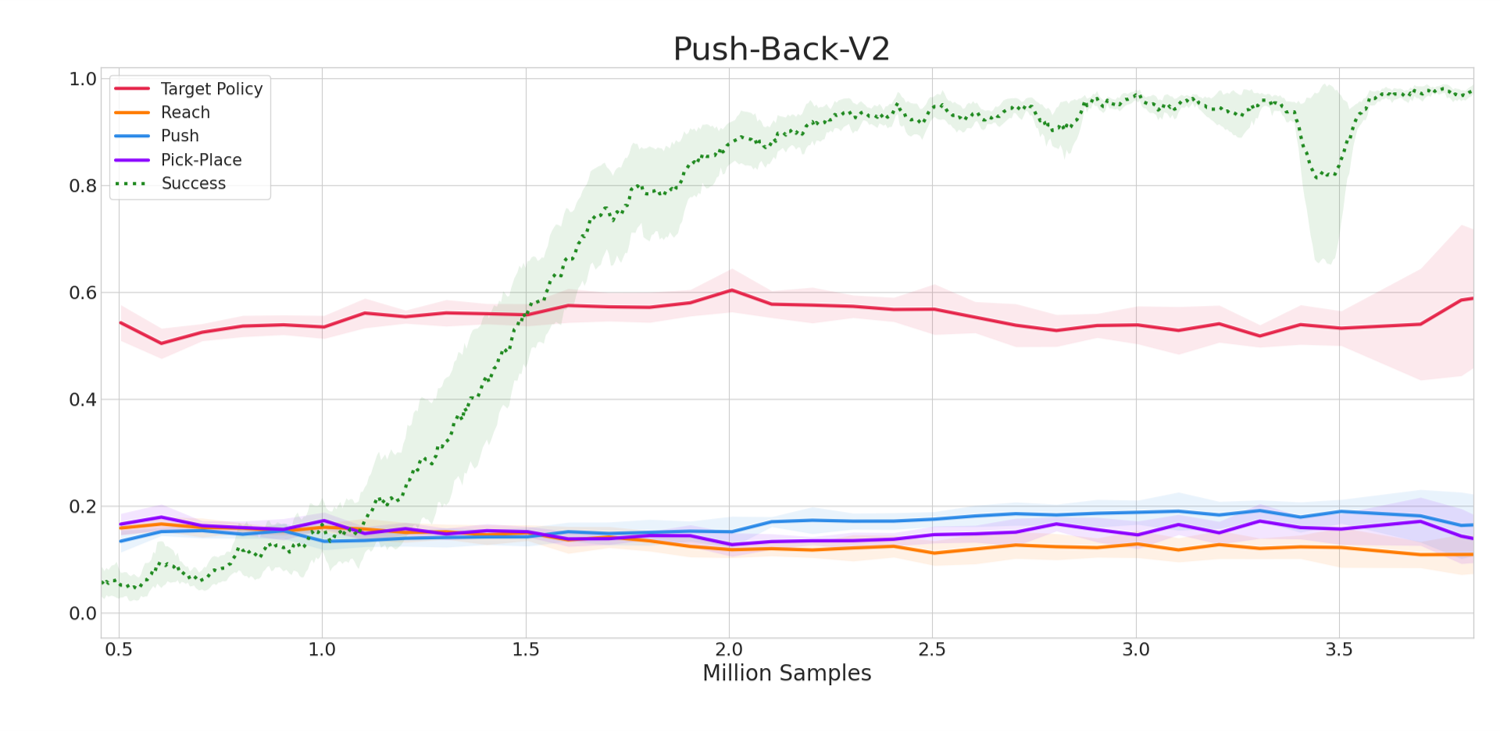}}
        \subfigure{\includegraphics[width=0.49\columnwidth]{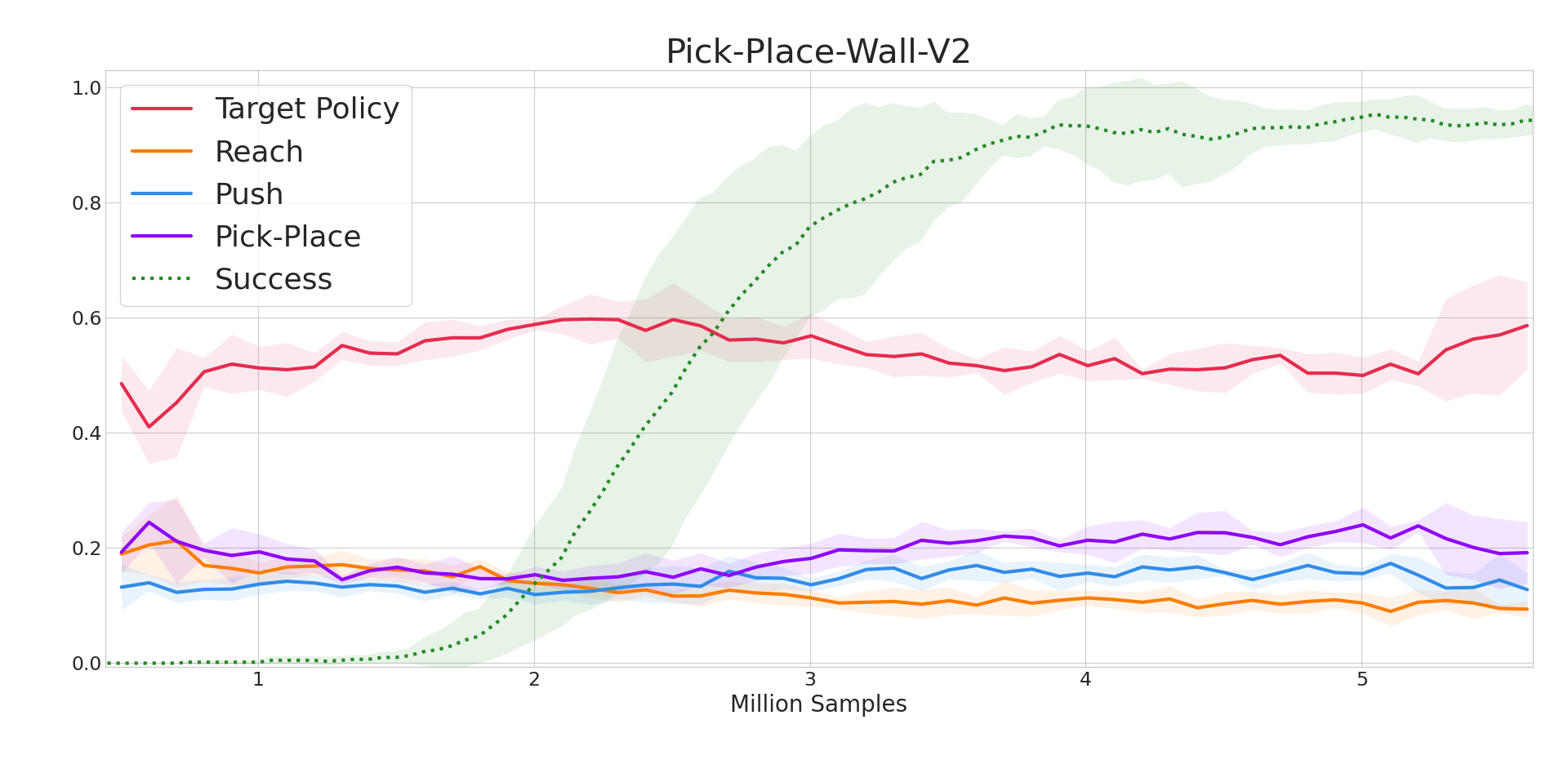}}
        \subfigure{\includegraphics[width=0.49\columnwidth]{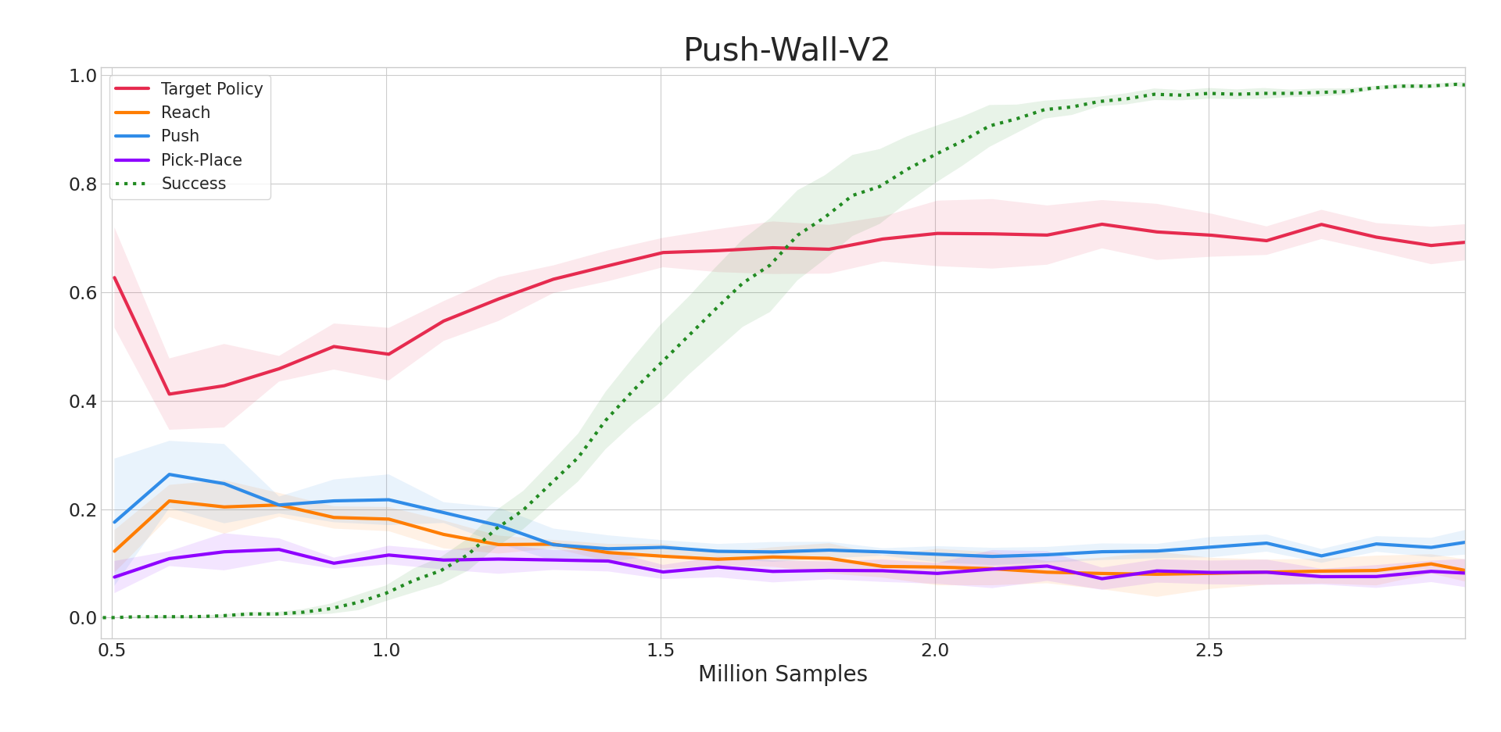}}
        \subfigure{\includegraphics[width=0.49\columnwidth]{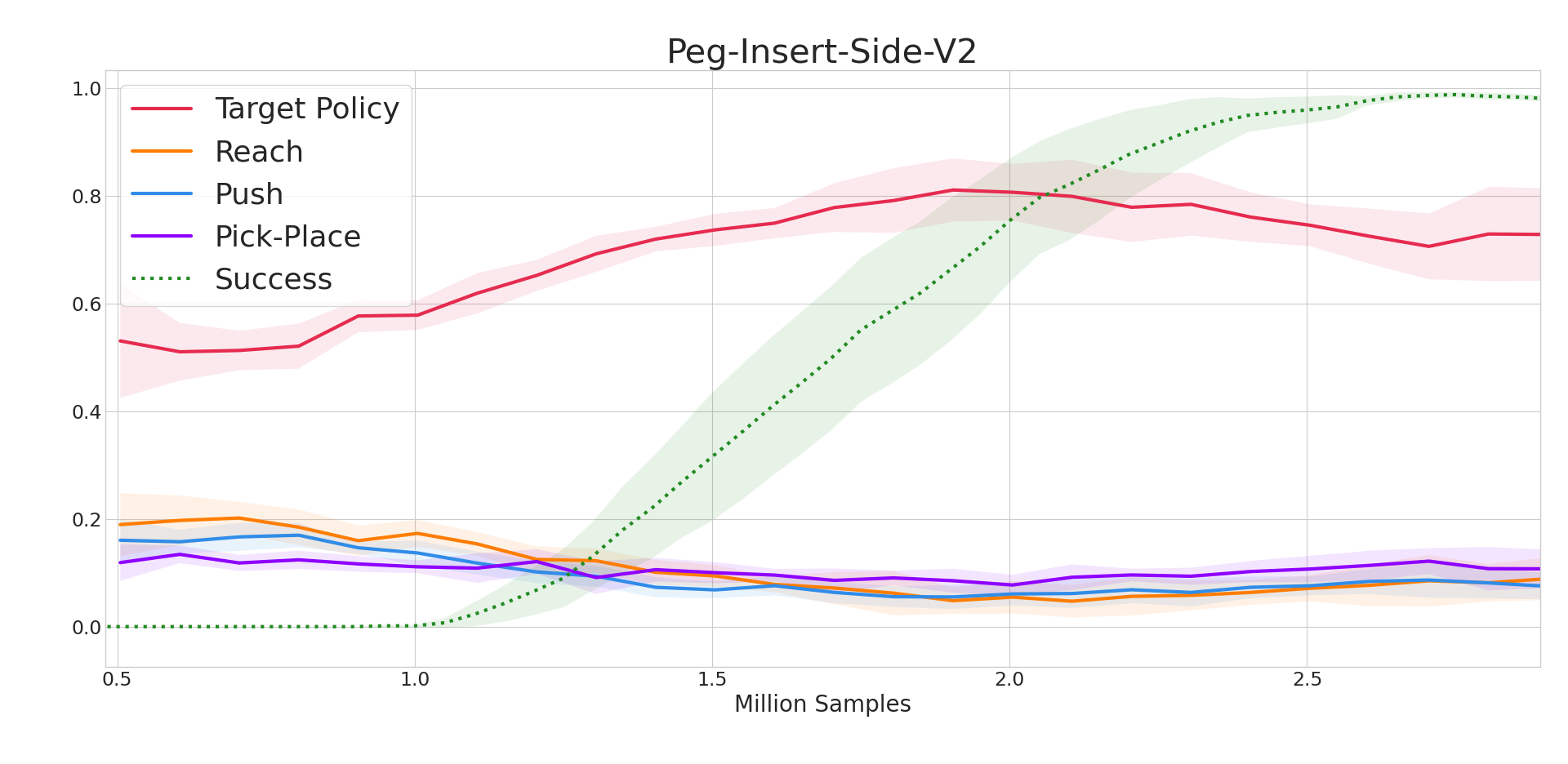}}
\caption{Percentages of source policies being selected by CUP during training on all six tasks. The green dashed line represents the target policy's success rate on the task.}
\label{allvis1}
\end{figure}

\begin{figure}[t]
\centering
        \subfigure{\includegraphics[width=0.49\columnwidth]{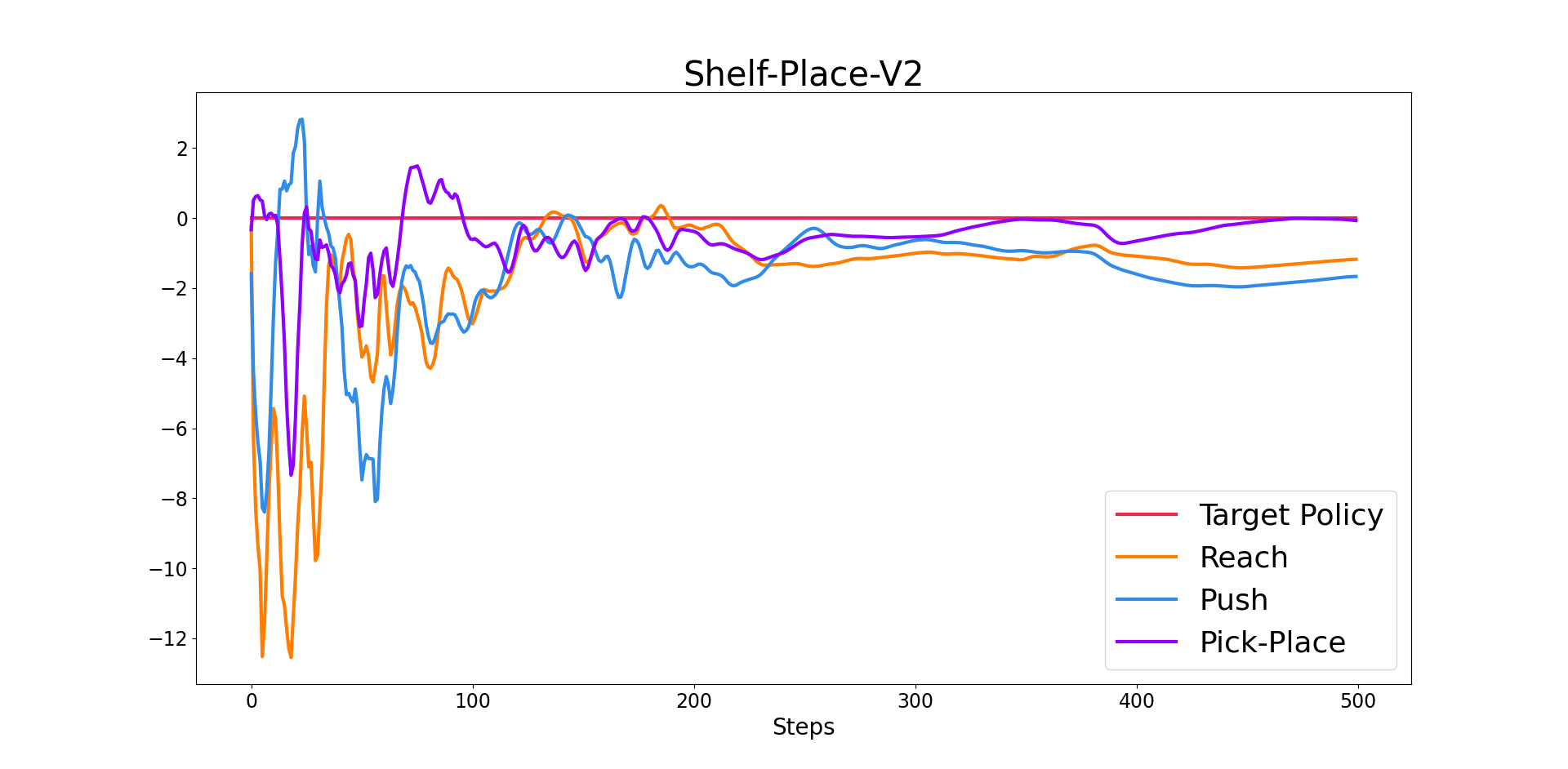}}
        \subfigure{\includegraphics[width=0.49\columnwidth]{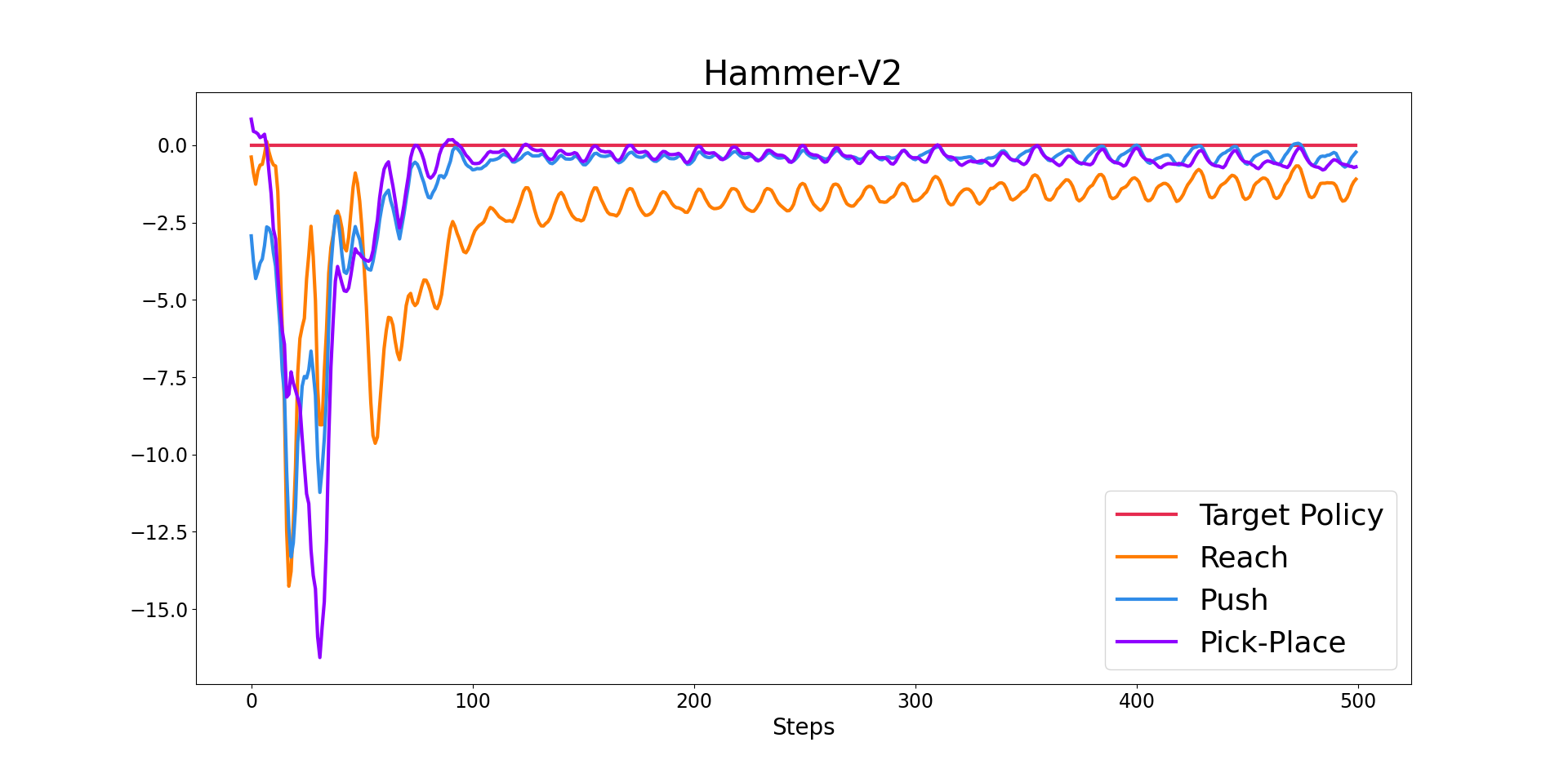}}
        \subfigure{\includegraphics[width=0.49\columnwidth]{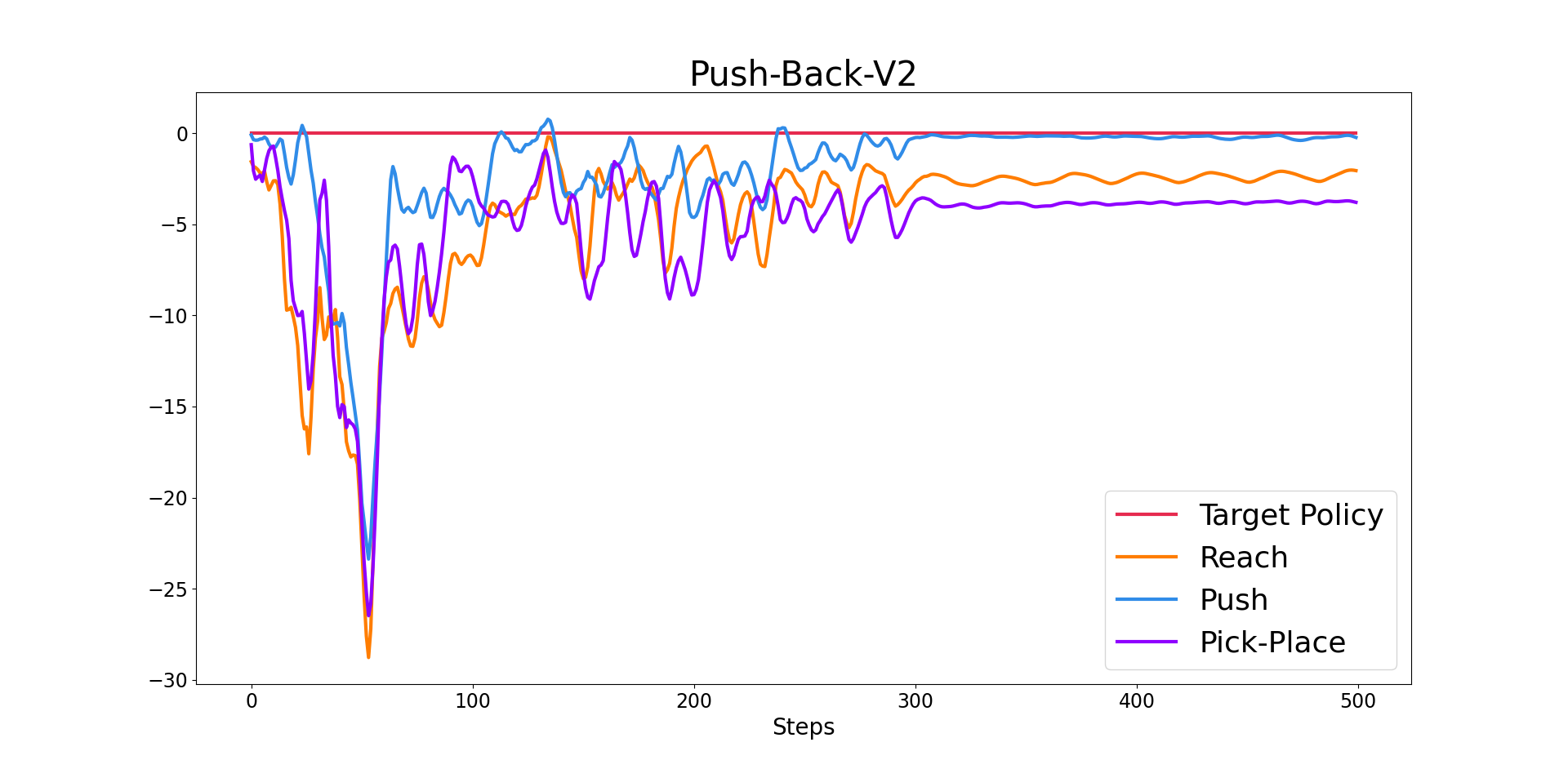}}
        \subfigure{\includegraphics[width=0.49\columnwidth]{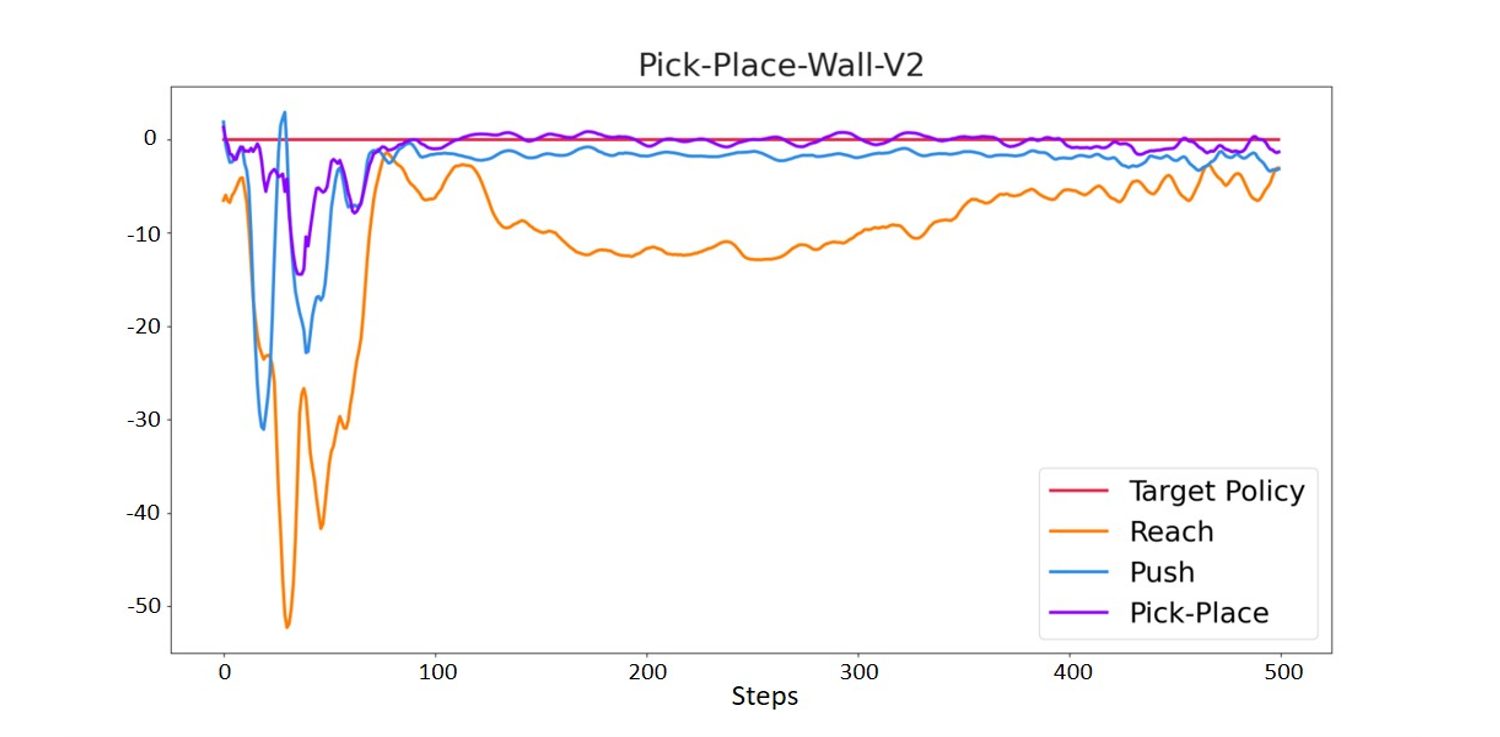}}
        \subfigure{\includegraphics[width=0.49\columnwidth]{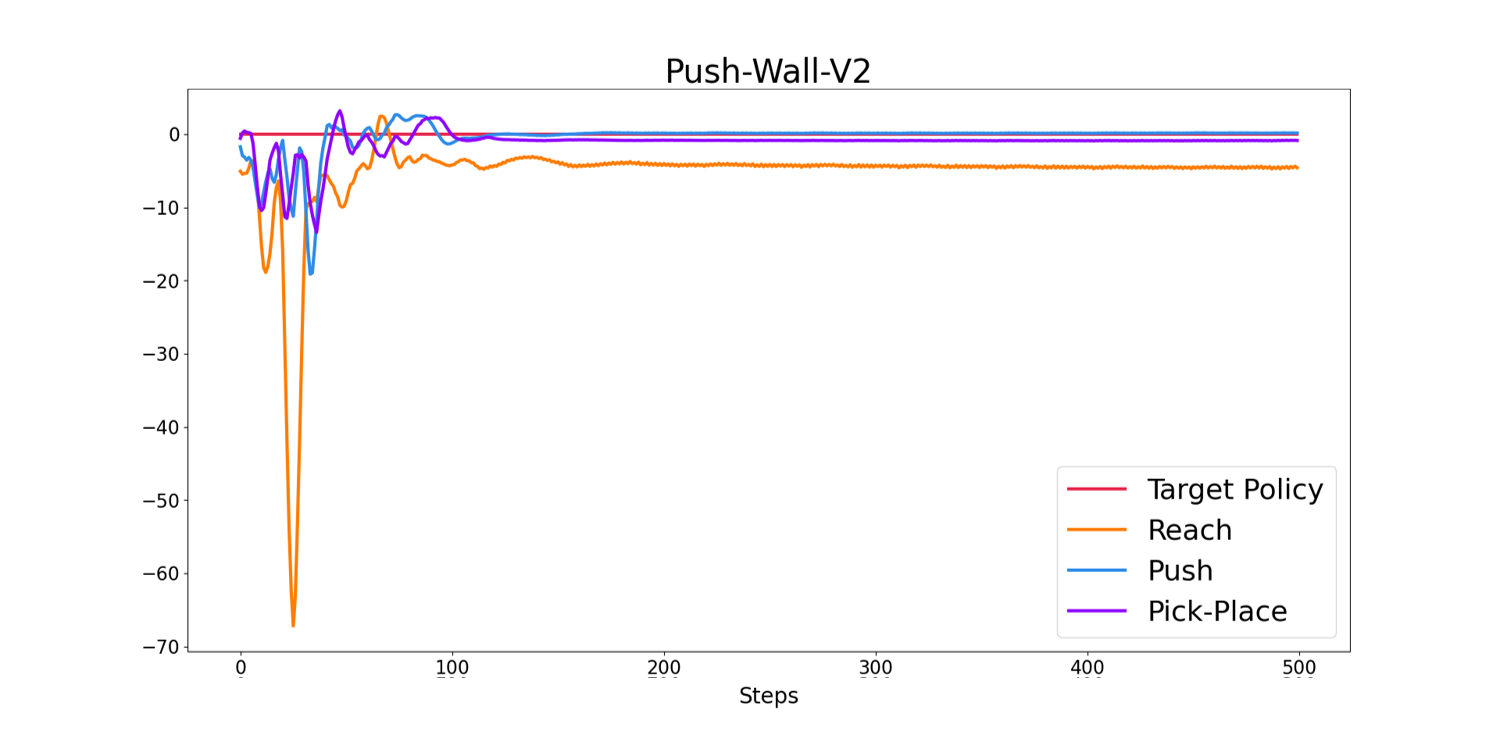}}
        \subfigure{\includegraphics[width=0.49\columnwidth]{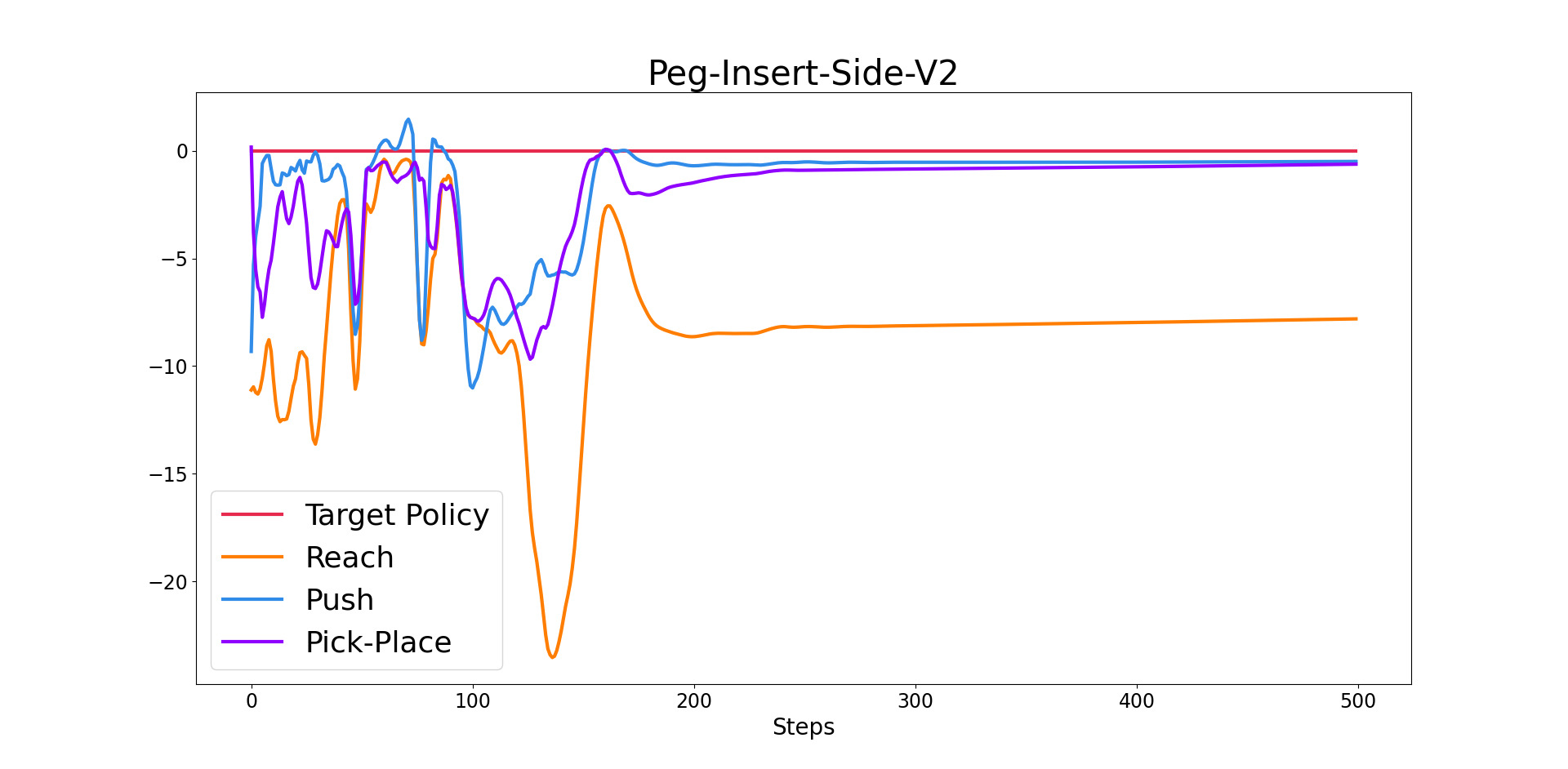}}
\caption{Expected advantages of source policies at convergence on all six tasks. The horizontal axis represents the environment steps of an episode.}
\label{allvis2}
\end{figure}

This subsection provides analyses of guidance policies on all six tasks, as shown in Fig. \ref{allvis1} and Fig. \ref{allvis2}. Results demonstrate that the Pick-Place source policy is the most useful in Shelf-Place, Hammer, and Pick-Place-Wall, while the Push source policy is the most useful in Push-Back and Push-Wall. In Peg-Insert-Side, both source policies are of similar usefulness.

\subsection{CUP's Ability to Use Additional Source Policies}
\label{CAU}
\begin{figure}[tb]
\centering
        \includegraphics[width=0.75\columnwidth]{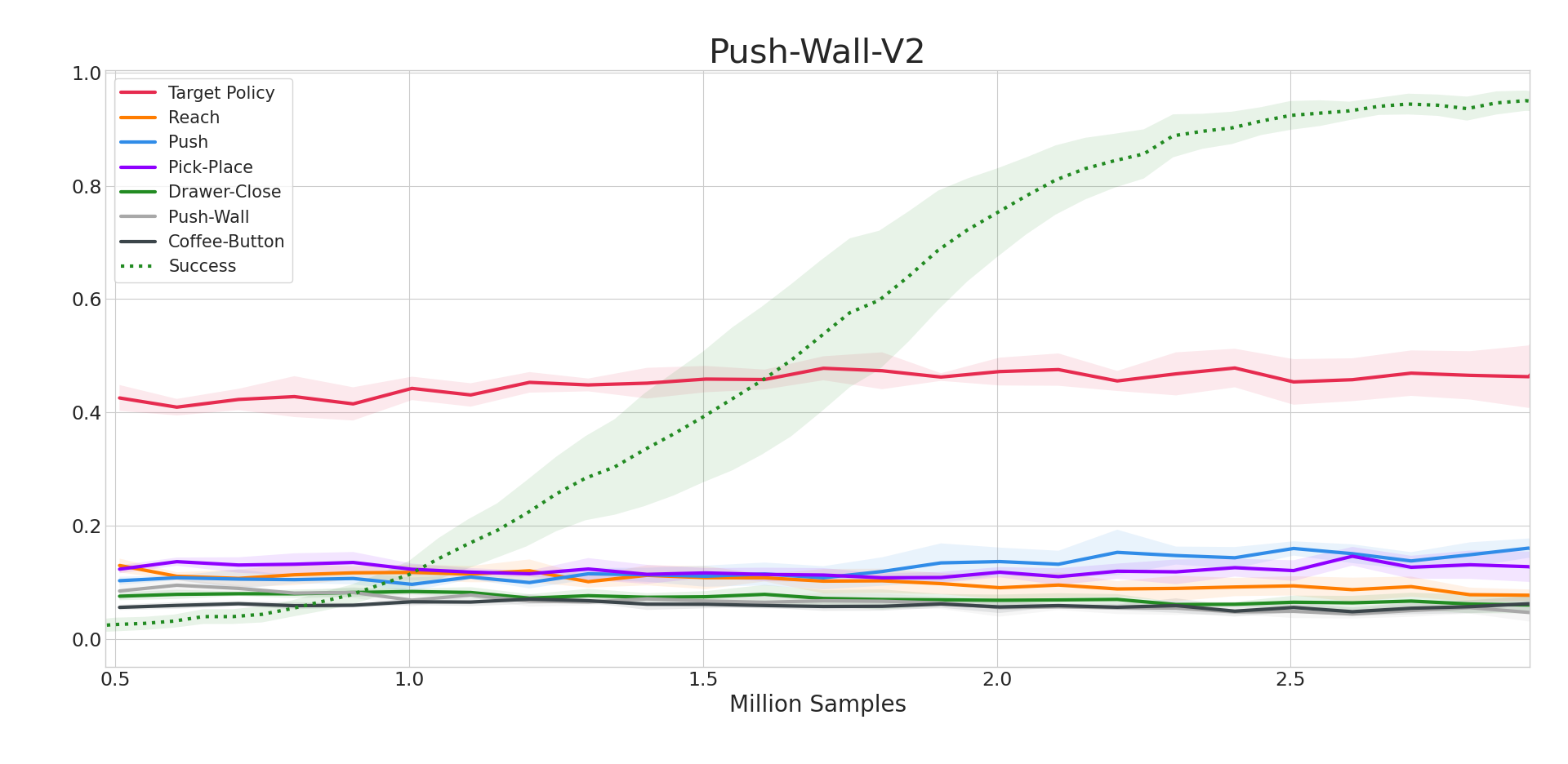}
\caption{Percentages of source policies being selected by CUP during training on Push-Wall with the original set of six source policies. The additional source policies are seldom selected, which suggests that the first three source policies already provide sufficient support for policy reuse.}  \label{ori6source}
\end{figure}

\begin{figure}[tb]
\centering
        \includegraphics[width=0.6\columnwidth]{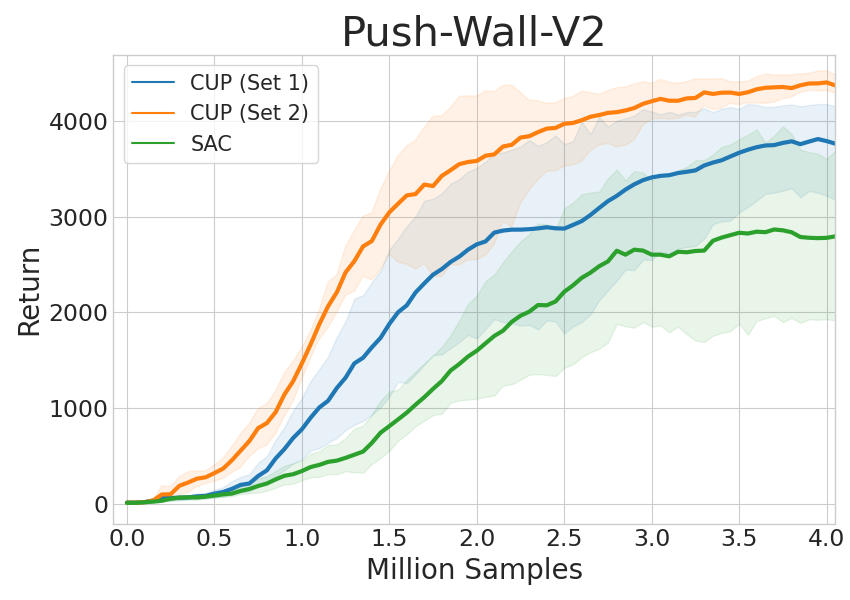}
\caption{CUP's performance on another two sets of source policies. CUP can efficiently utilize the additional source policies contained in Set 2. }  \label{additionalsource}
\end{figure}

 In Fig. \ref{curves2} the improvement brought by additional source policies is mild, because the original three source policies have already provided sufficient support for policy reuse, as demonstrated in Fig. \ref{ori6source}. To evaluate CUP's ability to utilize additional source policies, we design another two sets of source policies. Set 1 consists of three source policies that solve Reach, Peg-Insert-Side, and Hammer, while Set 2 adds source policies trained on Push-Back, Pick-Place-Wall, and Shelf-Place to Set 1. As Set 1 is less related to our target task Push-Wall, CUP must utilize the additional source policies in Set 2 to improve its performance. As demonstrated in Fig. \ref{additionalsource}, CUP can efficiently take advantage of the additional source policies to achieve efficient learning. 

\subsection{CUP's Source Policy Selection on the Source Task}
\begin{figure}[tb]
\centering
        \includegraphics[width=0.75\columnwidth]{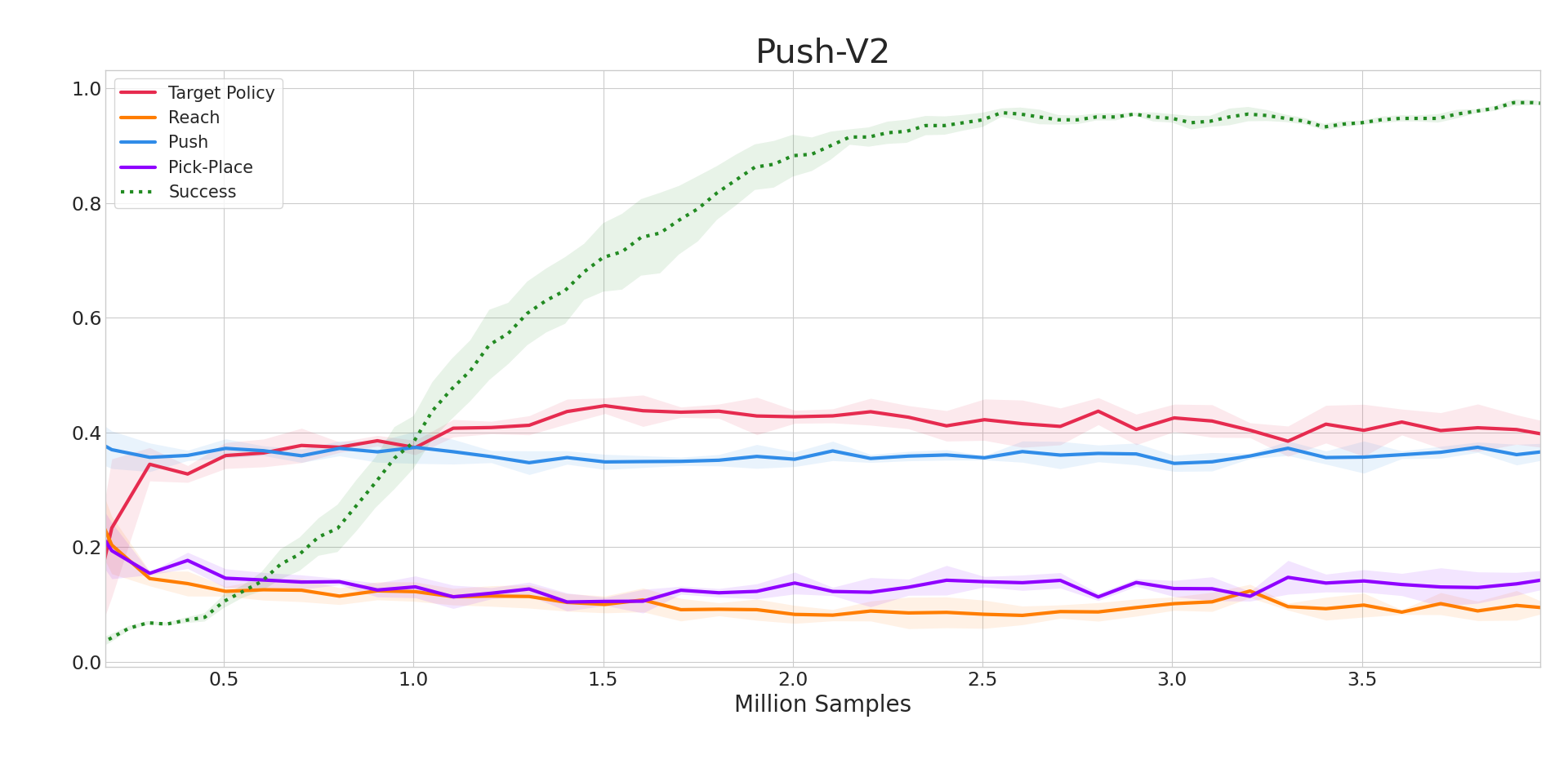}
\caption{Percentages of source policies being selected by CUP during training on Push. The green dashed line represents the target policy’s success rate on the task. The Push source policy is selected far more frequently than the other two source policies, and is selected for roughly the same frequency as the target policy at convergence.}  \label{pushonpush}
\end{figure}

To further investigate CUP's formation of the guidance policy, we train CUP on one of the source tasks, Push. As shown in Fig. \ref{pushonpush}, the corresponding source policy Push is selected frequently. After the target policy converges, CUP selects the target policy and the Push policy for roughly the same frequency, as they can both solve the task. 

\subsection{Experiments on Additional Tasks}
\begin{figure}[tb]
\centering
        \includegraphics[width=0.45\columnwidth]{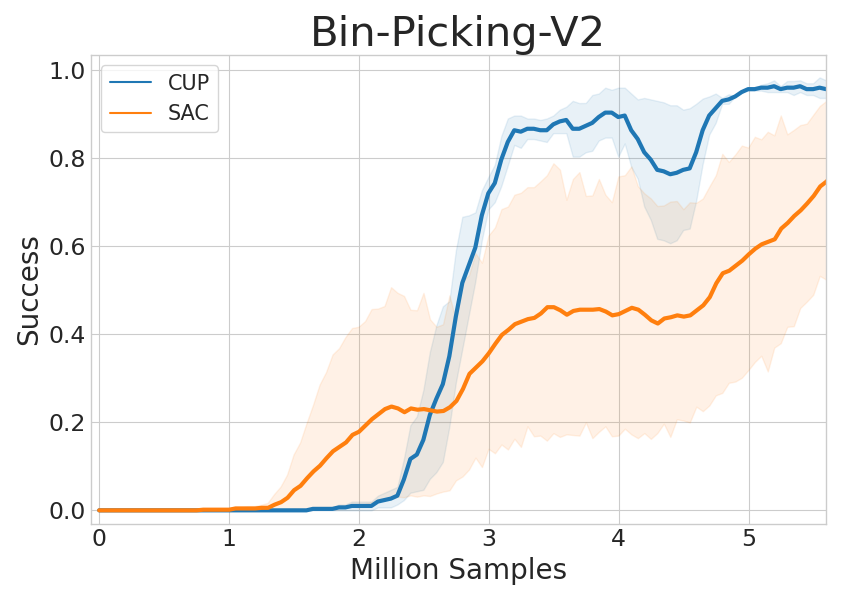}
        \includegraphics[width=0.45\columnwidth]{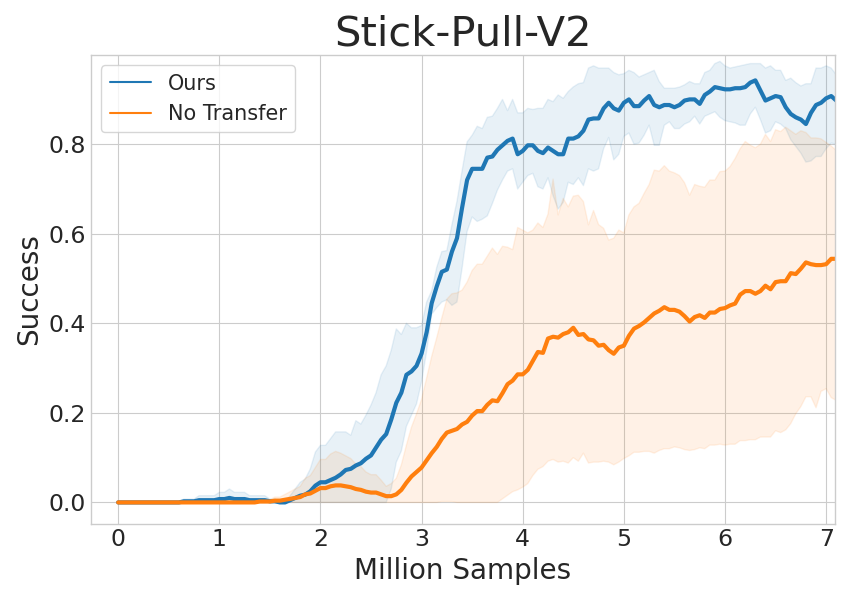}
\caption{Performance of CUP and SAC on two harder Meta-World tasks that require more environment steps to converge.}  \label{additionalexp}
\end{figure}

\begin{figure}[tb]
\centering
        \includegraphics[width=0.45\columnwidth]{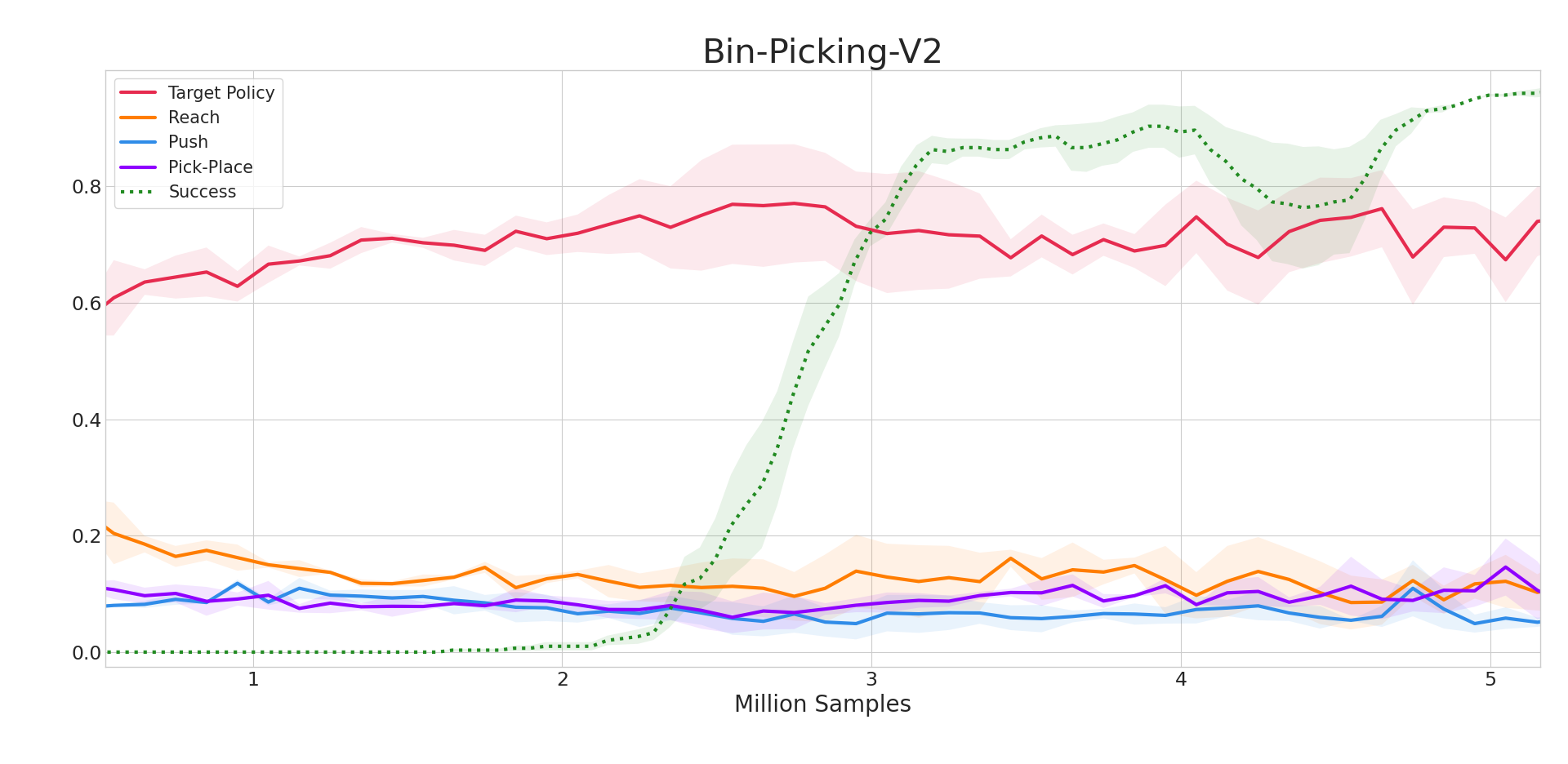}
        \includegraphics[width=0.45\columnwidth]{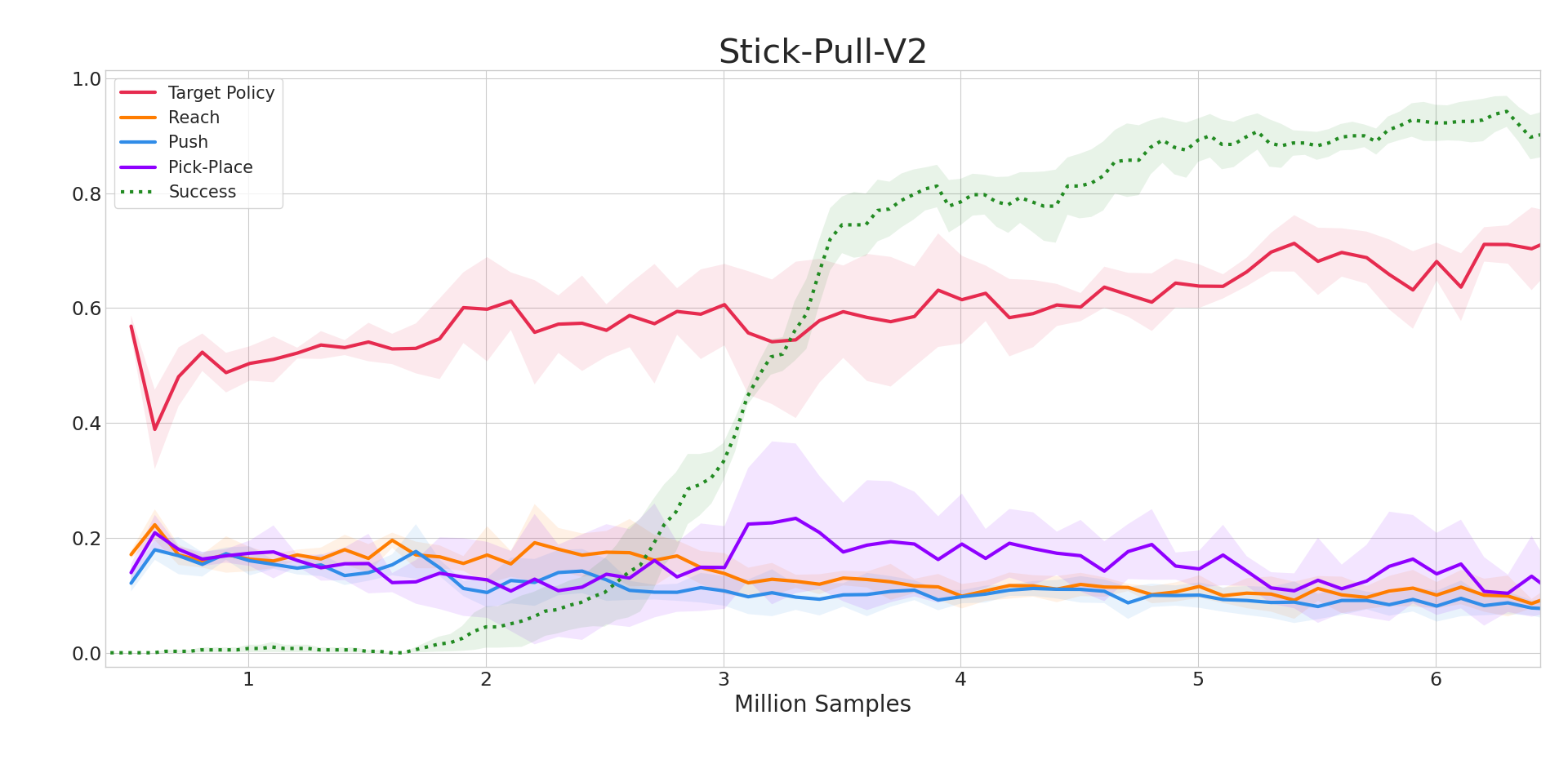}
\caption{Percentages of source policies being selected by CUP during training. In these two tasks, the source policies are chosen less frequently, which implies that source policies are less related to these tasks.}  \label{additionalexpany}
\end{figure}

We evaluate CUP on Bin-Picking and Stick-Pull, two tasks less related to the source policies. As demonstrated in Fig. \ref{additionalexp}, in this harder setting, CUP's performance improvement over SAC is smaller. To investigate this, we provide an analysis on CUP's source policy selection. As shown in Fig. \ref{additionalexpany}, the smaller improvement is because that source policies are less related to the target tasks, as they are selected less frequently.

\subsection{Analyzing Non-Stationarity in HRL Methods}
\label{anay}
\begin{figure}[tb]
\centering
        \subfigure[]{\includegraphics[width=0.49\columnwidth]{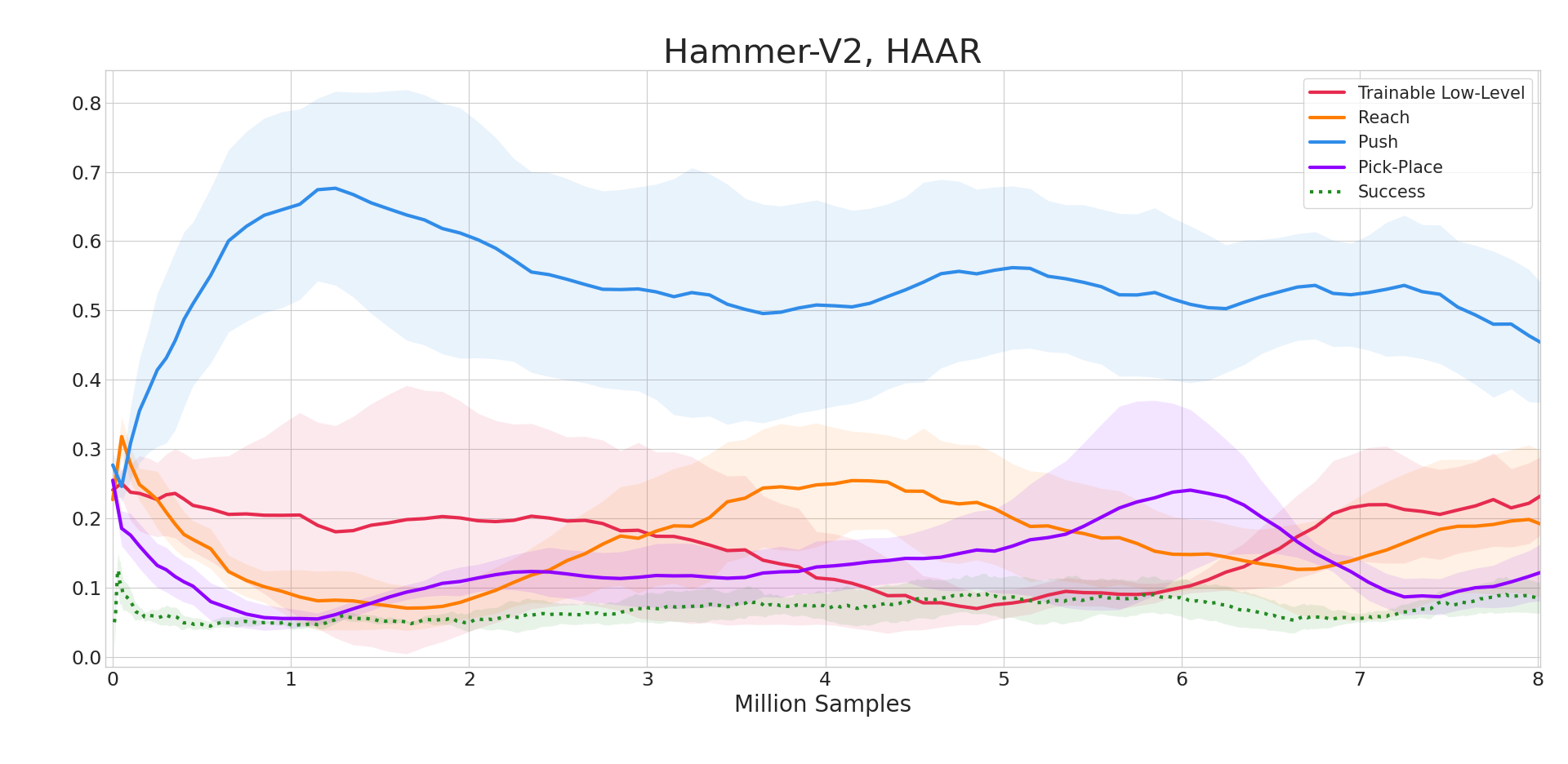}\label{analfig1}}
        \subfigure[]{\includegraphics[width=0.49\columnwidth]{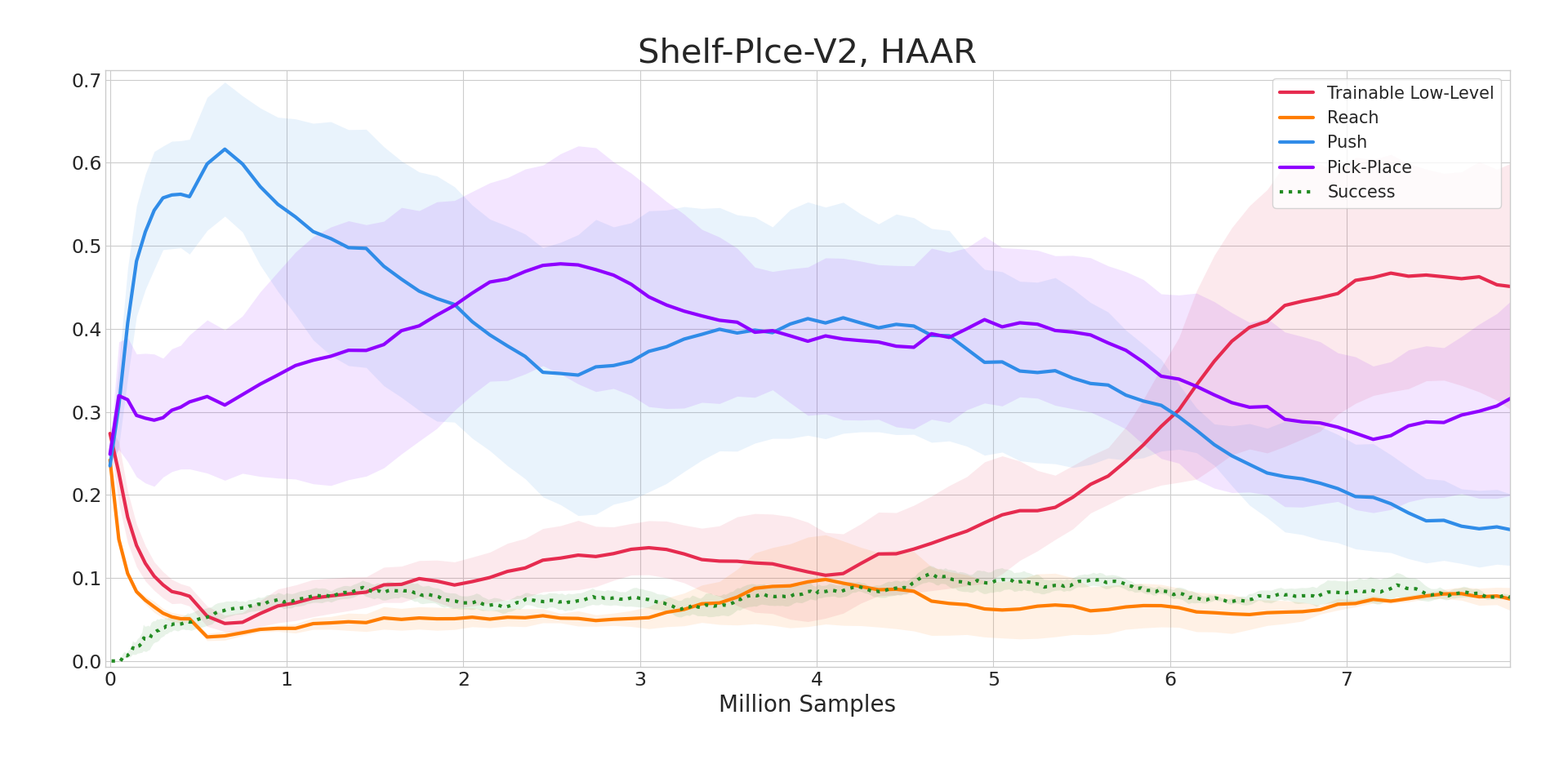}\label{analfig2}}
        \subfigure[]{\includegraphics[width=0.49\columnwidth]{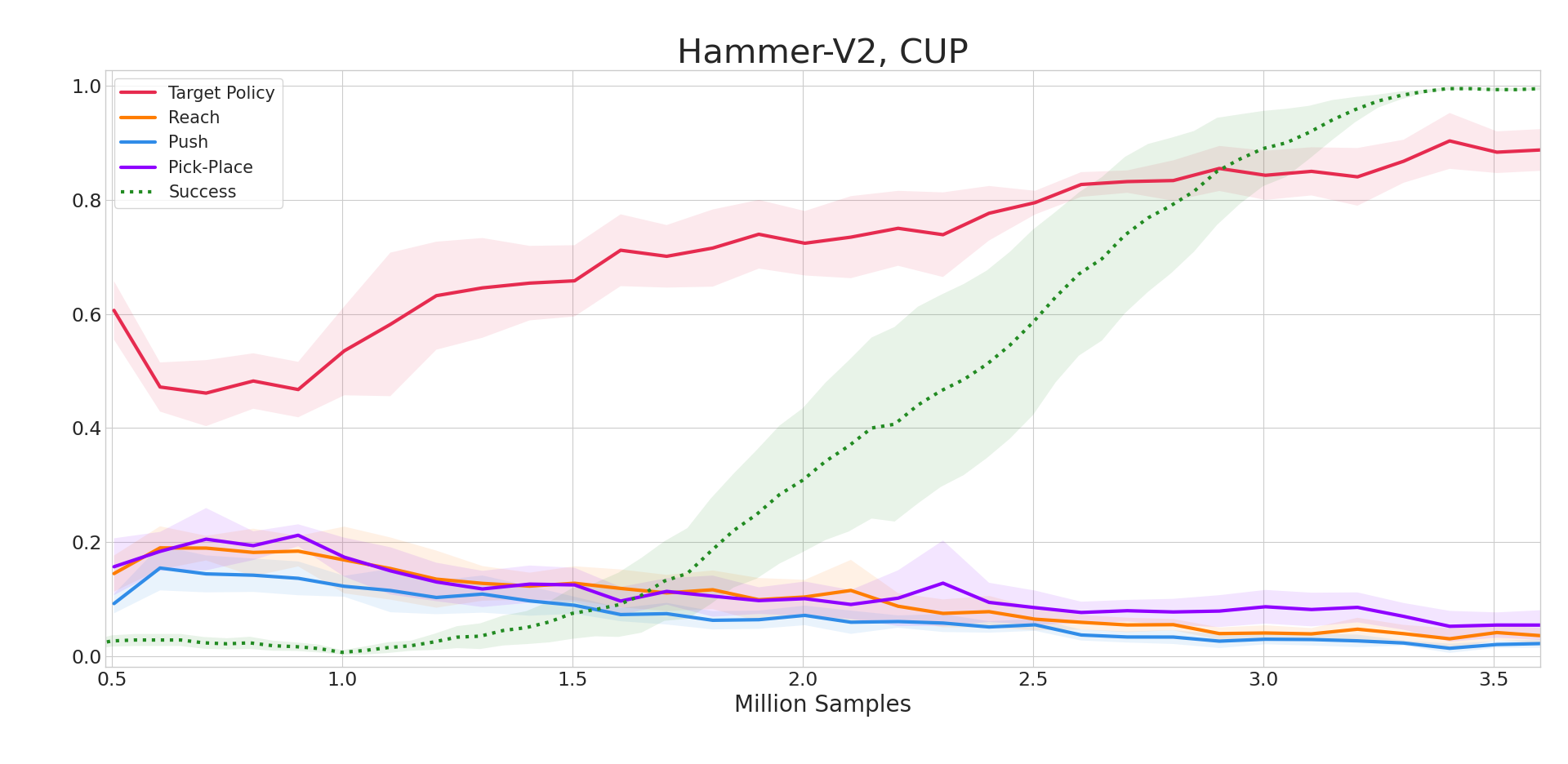}\label{analfig3}}
        \subfigure[]{\includegraphics[width=0.49\columnwidth]{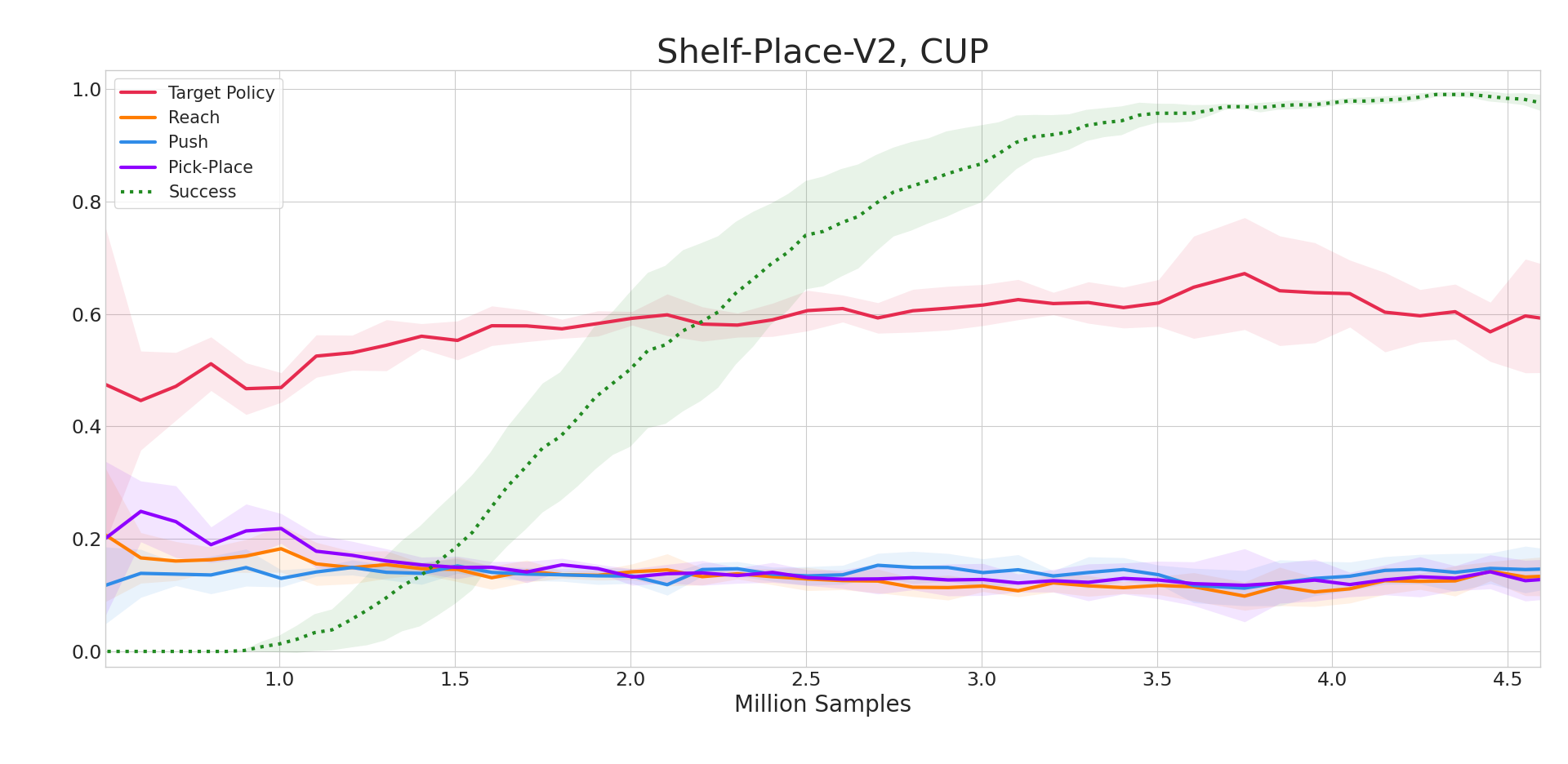}\label{analfig4}}
\caption{Comparison between HAAR and CUP's source policy selection on two representative tasks. Results are averaged over six random seeds. (a) and (b) demonstrates the percentages of each low-level policy being selected by HAAR's high-level policy. ``Trainable Low-Level'' is HAAR's additional trainable low-level policy, as mentioned in Appendix \ref{anay}. (c) and (d) demonstrates the the percentages of each source policy being selected by CUP. While HAAR suffers from the non-stationarity problem and has a large variance in source policy selection, CUP is much more stable and achieves superior performance, as CUP avoids the non-stationarity problem by avoiding training high-level policies.}
\label{analfig}
\end{figure}

To further analyze the advantages of CUP and demonstrate the non-stationarity problem of HRL methods, we illustrate the percentages of each low-level policy being selected by HAAR's high-level policy. HAAR's low-level policy set consists of the three source policies and an additional trainable low-level policy, which is expected to be selected at states where no source policies give useful actions. As demonstrated in Fig. \ref{analfig1} and Fig. \ref{analfig2}, HAAR's low-level policy selection suffers from a large variance over different random seeds, and oscillates over time. This is because that as the low-level policy keeps changing, the high-level transition becomes non-stationary and leads to unstable learning. In comparison, as shown in Fig. \ref{analfig3} and Fig. \ref{analfig4}, CUP's source policy selection is much more stable and achieves superior performance, as it selects source policies according to expected advantages instead of high-level policies, and avoids the non-stationarity problem.

\end{document}